\documentclass[twoside,11pt]{article}

\usepackage{bbm}
\usepackage{jmlr2e}
\usepackage{amsmath}
\usepackage{xcolor}
\usepackage{bm}
\usepackage{tikz}
\usetikzlibrary{bayesnet}
\usepackage{subcaption}
\usepackage{float}
\usepackage{stackengine}
\usepackage{array}
\usepackage{multirow}
\usepackage{stmaryrd}
\usepackage{pgfplots}
\usepgfplotslibrary{fillbetween}
\usetikzlibrary{patterns}
\usepackage[outdir=./]{epstopdf}

\pgfmathdeclarefunction{gauss}{3}{%
  \pgfmathparse{1/(#3*sqrt(2*pi))*exp(-((#1-#2)^2)/(2*#3^2))}%
}
\pgfmathdeclarefunction{sum_gauss}{5}{%
  \pgfmathparse{1/(3*#3*sqrt(2*pi))*exp(-((#1-#2)^2)/(2*#3^2)) + 2/(3*#5*sqrt(2*pi))*exp(-((#1-#4)^2)/(2*#5^2))}%
}

\newcommand{\kl}[2]{D_{\mathrm{KL}} \left[ \left. \left. #1 \right|\right| #2 \right] }
\DeclareMathOperator*{\argmax}{arg\,max}

\definecolor{blue-violet}{rgb}{0.54, 0.17, 0.89}
\definecolor{bluepigment}{rgb}{0.2, 0.2, 0.6}
\makeatletter
\newcommand*\bigcdot{\mathpalette\bigcdot@{.5}}
\newcommand*\bigcdot@[2]{\mathbin{\vcenter{\hbox{\scalebox{#2}{$\m@th#1\bullet$}}}}}
\makeatother
\def\delequal{\mathrel{\ensurestackMath{\stackon[1pt]{=}{\scriptstyle\Delta}}}}
\definecolor{Yellow}{RGB}{211, 176, 15}
\definecolor{Green}{rgb}{0.01, 0.75, 0.24}
\colorlet{Red}{red!50!black}
\colorlet{Blue}{blue!50!black}
\definecolor{Violet}{rgb}{0.56,0.14,0.56}
\newcommand{\indep}{\perp \! \! \! \perp}

\jmlrheading{1}{2020}{1-48}{4/00}{10/00}{meila00a}{Théophile Champion and Marek Grze\'s and Howard Bowman}

\ShortHeadings{Active Inference and Variational Message Passing}{Champion et al.}
\firstpageno{1}

\begin{document}

\title{Realising Active Inference in Variational Message Passing: the Outcome-blind Certainty Seeker}

\author{\name Théophile Champion \email TMAC3@KENT.AC.UK \\
       \addr University of Kent, School of Computing\\
       Canterbury CT2 7NZ, United Kingdom
       \AND
       \name Marek Grze\'s \email M.GRZES@KENT.AC.UK \\
       \addr University of Kent, School of Computing\\
       Canterbury CT2 7NZ, United Kingdom
       \AND
       \name Howard Bowman \email H.BOWMAN@KENT.AC.UK \\
       \addr University of Birmingham, School of Psychology,\\
       Birmingham B15 2TT, United Kingdom\\
       University of Kent, School of Computing\\
       Canterbury CT2 7NZ, United Kingdom
       }
       
\editor{\textbf{TO BE FILLED}}

\thispagestyle{plain}
\maketitle

\begin{abstract}
Active inference is a state-of-the-art framework in neuroscience that offers a unified theory of brain function. It is also proposed as a framework for planning in AI. Unfortunately, the complex mathematics required to create new models --- can impede application of active inference in neuroscience and AI research. This paper addresses this problem by providing a complete mathematical treatment of the active inference framework --- in discrete time and state spaces --- and the derivation of the update equations for any new model. We leverage the theoretical connection between active inference and variational message passing as describe by John Winn and Christopher M. Bishop in 2005. Since, variational message passing is a well-defined methodology for deriving Bayesian belief update equations, this paper opens the door to advanced generative models for active inference. We show that using a fully factorized variational distribution simplifies the expected free energy --- that furnishes priors over policies --- so that agents seek unambiguous states. Finally, we consider future extensions that support deep tree searches for sequential policy optimisation --- based upon structure learning and belief propagation.
\end{abstract}

\begin{keywords}
Active Inference, Variational Message Passing, Free Energy Principle, Reinforcement Learning, Kullback Leibler Control
\end{keywords}

\section{Introduction}

The free energy principle aims to provide a unified theory of the brain based on Bayesian probability theory \citep{Friston2010,BUCKLEY201755}. It takes root in Helmholtz's argument that observations are produced by hidden causes that must be inferred --- and the predictive coding formulation which argues that inference and learning emerges from the reduction of the error between predicted and actual observations. Active inference extends predictive coding to consider generative models of actions \citep{FRISTON2016862,AI_TUTO}.

In brief, active inference is a probabilistic framework that describes how agents should act in their environment. It starts with the definition of a generative (probabilistic) model that encodes the agent's beliefs about its environment. However, active inference does not rely on one particular generative model, instead it refers to a class of generative models that consider the impact of their actions in their environment. Active inference also relies on learning and inference to estimate the most likely states of the world and values of the model parameters. However, the concept behind active inference does not dependent on a particular inference method, which means that both variational inference \citep{VI_TUTO} and Monte Carlo Markov chains \citep{DeepAIwithMCMC} can, in principle, be used.

Active inference has been successfully applied in neuroscience to explain a wide range of brain phenomena such as habit formation \citep{FRISTON2016862}, Bayesian surprise \citep{bayes_surprise}, curiosity \citep{curiosity}, and dopaminergic discharges \citep{dopamine}. Active inference is also a form of planning as inference \citep{PAI} consistent with Occam's Razor \citep{occam} and can be seen as a generalisation of reinforcement learning \citep{DDQN,lample2016playing} and Kullback Leibler control \citep{KL_CONTROL}. This framework has also been used to ground active vision \citep{active_vision,10.3389/frai.2020.509354,10.3389/fnbot.2021.642780,10.3389/fncom.2016.00056,10.1371/journal.pone.0190429} within a strong theoretical framework.

This paper focuses on active inference using variational (a.k.a approximate Bayesian) inference and highlights its connection to variational message passing \citep{VMP_TUTO}. This ubiquitous message passing algorithm builds on the variational inference literature by leveraging the structure of the generative model to split the update equations into messages. Those messages transmit information about the new observations and --- by summing those messages --- it is possible to compute the posterior distribution over the parameters. The decomposition of the updates into messages formalises the modularity of the method, while remaining biologically plausible \citep{BP_AI}. Indeed, a key question in machine learning and computational neuroscience is how to identify compositional models --- an issue that was identified early in the development of connectionism \citep{kar30708,FODOR19883}. The central requirement being that higher-order representations (whether syntactic, semantic, perceptual, etc) can be constructed by ``plugging together" lower order representations, in such a way that the meanings of lower-order representations do not change (e.g. the ``Jane" in ``Jane loves John" is the same ``Jane" as in ``John loves Jane"). It may be that the structural modularity provided by message passing implementations of Bayesian networks enable compositionality of representations. According to modern trends, we use the formalism of Forney factor graphs \citep{FFG_TUTO} to represent the updates as messages sent along the graph edges.

Forney factor graphs are graphical representations used to realise generative models. They comprise of two kinds of round nodes that represent the observations and the latent variables of the model. If the notion of observations can be understood as the data available to the model, the notion of latent variables is a bit more abstract. As an example, let us consider the MNIST dataset \citep{mnist} composed of images of hand written digits. In this example, the pixels are observations made by the model and latent variables could be any variables encoding the digit being represented, such as its orientation or size. The last type of nodes --- square nodes --- represent the dependency between observed and latent variables. In other words, how does the digit being represented generate the pixels?

The first goal of this paper is to provide the reader with a full intuition of the mathematics underlying active inference and variational message passing. Then, this paper shows how to derive the update equations for any new generative models. The hope is to facilitate the development of new models that could, for example, play Atari games or model new brain mechanisms. Finally, we use our new generative model to prove that the update equations of active inference can be understood as variational message passing. This formal proof complements previous work that frames active inference as belief propagation \citep{BP_AI} and enables us to create an automatic and modular implementation of active inference \citep{Simul_AI,DBLP:journals/ijar/CoxLV19}. This message passing formulation has particular consequences for the expected free energy, which is effectively reduced by the change, resulting in an agent that seeks certainty, without any concern for outcomes, whether preferred or not. We argue that the resulting behaviour may have similarities to repetitive actions (sometimes called stimming) that are common, for example, in autism \citep{GABRIELS2005}.

Section \ref{sec:problem} describes the problem used to present the (classic) model widely used in the active inference literature. Sections \ref{sec:vi} and \ref{sec:ffg} introduce variational inference and Forney factor graphs, respectively. Next, Section \ref{sec:ai} presents active inference as a decision theory based on the Bayesian view of probability, followed by Section \ref{sec:ai_vmp} that introduces the notion of variational message passing. Then, Section \ref{sec:vmp-connection} formulates active inference as variational message passing under a fully factorised approximate posterior (i.e. variational distribution), and explains the implications of this approximation for the expected free energy that underwrites policy selection. Before starting the next section, readers new to the active inference literature might want to read Appendix D, which uses Bayes theorem to present the simplest generative model sufficient for active inference. 

\section{Problem statement}\label{sec:problem}

Active inference crops up in many areas that require an agent to interact with its environment. Throughout this paper, the explanations will be based on an agent named Bob, whose goal is to solve the food problem presented in section \ref{ss:food_problem}. But before we investigate this problem, let us have a look at how to simulate the interaction between Bob and his environment.

\subsection{Simulating active inference}

Most living beings are able to sense their environment through sensory inputs, and process this sensory information to act in the world. For example, carnivorous flowers use tiny trigger hairs on their leaves to detect flies (sensing). When those hairs are stimulated, the ion concentrations in the leaves increase (processing) resulting in an electrical current that closes the leaf trapping the fly (acting). Similarly, humans gather sensory information through their five senses (sensing), process this information to understand their environment (processing), and finally, make use of this understanding to act with intelligence (acting).

Sensing, processing and acting correspond to the three steps of the Action-Perception cycle. This cycle conveniently casts active inference as an infinite loop \citep{simulate_ai}. Each iteration begins by sampling the environment to obtain an observation, which is provided to the agent. Then, the observation is used to perform inference (and learning) that produce a higher level of understanding, for example, an image might be mapped to a representation of the objects that it contains. And finally, this representation is exploited when acting to prepare your diner, drive your kids to school or solve your favourite maths problem.

\subsection{The food problem} \label{ss:food_problem}

\begin{figure}[H]
	\centering
	{\includegraphics[width=0.7\textwidth]{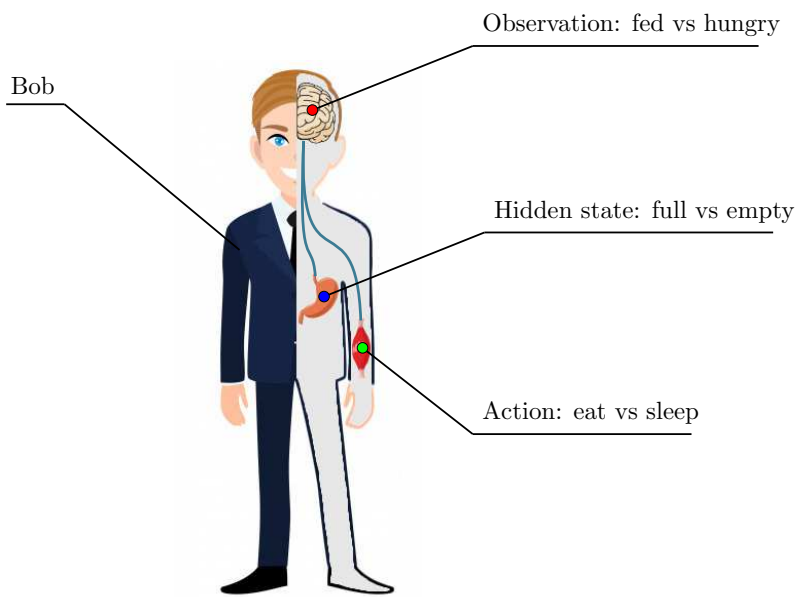}}
	\caption[The Food Problem]{This figure illustrates the food problem, where the goal of our agent --- Bob --- is to keep his stomach full. The first thing Bob needs to achieve his goal is to guess the state of his stomach, which can either be full or empty. This guess is informed by the observations he makes, when feeling hungry (high level of ghrelin) or fed (low level of ghrelin). Finally, once Bob has reduced his uncertainty about his stomach state, he can engage in exploitative behaviour by taking action in his environment, such as sleeping or eating.}
	\label{fig:food}
\end{figure}

Speaking of which, this section is concerned with the food problem initially proposed by Oleg \cite{oleg}. This problem concerns an agent, named Bob, striving to survive. To produce the energy needed by his body, Bob needs to ingest nutriments. During periods of starvation, Bob's stomach produces an hormone called ghrelin. This hormone travels to the brain through the blood and reaches a part of the brain, named the hippocampus. This area has been shown to monitor the level of ghrelin in the blood \citep{hippocampus_and_ghrelin}. At the moment ghrelin reaches the hippocampus, Bob's brain can estimate the content of his stomach. This information can then be exploited to choose between eating and sleeping. However, the best action depends on the outcomes that Bob wants to witness in the future. This paper assumes that mother nature has kindly set Bob's preferences to be biased towards the sensation of feeling fed (i.e. Bob enjoys observing low levels of ghrelin in his blood), which is arguably a favourable traits under a Darwinism view of evolution. Figure \ref{fig:food} summarises the food problem.

\section{Variational Inference}\label{sec:vi}

In Bayesian statistics, one assumes a prior distribution over latent (a.k.a hidden) variables that represent the process generating the data. When collecting more data, new observations bring information, allowing us to update our prior knowledge. The process of computing the most likely values of the hidden variables is called inference. A simple inference method is to use Bayes theorem to obtain the posterior probability distribution over the latent variable(s) of the model:
\begin{align*}
\underbrace{P(S|O)}_{\text{posterior}} = \frac{\overbrace{P(O|S)}^{\text{likelihood}}\overbrace{P(S)}^{\text{prior}}}{\underbrace{P(O)}_{\text{evidence}}} = \frac{P(O|S)P(S)}{\sum_S P(O|S)P(S)}.
\end{align*}

Since Bayes theorem is a corollary of the product rule of probability and no approximation is needed, it belongs to the field of exact inference. However, the computation of the evidence requires the marginalisation over all hidden variables, which makes it intractable for all but the simplest models.

To address this intractability, one can turn to approximate or sampling based methods. Variational inference belongs to the former and relies on an assumption of independence. As will be explained in Section \ref{ssec:vfed}, the idea behind variational inference is to use a distribution $Q(S)$ to approximate the true posterior $P(S|O)$. This can be accomplished by minimising the Kullback-Leibler (KL) divergence between some approximate and the true posterior: 
$$\kl{{\color{blue!50!black}Q(S)}}{{\color{red!90!black}P(S|O)}}.$$

Minimising this KL divergence is impossible because the true posterior $P(S|O)$ is unknown. Fortunately however, it is equivalent to minimising the variational free energy $\bm{F}$, known in machine learning as the negative evidence lower bound (ELBO). The variational free energy is defined as the Kullback-Leibler divergence between the variational distribution $Q(S)$ and the generative model $P(O,S)$:
\begin{align*}
\bm{F} &= \kl{{\color{blue!50!black} Q(S)}}{P(O,S)} = - ELBO\\
&= \kl{{\color{blue!50!black} Q(S)}}{{\color{red!90!black}P(S|O)}} + \ln P(O).
\end{align*}

The variational distribution $Q(S)$ is used to approximate the true posterior $P(S|O)$. In addition to the introduction of this approximate posterior, the mean-field approximation makes the computation tractable by assuming that all latent variables are independent:
\begin{align*}
{\color{blue!50!black}Q(S)} = \prod_i Q_i(S_i),
\end{align*}
where $Q_i(S_i)$ is the distribution over the i-th hidden state of the model and $Q(S)$ is the joint distribution over all latent variables. This assumption of independence constrains the expressiveness of the variational distribution, but allows the derivation of update equations, which can be evaluated efficiently.

At this point, an analogy might be useful to furnish an intuitive understanding of variational inference. Imagine you drop some coffee on a table, producing a stain with a complex shape. To compute the area of the stain, it might be useful to first assume an elliptic shape for the stain. However, since the stain is not actually elliptic, the solution will only be an approximation. In this analogy, the stain is the true posterior, and the ellipse is the approximate posterior.

\begin{figure}[H]
	\centering
	{\includegraphics[width=0.7\textwidth]{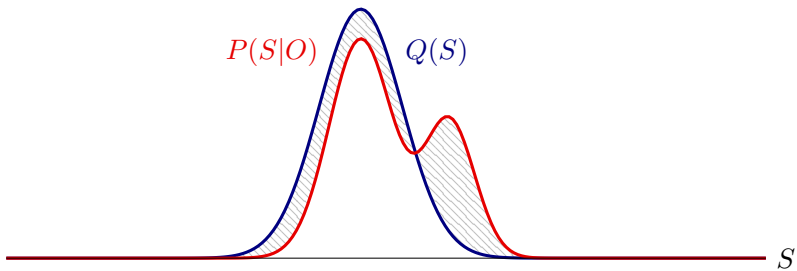}}
    \caption[Variational Inference - Approximate Posterior]{This figure illustrates the kind of result obtained using variational inference. The true posterior drawn in red has a complex shape and is approximated by the variational distribution drawn in blue. The grey area depicts the error made when using the variational distribution to approximate the true posterior.}
    \label{fig:approx-inf-0000}
\end{figure}

This analogy should help with the understanding of Figure \ref{fig:approx-inf-0000} that illustrates the kind of results obtained by variational methods. As will be demonstrated in Section \ref{ss:viu}, it is possible to prove \citep{VI_TUTO} that  minimising the variational free energy $\bm{F}$ with respect to $Q_k(S_k)$ can be performed by iterating one of the following update equations:
\begin{align} \label{eq:vi-optimal-update}
&\ln Q_k(S_k) \leftarrow \ln Q_k^*(S_k) = \langle \ln P(O,S) \rangle_{\sim Q_k}\\
\Leftrightarrow\quad \quad \quad &Q_k(S_k) \leftarrow Q_k^*(S_k) = \frac{1}{Z} \exp \langle \ln P(O,S) \rangle_{\sim Q_k},\nonumber
\end{align}
where $Q_k^*(S_k)$ is the optimal posterior, $Z$ is a normalisation constant and $\langle \bigcdot \rangle_{\sim Q_k}$ is the expectation over all factors but $Q_k$. Importantly, it is the coupling of the above update equations (i.e. one update per hidden variable $S_k$) that justifies the iteration of the updates until convergence to the free energy minimum.

\section{Forney Factor Graphs}\label{sec:ffg}

\begin{figure}[H]
	\centering
	{\includegraphics[width=0.5\textwidth]{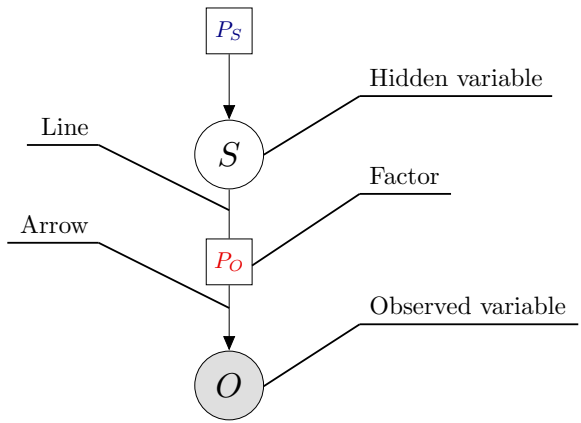}}
	\caption[Forney Factor Graph]{This figure illustrates the Forney factor graph corresponding to the following generative model: $P(O,S) = {\color{red!90!black}P_{O}}(O|S){\color{blue!50!black}P_{S}}(S)$. The hidden state is represented by a white circle with the variable's name at the centre, and the observed variable is depicted similarly but with a grey background. The factors of the generative model are represented by squares with a white background and the factor's name at the centre. Finally, arrows connect the factors to their target variable and lines link each factor to its predictor variables.}
    \label{fig:type-elements-ffg}
\end{figure}

Typically, generative models are represented graphically using a graphical model \citep{koller2009probabilistic} or Forney factor graph \citep{FFG_TUTO}. This section focuses on the latter representation introduced by David Forney in 2001, which uses three kinds of nodes. The nodes representing hidden and observed variables are depicted by white and grey circles, respectively. And factors are represented using white squares, which are linked to variable nodes by arrows or lines. Arrows are used to connect factors to their target variable, while lines link factors to their predictors. Figure \ref{fig:type-elements-ffg} shows an example of a Forney factor graph corresponding to the following generative model:
\begin{align} \label{eq:general-generative-model}
P(O,S) = {\color{red!90!black}P_{O}}(O|S){\color{blue!50!black}P_{S}}(S).
\end{align}

Generally, factor graphs only describe the model's structure --- in terms of the variables and their dependencies --- but not the individual factors. For example, the definitions of ${\color{red!90!black}P_{O}}$ and ${\color{blue!50!black}P_{S}}$ are not given by Figure \ref{fig:type-elements-ffg}, and additional information is required, e.g. ${\color{blue!50!black}P_{S}}(S) = \mathcal{N}(S;\mu, \sigma)$ specifies ${\color{blue!50!black}P_{S}}$ as a Gaussian distribution.

Initially, variables could only connect to a limited number of factors. However, a special kind of factor, called an equality node, dissolves this limitation. Purists tend to represent all equality nodes, while others make them implicit by allowing the variables to connect to an arbitrary number of factors. For sake of clarity, this paper keeps equality nodes implicit.

Finally, factors --- along with hidden and observed variables --- are sometimes called constraint, state and symbol, respectively. As explained by \citet{BP_and_DC}, those two terminologies refer to two views on Forney factor graphs, where factors encode probabilities and constraints encode costs. Infinite costs represent hard constraints, while finite costs encode soft constraints. Here, hard constraints define which configurations of the state space are forbidden (i.e. has a probability of zero) and soft constraints encode preferences over the state configurations (i.e. the higher the cost the smaller the state probability). This reveals an interesting link between Bayesian statistics and symbolic artificial intelligence, and prompts the question of whether Bayesian statistics can be regarded as a generalisation of symbolic artificial intelligence. For example, one could start by framing the problem of constraint satisfaction, as an inference process on a Forney factor graph that encodes the problem constraints.

\section{Active Inference}\label{sec:ai}

So far, we have discussed variational inference and Forney factor graphs. We now present the intuition behind the various equations that comprise the active inference framework. We will be working with the food problem that was introduced in Section \ref{sec:problem}.

\subsection{Generative model} \label{ssec:GM}

We begin by presenting the generative model introduced by \citet{AI_DISCRET}. Instead of presenting the full generative model at once, the next subsections build this model progressively. This should help the reader to understand both the model and its corresponding Forney factor graph.

\subsubsection{The D vector} \label{ssec:D-matrix}

As we shall see shortly, the full generative model represents the world as a sequence of hidden states, and those states generate the observations made by the agent. For the sake of organisation, those states are arranged chronologically using the index $\tau$ that runs from the initial state ($S_0$) to the state of the last time step ($S_T$). This section focuses on the initial state, whose distribution is a categorical, defined as follows:
\begin{align}\label{eq:PS0-def}
P_{S_0}(S_0|\bm{D}) = \text{Cat}(S_0; \bm{D}),
\end{align}
where $\bm{D}$ is a vector containing the parameters of the categorical distribution. In addition to the categorical distribution, the model assumes a Dirichlet prior over the parameters $\bm{D}$, leading to:
\begin{align}\label{eq:PD-def}
P_{D}(\bm{D}) = \text{Dir}(\bm{D};d).
\end{align}

\begin{figure}[H]
	\centering
	{\includegraphics[width=0.7\textwidth]{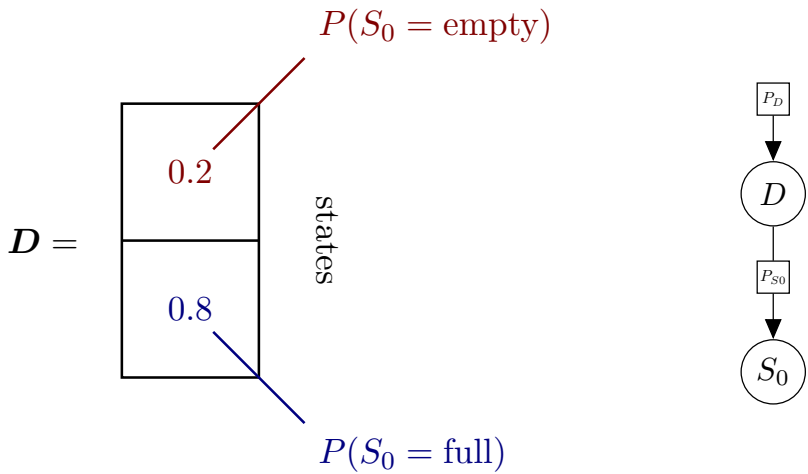}}
  \caption[Generative model - D matrix]{This figure illustrates the vector $\bm{D}$ that defines Bob's beliefs about the initial hidden state, and the Forney factor graph corresponding to (\ref{eq:PS0-def}) and (\ref{eq:PD-def}). Since the probability of $S_0$ being full is higher than the probability of it being empty, Bob thinks that at the beginning of each trial, his stomach is more likely to be full than empty.}
   \label{fig:D-matrix}
\end{figure}

In this context, the parameters $d$ of the Dirichlet distribution are called hyperparameters, because they control the distribution of the parameters $\bm{D}$. Figure \ref{fig:D-matrix} summarises this part of the model by presenting an example of the vector $\bm{D}$, and the Forney factor graph corresponding to the two distributions constituting Bob's generative model.

\subsubsection{The A matrix} \label{ssec:A-matrix}

\begin{figure}[H]
	\centering
	{\includegraphics[width=0.8\textwidth]{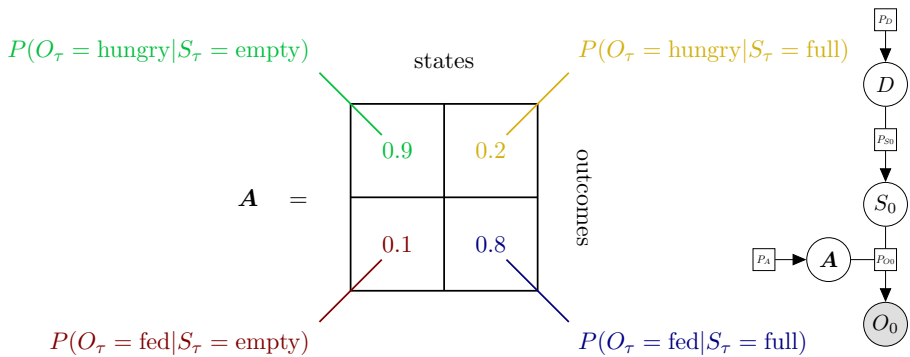}}
  \caption{This figure illustrates the matrix $\bm{A}$ that defines how the hidden states generate the observations. In our example with Bob, this matrix defines the probability of Bob feeling hungry or fed while his stomach is full or empty. Furthermore, the new version of the generative model is shown on the right.}
   \label{fig:A-matrix}
\end{figure}

We have already mentioned that the probability of an observation (a.k.a outcome), such as feeling hungry, depends on the value of the hidden state, i.e. whether Bob' stomach is full or empty. This dependency is represented by a conditional distribution, such that the likelihood of an observation --- given a particular value of the hidden states --- is defined by a categorical distribution, as follows:
$$P_{O_\tau}(O_\tau | S_\tau = j, \bm{A}) = \text{Cat}(O_{\tau};\bm{A}_{\bigcdot j}),$$
where the j-th column of $\bm{A}$, denoted $\bm{A}_{\bigcdot j}$, contains the parameters of the categorical distribution encoding the probability of the outcomes given that $S_\tau = j$. Additionally, we can re-write the above equation more concisely by letting $S_\tau$ be a one hot vector, whose j-th element is equal to one, such that:
$$P_{O_\tau}(O_\tau | S_\tau, \bm{A}) = \text{Cat}(O_{\tau};\bm{A}S_\tau),$$
where because $S_\tau$ is a one hot vector, the multiplication of $\bm{A}$ and $S_\tau$ selects the j-th column of $\bm{A}$. Similarly to the treatment of the vector $\bm{D}$, a prior over the columns of $\bm{A}$ is used. To ensure the conjugacy between the distributions of the model, a Dirichlet prior is used for each column. The probability of the overall matrix is then given by the following product of Dirichlet:
$$P_A(\bm{A}) = \prod_i \text{Dir}(\bm{A}_{\bigcdot i};a_{\bigcdot i}),$$
where $a$ is a matrix containing the parameters of the Dirichlet distributions, i.e., each column of $a$ contains the parameters of one Dirichlet distribution. Note that because each column of the matrix $\bm{A}$ is a categorical distribution, then the conjugate prior of each column is a Dirichlet distribution. Assuming independence of the columns of $\bm{A}$, the conjugate prior of the entire matrix $\bm{A}$ is a product of Dirichlet distributions. Importantly, the prior over $\bm{A}$ is not a Dirichlet distribution whose parameters are obtained by concatenation of the columns of $\bm{A}$. Indeed, if we sample from such a (concatenated) prior, then the elements of the entire matrix will sum up to one but the columns would not. This is problematic because each column of $\bm{A}$ is supposed to be a categorical distribution that sum up to one. We conclude this section with Figure \ref{fig:A-matrix} that illustrates the likely matrix $\bm{A}$, along with the resulting version of the generative model for Bob's problem.

\subsubsection{The B matrices} \label{ssec:B-matrices}

Now that the reader is familiar with the definition of the likelihood matrix $\bm{A}$, we focus on the temporal transitions between any pair of successive states. Those transitions are modelled similarly to the matrix $\bm{A}$ that concerns the generation of observations from hidden states. However here, we are concerned with the transition matrices that maps from states at one point on time to the next. Crucially, there are as many of these matrices as the number of allowable actions on the state in question. This follows from the idea that each action has the potential to modify Bob's stomach differently: for example, eating is more likely to change Bob's stomach from empty to full than sleeping. Accordingly, the transition between two consecutive hidden states is defined by a set of matrices, called the transition or $\bm{B}$ matrices, such that:
\begin{align}\label{eq:PStau1-def}
P_{S_{\tau + 1}}(S_{\tau + 1} | S_\tau = i, \pi = j, \bm{B}) &= \text{Cat}(S_{\tau + 1}; \bm{B}[U_\tau^j]_{\bigcdot i}) \nonumber\\
&\delequal \text{Cat}(S_{\tau + 1}; \bm{B}[U]_{\bigcdot i}),
\end{align}
where $\delequal$ means equal by definition, $U \delequal U_\tau^j$ is the action predicted at time step $\tau$ by the j-th policy, and $\bm{B}[U]$ is the matrix corresponding to the action $U$. Furthermore, active inference defines policies as action sequences (cf. next section). By replacing the index $i$ by a one hot vector as in the previous section, Equation \ref{eq:PStau1-def} can be re-written as:
$$P_{S_{\tau + 1}}(S_{\tau + 1} | S_\tau, \pi, \bm{B}) = \text{Cat}(S_{\tau+1};\bm{B}[U]S_\tau).$$

\begin{figure}[H]
	\centering
	{\includegraphics[width=1\textwidth]{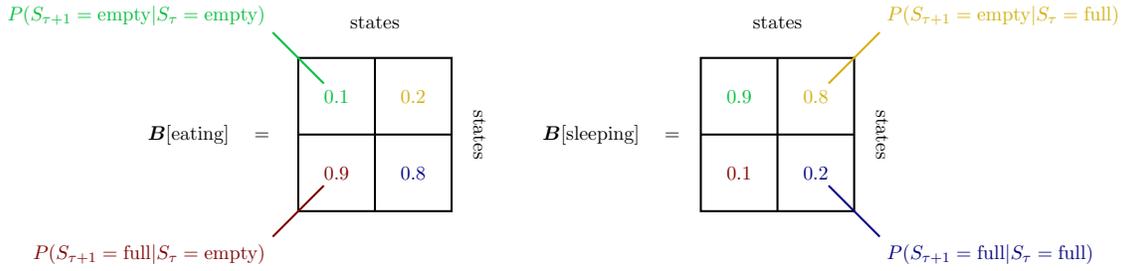}}
  \caption{This figure illustrates the matrices $\bm{B}$ that define the transition between any two consecutive hidden states. In the context of the food problem, those matrices encode the probability of transitioning from a full or empty stomach at time $\tau$ to a full or empty stomach at time $\tau + 1$.}
   \label{fig:B-matrices}
\end{figure}

\begin{figure}[H]
	\centering
	{\includegraphics[width=0.4\textwidth]{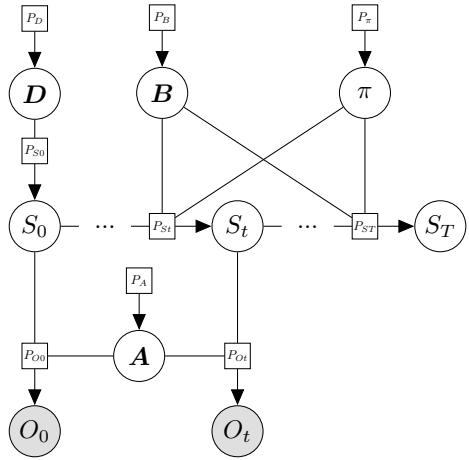}}
  \caption{This figure shows the next version of the generative model, where the transition between hidden states is specified by a set of $\bm{B}$ matrices and the policies $\pi$. At this point, it should be emphasized that the generation of outcomes through the matrix $\bm{A}$ stops after the current time step $t$. This follows naturally from the idea that we cannot observe future outcomes. Finally, the factor $P_\pi$ has not been defined yet: it will be the subject of the next section.}
   \label{fig:ffg-B-matrices}
\end{figure}

A Dirichlet prior is assumed for each column of the transition matrices $\bm{B}$, leading to the following prior:
$$P_B(\bm{B}) = \prod_{i,j} \text{Dir}(\bm{B}[i]_{\bigcdot j};b[i]_{\bigcdot j}),$$
where $b$ are the parameters of the Dirichlet distributions, $i$ and $j$ iterate over all possible actions and states, respectively. Finally, Figures \ref{fig:B-matrices} and \ref{fig:ffg-B-matrices} conclude this subsection by illustrating the matrices $\bm{B}$, and the updated version of the generative model.

\subsubsection{The prior over policies} \label{ssec:prior-policies}

We now consider the prior over the policy that was left undefined in Figure \ref{fig:ffg-B-matrices}. But what do we exactly mean by policies? In active inference, a policy is a sequence of actions over time, i.e. $\{U_t, ..., U_{T - 1}\}$. As a consequence, even if the agent expects the environment to be in the same state at two different time steps, picking two different actions at those time steps is still possible. Therefore, an active inference agent can perform an epistemic action as long as there is some uncertainty to be reduced and then switch to exploitative behaviours. Note that this definition of policy is in opposition to most of the model-free reinforcement learning literature, where a policy is a mapping from states to actions. In particular, states in the context of model-free reinforcement learning are observed and therefore are closer to the notion of observations in active inference. Technically, active inference takes us out of the world of fixed state-action policies (where the same action is taken from each state) into the world of sequential policy optimisation, where different actions can be taken from the same state --- crucially, in a way that depends upon (Bayesian) beliefs about hidden states.

The last ingredient required to obtain the prior over the policies is a notion of policy quality. In active inference, good policies are the ones that minimise the expected free energy; that is, the free energy expected in the future, which is defined as follows:
\begin{align}\label{eq:1}
\bm{G}(\pi) \approx \sum_{\tau=t + 1}^T \Bigg[ \underbrace{D_{\mathrm{KL}}[\overbrace{Q(O_\tau|\pi)}^{\text{expected outcomes}}||\overbrace{P(O_\tau)}^{\text{prior preferences}}]}_{\text{risk}}\,\, +\,\, \underbrace{\mathbb{E}_{Q(S_\tau|\pi)}[\text{H}[P(O_\tau | S_\tau)]]}_{\text{ambiguity}}\Bigg],
\end{align}
where $\text{H}[\cdot]$ is the Shannon entropy, $\bm{G}$ is a vector containing as many elements as the number of policies, and the i-th element of $\bm{G}$ represents the quality of the i-th policy. The reader interested in the derivation of the expected free energy is referred to Appendix C. We should mention here that $Q(O_\tau|\pi)$ and $Q(S_\tau|\pi)$ are computed based on the result of the inference process of the previous action-perception cycle. Therefore, $\bm{G}$ can be regarded as a model parameter and is not represented as a random variable in the Forney factor graph. The definition and justification of the expected free energy are provided in Appendix C and a recent paper by \citet{millidge2020expected}. Also, the expected free energy arises naturally in mathematical treatments of the free energy principle, when considering self-organisation at non-equilibrium steady-state \citep{friston2019free,doi:10.1098/rsta.2019.0159}. At this point, we should take a moment to understand the intuition behind the expected free energy.

Let us begin with the second term of Equation \ref{eq:1}. For each value of the hidden state, $P(O_\tau|S_\tau = i)$ is a categorical distribution whose parameters correspond to the i-th column of $\bm{A}$. This distribution defines the probability of future outcomes. Thus, the closer this distribution is to a uniform distribution, the more uncertain we are about future outcomes. This uncertainty is measured by the Shannon entropy, and the average of this quantity over all possible values of $S_\tau$ is called the ambiguity. Therefore, the ambiguity quantifies the degree to which a particular observation disambiguates among its hidden or latent causes.

\begin{figure}[H]
	\centering
	{\includegraphics[width=0.4\textwidth]{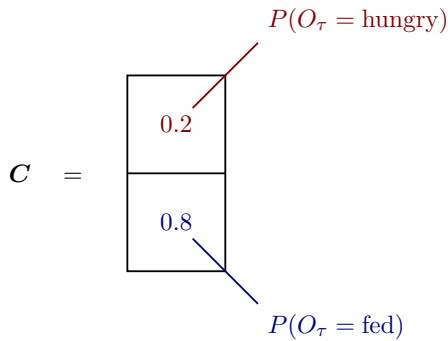}}
	\caption{This figure illustrates the vector $\bm{C}$ that defines Bob's prior preferences over future outcomes. This vector corresponds to the case where Bob prefers to feel fed rather than hungry, and the intensity of those preferences can be changed by tweaking the probabilities of the vector $\bm{C}$. For example, $\bm{C} = (0, 1)$ corresponds to an extreme preference towards feeling fed.}
   	\label{fig:C-matrix}
\end{figure}

Next, we need to encode Bob's preferences over future outcomes, which are called prior preferences. Formally, those preferences are defined as a categorical distribution whose parameters are stored in the vector $\bm{C}$. Figure \ref{fig:C-matrix} illustrates this vector. It should be noted that those preferences define the goodness of future outcomes, and we shall come back to this when discussing the link between active inference and reinforcement learning, cf. Appendix A.

\begin{figure}[H]
	\centering
	{\includegraphics[width=0.4\textwidth]{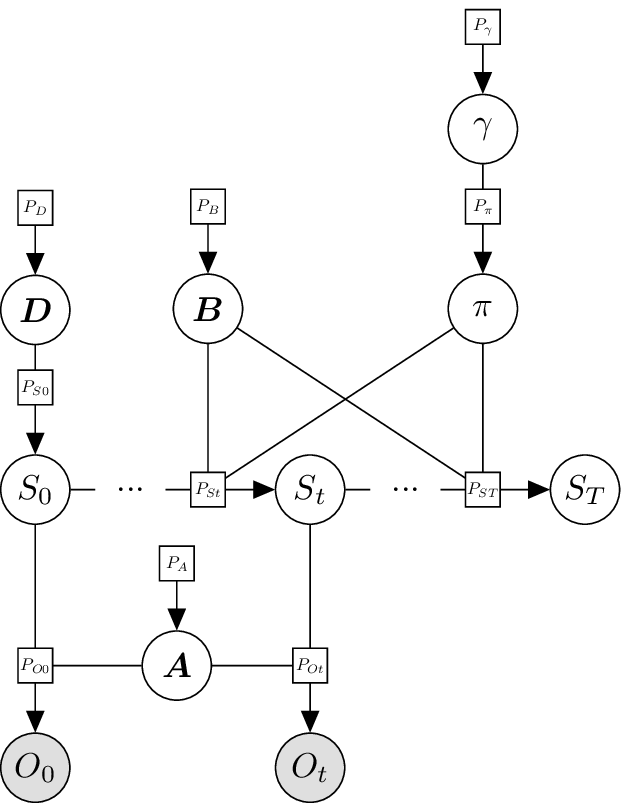}}
    \caption{This figure illustrates the Forney factor graph of the entire generative model of the sort presented by \citet{FRISTON2016862}. Section \ref{ssec:D-matrix} described how the probability of the initial states is defined by the vector $\bm{D}$, and as discussed in Section \ref{ssec:A-matrix}, the matrix $\bm{A}$ defines the probability of the observations given the hidden states. Section \ref{ssec:B-matrices} explained that the $\bm{B}$ matrices define the transition between any successive pair of hidden states. This transition depends on the action performed by the agent, i.e. on the policy $\pi$. Furthermore, the prior over the policies has been chosen in Section \ref{ssec:prior-policies}, such that policies minimising the expected free energy are more probable. Finally, we see in section \ref{sec:ppp} that the precision parameter $\gamma$ (which modulates the stochasticity of the agent behaviour) is distributed according to a gamma distribution.}
    \label{fig:full_GM}
\end{figure}

\begin{figure}[H]
	\centering
	{\includegraphics[width=0.7\textwidth]{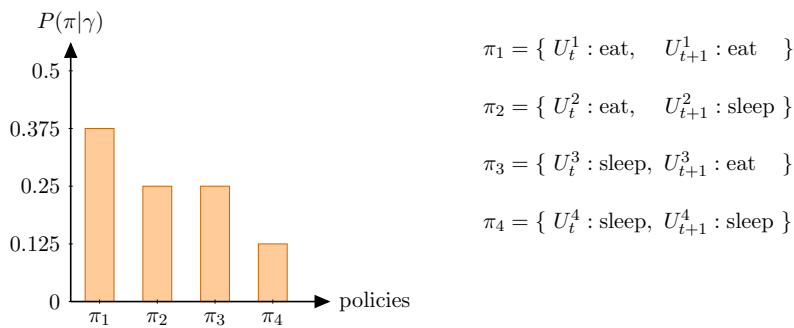}}
    \caption[Probability of policies]{A distribution over the policies that gives high probability to policies fulfilling Bob's preferences in the future. For example, the first policy where Bob is constantly eating has high probability, while the fourth policy where Bob is constantly sleeping has low probability. This is congruent with the notion that eating is more likely to make Bob feel fed than hungry, and similarly, sleeping is more likely to make Bob feel hungry than fed.}
    \label{fig:proba_agent_policies}
\end{figure}

To conclude, we need to consider the predicted or expected outcomes. One way to predict future outcomes would be to compute the marginal distribution over $O_\tau$ using for example the sum product algorithm \citep{910572}. However, this might be computationally expensive, so we will proceed with the following formula:
$$Q(O_\tau|\pi) = \sum_{i} P(O_\tau|S_\tau = i,A)Q(S_\tau = i|\pi) = \bm{A}\bm{s}_\tau^\pi,$$
where as will be discussed in Section \ref{section:VD}, $Q(S_\tau|\pi) \delequal \text{Cat}(S_\tau;\bm{s}_\tau^\pi)$. This equation can be understood as a form of marginalization, where the approximate posterior $Q(S_\tau|\pi)$ is our most informed belief about the hidden states. Finally, the KL divergence between the expected outcomes and the prior preference is called risk (cf. Appendix A for additional details). The risk part of expected free energy is simply the divergence between the expected outcomes and the preferred outcomes. It is this part of expected free energy that underwrites policies that lead to preferred outcomes under uncertainty. Minimising expected free energy therefore minimises risk (i.e., the divergence between anticipated and preferred outcomes) and ambiguity (i.e., the conditional uncertainty about outcomes, given the causes). The resulting prior over the policies is defined as:
$$P_\pi(\pi|\gamma) = \sigma(-\gamma \bm{G}),$$
where $\sigma(\cdot)$ is the softmax function, $\bm{G}$ is the expected free energy, $\gamma$ determines the sensitivity of policy selection to the expected free energy of each policy, and the negative sign gives high probability to policies minimising expected free energy. Importantly, the prior over policies is an empirical prior because the expected free energy depends on the observations, which means that it must be re-evaluated each time a new observation is made by the agent. In other words, the prior over the policies is a Boltzmann distribution with $\gamma$ being the inverse temperature. Taking this view, small values for $\gamma$ means a high temperature and less precise prior beliefs about which policy should --- or is --- being pursued. Figure \ref{fig:proba_agent_policies} shows an example of this distribution and Figure \ref{fig:full_GM} illustrates the current generative model.

\subsubsection{The prior over the precision parameter}\label{sec:ppp}

We now turn to the last part of the generative model, i.e. the prior over the precision parameter $\gamma$. Importantly, this precision parameter has been associated with the neuromodulator dopamine through what is called the ``precision hypothesis" \citep{dopamine}. This association of dopamine and the precision parameter claims to unify two perspectives on the role of dopamine. The first frames dopamine as an error signal on predicted reward \citep{Schultz1593} and uses the framework of TD-learning. The second, called the incentive salience hypothesis, frames dopamine as ``associating salience and attractiveness to visual, auditory, tactile, or olfactory stimuli" \citep{Berridge2007}. 

But, let us come back to the prior over the precision parameters $\gamma$. In neurobiological treatments, this prior usually takes the form of a gamma distribution with a rate parameter $\beta$ and a shape parameter fixed to one:
$$P_\gamma(\gamma) = \Gamma(\gamma;1,\beta).$$

\begin{figure}[H]
	\centering
	{\includegraphics[width=1\textwidth]{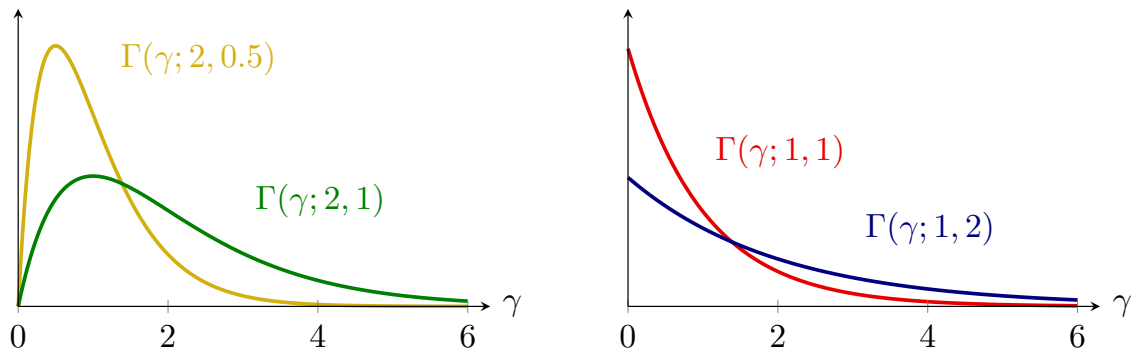}}
	\caption{This figure illustrates four gamma distributions where the values of the parameters have been changed. The graph on the right shows the kind of prior the model believes in by forcing the shape parameter to equal one.}
    \label{fig:gamma_prior}
\end{figure}

The graph on the right of Figure \ref{fig:gamma_prior} illustrates two variations of this prior for $\beta = 1$ and $\beta = 2$. Also, we should mention that a more flexible prior can be obtained by removing the constraint on the shape parameter \citep{EPISTEMIC_VALUE}, and the left hand side of Figure \ref{fig:gamma_prior} illustrates this extension. However, in most artificial intelligence applications (that are not concerned with biological implementation or dopamine), $\gamma$ is usually assumed to be one. Mainly, this design choice is made for the sake of simplicity, even if in practice forcing $\gamma$ to be one reduces the model flexibility, i.e. $\gamma$ can no longer be learnt.

\newpage

\begin{table}[H]
\centering
\begin{tabular}{cl}
\hline 
\\[-0.25cm]
Notation & Meaning \\
\\[-0.25cm]
\hline
\hline
\\[-0.4cm]
$T$ & The time horizon \\
\\[-0.4cm]
\hline
\\[-0.4cm]
$t$ & The current time steps \\
\\[-0.4cm]
\hline
\\[-0.4cm]
$\tau$ & An iterator over time step \\
\\[-0.4cm]
\hline
\\[-0.4cm]
$O_{0:t}$ & The sequence of observations between time step 0 and t \\
\\[-0.4cm]
\hline
\\[-0.4cm]
$S_{0:T}$ & The sequence of hidden states between time step 0 and T \\
\\[-0.4cm]
\hline
\\[-0.4cm]
$\pi$ & The policies\\
\\[-0.4cm]
\hline
\\[-0.4cm]
$U_\tau^m \delequal U$ & The action or control state predicted by the m-th policy at time step $\tau$\\
\\[-0.4cm]
\hline
\\[-0.4cm]
\multirow{2}{*}{$\bm{A}$}
& The matrix defining the likelihood mapping from the hidden states to the\\
& observations\\
\\[-0.4cm]
\hline
\\[-0.4cm]
$\bm{A}_{\bigcdot i}$ & The i-th column of the matrix $\bm{A}$\\
\\[-0.4cm]
\hline
\\[-0.4cm]
\multirow{2}{*}{$\bm{B}$}
& The set of transition matrices defining the mappings between any two consecutive\\
& hidden states\\
\\[-0.4cm]
\hline
\\[-0.4cm]
$\bm{B}[U]_{\bigcdot i}$ & The i-th column of the transition matrix $\bm{B}[U]$ corresponding to action $U$\\
\\[-0.4cm]
\hline
\\[-0.4cm]
$\bm{D}$ & The prior over the initial hidden states\\
\\[-0.4cm]
\hline
\\[-0.4cm]
$a$, $b$, $d$ & The parameters of the prior over $\bm{A}$, $\bm{B}$ and $\bm{D}$\\
\\[-0.4cm]
\hline
\\[-0.4cm]
$a_{\bigcdot i}$ & The i-th column of the matrix $a$\\
\\[-0.4cm]
\hline
\\[-0.4cm]
$b[U]_{\bigcdot i}$ & The i-th column of the matrix $b[U]$ corresponding to action $U$\\
\\[-0.4cm]
\hline
\\[-0.4cm]
$\gamma$ & The precision parameter related to neuromodulators such as dopamine\\
\\[-0.4cm]
\hline
\\[-0.4cm]
$\sigma(x)$ & The softmax function\\
\\[-0.4cm]
\hline
\\[-0.4cm]
$\bm{G}$ & The expected free energy\\
\\[-0.4cm]
\hline
\\[-0.4cm]
$\Gamma (\gamma;\alpha, \beta)$ & Gamma distribution with shape and inverse scale parameters $\alpha$ and $\beta$\\
\\[-0.4cm]
\hline
\\[-0.4cm]
$\text{Cat}(S_0;\bm{D})$ & Categorical distribution over $S_0$ with parameter $\bm{D}$\\
\\[-0.4cm]
\hline
\\[-0.4cm]
$\text{Dir}(\bm{D};d)$ & Dirichlet distribution\\
\\[-0.4cm]
\hline
\end{tabular}
\caption{Generative Model notation}\label{tab:active_inf_notation}
\end{table}

\newpage

\subsubsection{The entire generative model}

Throughout this section, we have assembled incrementally the generative model usually used in active inference, whose Forney factor graph is represented in Figure \ref{fig:full_GM}. The last step is to write down the equations that constitute its formal definition: 

\begin{align}
P(O_{0:t}, S_{0:T}, \pi, \bm{A}, \bm{B}, \bm{D}, \gamma)\,\, =\,\,\,\, &P(\pi|\gamma) P(\gamma) P(\bm{A}) P(\bm{B}) P(S_0|\bm{D}) P(\bm{D})\nonumber\\
&\prod^{t}_{\tau = 0} P(O_\tau|S_\tau,\bm{A}) \prod^{T}_{\tau = 1} P(S_\tau|S_{\tau-1},\bm{B},\pi),\label{eq:gm}
\end{align}
where:
\begin{align*}
&P(\pi|\gamma) = \sigma(-\gamma \bm{G}) & & P(\gamma) = \Gamma (\gamma;1, \beta) {\color{white}\prod_i }\\
&P(\bm{A}) = \prod_i \text{Dir}(\bm{A}_{\bigcdot i};a_{\bigcdot i}) & &  P(\bm{B}) = \prod_{i,j} \text{Dir}(\bm{B}[i]_{\bigcdot j};b[i]_{\bigcdot j})\\
&P(S_0|\bm{D}) = \text{Cat}(S_0;\bm{D}) & & P(\bm{D}) = \text{Dir}(\bm{D};d){\color{white}\prod_i }\\
&P(O_\tau|S_\tau,\bm{A}) = \text{Cat}(O_\tau;\bm{A}S_\tau) & & P(S_{\tau}|S_{\tau - 1},\bm{B}, \pi) = \text{Cat}(S_{\tau};\bm{B}[U]S_{\tau - 1}){\color{white}\prod_i }
\end{align*}

Note that to keep the notation uncluttered, we have dropped the subscripts such that $P_{S_0}(S_0|\bm{D})$ becomes $P(S_0|\bm{D})$, $P_A(\bm{A})$ becomes $P(\bm{A})$ and so forth. Table \ref{tab:active_inf_notation} provides a complete description of the notation used to define the generative model.

\subsection{Variational Distribution}\label{section:VD}

We now turn to the definition of the variational distribution, which is used to approximate the true posterior during variational inference (a.k.a approximate Bayesian inference), i.e. $Q(x) \approx P(x|o)$ where $x$ and $o$ denote the hidden variables and the observations, respectively. Let us first recall that variational inference leverages independence between latent variables in what is known as a mean-field approximation. A structured approximation, often made in the active inference literature\footnote{An instance where this general assumption is not made can be found in \citep{ParrDec2019}.} to simplify computations is that all latent variables are independent except for the hidden states and the policy. This leads to the following variational distribution:
\vspace{-0.2cm}
\begin{align}\label{vd}
Q(S_{0:T}, \pi, \bm{A}, \bm{B}, \bm{D}, \gamma) = Q_\pi(\pi)Q_A(\bm{A})Q_B(\bm{B})Q_D(\bm{D})Q_\gamma(\gamma) \prod_{\tau=0}^{T} Q_{S\tau}(S_\tau|\pi),
\end{align}
where:
\begin{align*}
&Q_{S\tau}(S_\tau|\pi) = \text{Cat}(S_{\tau};\bm{s}_\tau^\pi) && Q_\pi(\pi) = \text{Cat}(\pi;\bm{\pi})\\
&Q_\gamma(\gamma) = \Gamma(\gamma;1,\bm{\beta}) && Q_D(\bm{D}) = \text{Dir}(\bm{D};\bm{d})\\
&Q_A(\bm{A}) = \prod_i \text{Dir}(\bm{A}_{\bigcdot i};\bm{a}_{\bigcdot i}) && Q_B(\bm{B}) = \prod_{i,j} \text{Dir}(\bm{B}[i]_{\bigcdot j};\bm{b[i]_{\bigcdot j}})
\end{align*}

Once again, for the sake of compactness, the subscript will be dropped, e.g. $Q_{S\tau}(S_\tau|\pi)$ will be replaced by $Q(S_\tau|\pi)$. Table \ref{tab:variational_distribution_notation} summarises the notation used to define this variational distribution. It is much easier to understand this distribution by comparing it to the definition of the generative model in Equation \ref{eq:gm}. Indeed, the distributions over $\bm{A}$, $\bm{B}$ and $\bm{D}$ remain Dirichlet distributions, and the distributions over $\gamma$ and $S_\tau$ remain gamma and categorical distributions, respectively. Only the distribution over $\pi$ changes from a Boltzmann to a categorical distribution. However, both the Boltzmann and the categorical are discrete distributions.
\begin{table}[H]
\centering
\begin{tabular}{cl}
\hline 
\\[-0.25cm]
Notation & Meaning \\
\\[-0.25cm]
\hline
\hline
\\[-0.4cm]
$\bm{s}_{\tau}^{\pi}$ & The parameters of the posterior over $S_{\tau}$ for each policy, i.e. a vector\\
\\[-0.4cm]
\hline
\\[-0.4cm]
$\bm{s}_{\tau}^{\,\bigcdot}$ & The parameters of the posterior over $S_{\tau}$ for all policies, i.e. a matrix\\
\\[-0.4cm]
\hline
\\[-0.4cm]
$\bm{\pi}$ & The parameters of the posterior over $\pi$, i.e. a vector\\
\\[-0.4cm]
\hline
\\[-0.4cm]
\multirow{2}{*}{$\bm{a}$, $\bm{b}$, $\bm{d}$}
& The parameters of the posterior over $\bm{A}$, $\bm{B}$ and $\bm{D}$, i.e. a matrix, \\
& a set of matrices and a vector, respectively\\
\\[-0.4cm]
\hline
\\[-0.4cm]
$\bm{\beta}$ & The (inverse temperature) parameter of the posterior over $\gamma$\\
\\[-0.4cm]
\hline
\end{tabular}
\caption{Variational distribution notation}\label{tab:variational_distribution_notation}
\end{table}

\subsection{Variational Free Energy}

Above, we have unpacked the generative model and variational distribution used in active inference. This section combines those two concepts to form the second cornerstone of the active inference framework, i.e. the variational free energy. Section \ref{ssec:vfed} will explain how the following equation can be derived from the Kullback-Leibler divergence between the variational distribution and the true posterior. However, this section explains the intuition behind the variational free energy, which is defined as follows:
\begin{align}
\bm{F} &= \mathbb{E}_{Q}[\ln Q(S_{0:T}, \pi, \bm{A}, \bm{B}, \bm{D}, \gamma) - \ln P(O_{0:t},S_{0:T}, \pi, \bm{A}, \bm{B}, \bm{D}, \gamma)]\nonumber\\
&= \underbrace{\kl{Q(x)}{P(x|o)}}_{\text{relative entropy}}\,\,\, - \underbrace{\ln P(o)}_{\text{log evidence}}, \label{eq:2}
\end{align}
where $x = \{S_{0:T}, \pi, \bm{A}, \bm{B}, \bm{D}, \gamma\}$ refers to the model's hidden variables, and $o = \{O_{0:t}\}$ refers to the sequence of observations made by the agent. Equation \ref{eq:2} highlights some important properties of the variational free energy. Indeed, the relative entropy (a.k.a KL divergence) ensures that the variational distribution $Q(x)$ tends to get closer to the true posterior $P(x|o)$, as the free energy is reduced. Furthermore, it shows that the variational free energy is an upper bound on the negative log evidence, because the relative entropy cannot be negative. Also, if the variational distribution is equal to the true posterior, then the variational free energy is equal to the (-ve) log evidence. The variational free energy can also be re-arranged as:
\begin{align}
\bm{F} &= \underbrace{\kl{Q(x)}{P(x)}}_{\text{complexity}} - \underbrace{\mathbb{E}_{Q(x)}[\ln P(o|x)]}_{\text{accuracy}},\label{eq:3}
\end{align}

showing the trade-off between complexity and accuracy. The complexity penalises the divergence of the posterior $Q(x)$ from the prior $P(x)$. The accuracy scores how likely the observations are given the generative model and current belief of the hidden states. Interestingly, in opposition to the Akaike information criterion (AIC) and Bayesian information criterion (BIC), the complexity does not depend on the number of parameters. Consequently, a model with a lot of parameters, but that does not vary from the prior will have zero complexity, and a model with a small number of parameters that moves away a lot from the prior will have a large complexity. Taking this view, a model is complex whenever the knowledge encoded by the prior fails to explain the observed data accurately. In other words, complexity scores the degree of belief updating that moves posterior beliefs away from prior beliefs to provide an accurate account of any observations.

Comparison of the expression for expected free energy and variational free energy reveals an intimate relationship. One can see that the risk is the expected complexity, while ambiguity is expected inaccuracy. These expectations are under the posterior predictive beliefs about outcomes in the future under the policy in question. This is why $\bm{G}$ is called expected free energy.

\subsection{Update equations} \label{sec:ueq}

All the update equations presented below come from the minimisation of the variational free energy. This section presents the intuition behind those updates using the notations summarized in Table \ref{tab:message_passing_notation}. Let us start with the optimal updates of $\bm{A}$, $\bm{B}$ and $\bm{D}$ that are given by:
\begin{align}
Q^*(\bm{D}) &= \text{Dir}(\bm{D}; \bm{d}) \quad\quad\quad\quad\quad\,\,\,\, \text{ where } \quad \bm{d} = d + \bm{s}_0\\
Q^*(\bm{A}) &= \prod_i \text{Dir}(\bm{A}_{\bigcdot i}, \bm{a}_{\bigcdot i}) \quad\quad\quad\,\, \text{ where }\quad \bm{a} = a + \sum_{\tau = 0}^t \bm{o}_\tau \otimes \bm{s}_\tau\\
Q^*(\bm{B}) &= \prod_{u,i} \text{Dir}(\bm{B}[u]_{\bigcdot i}, \bm{b}[u]_{\bigcdot i}) \,\quad \text{ where } \quad \bm{b}[u] = b[u] + \sum_{(k,\tau) \in \Omega_u} \bm{s}^k_{\tau} \otimes \bm{s}^k_{\tau - 1} \bm{\pi}_k
\vspace{-0.5cm}
\end{align}

Looking at the above equations, these updates can be understood as counting the number of times an event appears. For example, the update of $\bm{A}$ counts the number of times a pair of states-observations have been observed. Taking this view, $a$ is the pseudo count of previously occurring states-observations pairs, and $\bm{o}_\tau \otimes \bm{s}_\tau$ takes into account the new observations. Similarly, the update of the $\bm{B}$ and $\bm{D}$ matrices, respectively count how many times the state transitions and initial states have been observed. Additionally, the updates of the hidden states are:
\begin{align}
Q^*(S_0|\pi) &= \sigma \Big(\bm{\bar{D}} \hspace{-1.67cm} &+\, I(0 \leq t)\,  \bm{o}_{0}\, \cdot\,  \bm{\bar{A}} &+\,\, \bm{s}^\pi_{1}\quad \cdot \bm{\bar{B}}[U_{0}^\pi] \Big){\color{white}\sum_i^T}\label{eq:6}\\
Q^*(S_\tau|\pi) &= \sigma \Big( \bm{\bar{B}}[U_{\tau - 1}^\pi] \bm{s}^\pi_{\tau - 1} \hspace{-1.6cm} &+\, I(\tau \leq t)\,  \bm{o}_{\tau} \,\cdot\,  \bm{\bar{A}} &+\,\, \bm{s}^\pi_{\tau + 1} \cdot \bm{\bar{B}}[U_{\tau}^\pi] \Big)\label{eq:5}\\
Q^*(S_T|\pi) &= \sigma \Big(\underbrace{\bm{\bar{B}}[U_{T - 1}^\pi] \bm{s}^\pi_{T - 1}}_{\text{past or prior}} \, \hspace{-1.6cm}&+ \underbrace{I(T \leq t)\, \bm{o}_{T} \cdot \bm{\bar{A}}}_{\text{likelihood}} &{\color{white}+ } \,\,\,\underbrace{{\color{white}\bm{\bar{B}}[U_{T - 1}^\pi]}\quad\,\,\,\,\,}_{\text{future}} \Big){\color{white}\sum^T}\label{eq:7}
\end{align}
where $t$ can be thought of as a global variable referring to the present time point, and $I(\bigcdot)$ is an indicator function that equals one if the condition is true and zero otherwise. A closer look at these updates reveals that the hidden states are updated by gathering information from the past, the future, and the likelihood mapping. In Equation \ref{eq:6}, the information from the past is replaced by some information from the prior over the initial state, and in Equation \ref{eq:7}, the information from the future disappears because we have reached the limits of the time horizon (i.e. $\tau == T$). Similarly, in Equations \ref{eq:5} and \ref{eq:7}, the indicator function ensures that there is no information from the likelihood mapping after the current time step $t$ because no observations are available. For additional information about the above updates, the reader is referred to Sections \ref{sec:msg_sum} and \ref{sec:msg_vs_up_eq} as well as Appendix G. Interestingly, \citet{Parr304782} proposed a model in which future observations are latent variables, and in this case, information will be sent along the edges connecting future states and future observations. Finally, the update of $\gamma$ and $\pi$ takes the following form:
\begin{align*}
&Q^*(\gamma) = \Gamma \Big(\gamma; 1, \beta + \bm{G} \cdot (\bm{\pi} - \bm{\pi}_0)\Big)\\
&Q^*(\pi) = \sigma \Big(-\frac{1}{\bm{\beta}}\bm{G} - \mathcal{F} \Big)
\end{align*}
where $\bm{\pi}_0 = \sigma(-\gamma \cdot \bm{G})$, $\sigma(\cdot)$ is the softmax function, and $\mathcal{F}$ is a vector whose $\pi$-th element is defined as:
\begin{align*}
\mathcal{F}_{\pi} = \bm{s}^\pi_{0} \cdot ( \ln \bm{s}^\pi_{0} - \bm{\bar{D}} ) + \sum_{\tau = 1}^T \bm{s}^\pi_{\tau} \cdot ( \ln \bm{s}^\pi_{\tau} - \bm{\bar{B}}[U] \bm{s}^\pi_{\tau - 1}) - \sum_{\tau = 0}^t \bm{o_\tau} \cdot \bm{\bar{A}} \bm{s}^\pi_{\tau}.
\end{align*}
Section \ref{sec:vmp-connection} will derive update equations similar to those above that can be decomposed as a sum of messages coming from the parent, children and co-parents of each node.
\begin{table}
\vspace{-0.75cm}
\begin{center}
\begin{tabular}{cl}
\hline
\\[-0.25cm]
Notation & Meaning \\
\\[-0.25cm]
\hline
\hline
\\[-0.4cm]
$A \otimes B = AB^T$, $A \cdot B = A^TB$ & outer and inner products\\
\\[-0.4cm]
\hline
\\[-0.4cm]
$\llbracket a, b \rrbracket$ & all the natural numbers between $a$ and $b$\\
\\[-0.4cm]
\hline
\\[-0.4cm]
\multirow{2}{*}{$\Omega_u = \Big\{(k, \tau): U^k_{\tau - 1} = u, \tau \in \llbracket 1, T \rrbracket\Big\}$}
& all $(k, \tau)$ such that the $k$-th policy predicts\\
& action $u$ at time $\tau - 1$\\
\\[-0.4cm]
\hline
\\[-0.4cm]
$\bm{s}_{\tau} = \bm{s}^{\,\bigcdot}_{\tau} \cdot \bm{\pi}$ & the expected state at time $\tau$\\
\\[-0.4cm]
\hline
\\[-0.4cm]
$\langle f(X) \rangle_{P_X} \delequal \mathbb{E}_{P_X}[f(X)]$ & is the expectation of $f(X)$ over $P_X$\\
\\[-0.4cm]
\hline
\\[-0.4cm]
\multirow{2}{*}{$\psi(x)$} & the digamma function used to compute \\
&analytical solutions, e.g. for $\langle \ln \bm{D}_i \rangle_{Q_D}$.\\
\\[-0.4cm]
\hline
\\[-0.4cm]
$\bm{\bar{D}_i} = \langle \ln \bm{D}_i \rangle_{Q_D} = \psi(\bm{d}_{i}) - \psi(\sum_i \bm{d}_{i})$ & the expected logarithm of $\bm{D}$\\
\\[-0.4cm]
\hline
\\[-0.4cm]
$\bm{\bar{A}}_{i j} = \langle \ln \bm{A}_{i j} \rangle_{Q_A} = \psi(\bm{a}_{ij}) - \psi(\sum_k \bm{a}_{kj})$ & the expected logarithm of $\bm{A}$\\
\\[-0.4cm]
\hline
\\[-0.4cm]
$\bm{\bar{B}}[u]_{i j} = \langle \ln \bm{B}[u]_{i j} \rangle_{Q_B} = \psi(\bm{b}[u]_{ij}) - \psi(\sum_k \bm{b}[u]_{kj})$ & the expected logarithm of $\bm{B}$\\
\\[-0.4cm]
\hline
\end{tabular}
\caption{Update equations notation}\label{tab:message_passing_notation}
\end{center}
\vspace{-0.75cm}
\end{table}

\subsection{Action selection}

This section focuses on the various strategies available to pick the next action(s) that the agent will then perform. In active inference, the action selection process is performed after iteration of the update equations. Indeed, according to the Action-Perception cycle presented in Section \ref{sec:problem}, the agent first minimises the variational free energy and then acts in its environment. The first strategy entails summing the posterior evidence for the policies predicting each action, and to execute the action with the highest sum of posterior evidence:
\begin{align*}
u_t^* = \argmax_{u} \sum_{m = 1}^{|\pi|} \delta_{u,U^m_t} Q(\pi = m),
\end{align*}
where $|\pi|$ is the number of policies, $U^m_t$ is the action predicted at the current time step by the policy $\pi$, and $\delta_{u,U^m_t}$ is an indicator function that equals one if $u = U^m_t$ and zero otherwise. Since the model knows the posterior over the policies (i.e. sequences of actions) another strategy is to simply sample an entire policy (e.g. a sequence of actions) without re-computing the posterior at each timestep, i.e. Bob selects a policy, closes his eyes and performs the sequence of actions entailed by that multi-step policy. In the case of single-step policies, this is equivalent to the first strategy. This leads to a trade-off between computational time and quality of the actions selected. Indeed, the more actions selected at once, the less computational time required, but the less informed those actions will be.

Another strategy used in planning is called a Monte Carlo tree search \citep{6145622}. The most well-known example of Monte Carlo tree search is probably the victory of AlphaGo against Lee Sedol --- the go world champion --- in 2016 \citep{Go}. Interestingly, this method has been used recently with an active inference agent \citep{DeepAIwithMCMC}. The simplest version of this algorithm starts with an empty tree, i.e. a single node representing the current state. Then, the root node is expanded such that the states that are reachable from the current state become its children. Those children are linked to the root node by edges representing the actions leading to those states. Afterwards, simulations of the environment are run to evaluate how good those new child states are. In the context of reinforcement learning, the goodness of the states corresponds to whether or not rewarding terminal states are reached during the simulations. Similarly, in the context of active inference, the expected free energy scores the goodness of outcomes. Finally, the reward or EFE is back-propagated upward in the tree. Iterating this four-steps process (i.e. selection, expansion, simulation and backpropagation) furnishes a posterior over the best action to perform next.

\section{Variational Message Passing}\label{sec:ai_vmp}

In the previous sections, our focus was on explaining the intuition behind active inference. The current section is more technical. We begin with the KL divergence between the variational distribution $Q(x)$ and the true posterior $P(x|o)$, which underwrites the minimisation of the variational free energy. Then, we derive two update equations well known from the Bayesian statistics community. The first explains how the approximate posterior can be computed using variational inference. And the second reveals that the optimal posterior can be thought of as a sum of messages. Finally, the message based equation is specialised for the class of exponential conjugate models that we use to describe the method of \citet{VMP_TUTO} as a five-step process. During this section, we will be using a few properties that are summarised in Appendix B.

\subsection{Justification of the Variational Free Energy} \label{ssec:vfed}

As mentioned in Section \ref{sec:vi}, the computation of the true posterior --- using Bayes theorem quickly becomes intractable as the number of hidden states increases. The variational free energy (VFE), or equivalently, the negative evidence lower bound (-ELBO), aims to solve this intractability problem by approximating the true posterior with another distribution: the variational distribution. To justify the use of the variational free energy, let us first note that the following expression can be obtained from the product rule:

\begin{equation}\label{eq:cond}
P(x|o) = \frac{P(o,x)}{P(o)}.
\end{equation}

Since the KL divergence measures the distance between two distributions, we can minimise the KL divergence between the variational distribution and the true posterior. And this will keep the variational distribution close to the true posterior. Starting with this KL divergence, and substituting Equation \ref{eq:cond} within it, we obtain:
\begin{align*}
\kl{Q(x)}{P(x|o)} &= \kl{Q(x)}{P(x,o)} + \mathbb{E}_{Q(x)}[\ln P(o)]\\
&= \underbrace{\kl{Q(x)}{P(x,o)}}_{\text{VFE = -ELBO}} + \underbrace{\ln P(o)}_{\text{log evidence}},
\end{align*}

where the expectation over the log evidence can be dropped due to the lack of a dependence of $\ln P(o)$ on $Q(x)$. Because the log evidence does not depend on the latent variables, it can be safely ignored during the minimisation process. In other words, minimising the variational free energy is equivalent to minimising the KL divergence between the variational distribution and the true posterior, and ensuring that the variational distribution is a good approximation of the true posterior.

\subsection{Variational Inference Updates} \label{ss:viu}

As we have just noted, variational methods rely on the minimisation of the variational free energy, or equivalently, the maximisation of an evidence lower bound. So, let us start with the former:
$$\kl{Q(x)}{P(o,x)} = \mathbb{E}_{Q(x)}[\ln Q(x) - \ln P(x,o)].$$

Using the mean-field assumption $Q(x) = \prod_i Q_i(x_i)$, the log property, and the linearity of expectation. The above equation can be rewritten as:
$$\kl{Q(x)}{P(o,x)} = \mathbb{E}_{Q(x)}[\ln Q_k(x_k)] + \mathbb{E}_{Q(x)}[\ln \prod_{j \neq k} Q_j(x_j)] - \mathbb{E}_{Q(x)}[\ln P(x,o)].$$

Note that $\ln Q_k(x_k)$ is a constant w.r.t all factors but $Q_k(x_k)$, and $\ln \prod_{j \neq k} Q_j(x_j)$ is a constant w.r.t $Q_k(x_k)$. Using the expectation of a constant, the above equation can be rewritten as:
$$\kl{Q(x)}{P(o,x)} = \mathbb{E}_{Q_k(x_k)}[\ln Q_k(x_k)] + \mathbb{E}_{\sim Q_k(x_k)}[\ln \prod_{j \neq k} Q_j(x_j)] - \mathbb{E}_{Q(x)}[\ln P(x,o)],$$
where $\mathbb{E}_{\sim Q_k(x_k)}[\cdot]$ is the expectation over all factors but $Q_k(x_k)$. If the goal is to minimise the free energy w.r.t $Q_k(x_k)$, the second term can be safely considered as a constant $C$. Also, using the factorisation of the variational distribution, the third term can be rewritten
as $\mathbb{E}_{Q_k(x_k)}[\mathbb{E}_{\sim Q_k(x_k)}[\ln P(x,o)]]$, leading to:
\begin{align*}
\kl{Q(x)}{P(o,x)} &= \mathbb{E}_{Q_k(x_k)}[\ln Q_k(x_k)] - \mathbb{E}_{Q_k(x_k)}[\mathbb{E}_{\sim Q_k(x_k)}[\ln P(x,o)]] + C\\
&= \mathbb{E}_{Q_k(x_k)}\Big[\ln Q_k(x_k) - \mathbb{E}_{\sim Q_k(x_k)}[\ln P(x,o)]\Big] + C\\
&\delequal \mathbb{E}_{Q_k(x_k)}\Big[\ln Q_k(x_k) - \ln Q_k^*(x_k)\Big] + C\\
&= \kl{Q_k(x_k)}{Q_k^*(x_k)} + C,
\end{align*}
where $\delequal$ means equal by definition, and $\ln Q_k^*(x_k) \delequal \mathbb{E}_{\sim Q_k(x_k)}[\ln P(x,o)]$. The KL divergence can not be negative which means that $Q_k(x_k) = Q_k^*(x_k)$ minimises the free energy, and for this reason $Q_k^*(x_k)$ is called the optimal posterior. 

\subsection{Variational Message Passing Updates}

\begin{figure}[H]
	\centering
	{\includegraphics[width=0.5\textwidth]{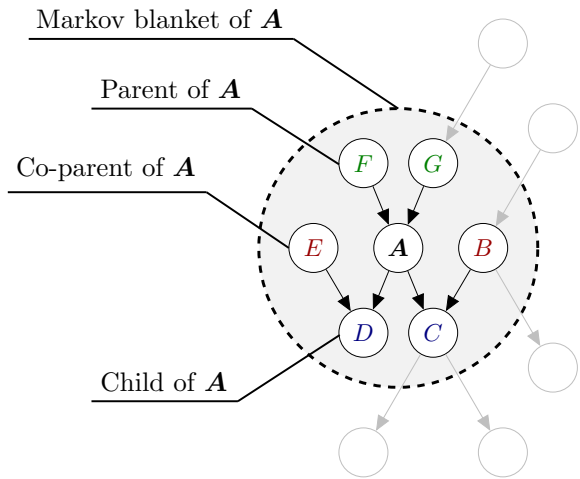}}
	\caption{This figure illustrates the Markov blanket of node $\bm{A}$, which is drawn in grey surrounded by a dashed line. The nodes $\textcolor{Green}{F}$ and $\textcolor{Green}{G}$ are the parents of $\bm{A}$ and the nodes $\textcolor{Blue}{C}$ and $\textcolor{Blue}{D}$ are the children of $\bm{A}$. The node $\textcolor{Red}{E}$ is the co-parent of $\bm{A}$ with respect to $\textcolor{Blue}{D}$ and the node $\textcolor{Red}{B}$ is the co-parent of $\bm{A}$ with respect to $\textcolor{Blue}{C}$.}
   \label{fig:MB}
\end{figure}

Restarting with the definition of $Q_k^*(x_k)$ and using the factorisation of the generative model, we get:
\begin{align*}
\ln Q_k^*(x_k) &\delequal \mathbb{E}_{\sim Q_k(x_k)}[\ln P(x,o)]\\
&= \mathbb{E}_{\sim Q_k(x_k)}[\ln \prod_i P(N_i| \text{pa}_i)],
\end{align*}
where $N_i$ iterates over all nodes, i.e. all latent and observed variables, and $\text{pa}_i$ are the parents of $N_i$. The term in the above product can be classified into three groups: the terms that do not depend on $x_k$, the terms whose target variable ($N_i$) is $x_k$ and the terms whose predictors ($\text{pa}_i$) contains $x_k$. Building on this observation, one can use the log property and the linearity of expectation to isolate the terms that depend on $x_k$:
\begin{align}
\ln Q_k^*(x_k) &= \langle \ln \prod_i P(N_i| \text{pa}_i) \rangle_{\sim Q_k} \nonumber \\
&= \langle \ln P(x_k|{\color{purple}\text{pa}_k}) \rangle_{\sim Q_k} + \sum_{c_j \in {\color{red}\text{ch}_k}} \langle \ln P(c_j|x_k, {\color{violet}\text{cp}_{kj}}) \rangle_{\sim Q_k} + C, \label{eq:VMP_42}
\end{align}
where $\langle \cdot \rangle_{\sim Q_k}$ is just another notation for $\mathbb{E}_{\sim Q_k(x_k)}[ \cdot ]$, and the constant $C$ comes from the terms of the product that do not depend on $x_k$. Equation \ref{eq:VMP_42} is the variational message passing equation that tells us how to compute the optimal posterior of any hidden state $x_k$ based on its Markov blanket, i.e. $x_k$'s parents ${\color{purple}\text{pa}_k}$, children ${\color{red}\text{ch}_k}$ and co-parents ${\color{violet}\text{cp}_{kj}}$. For readers unfamiliar with the notion of Markov blankets, Figure \ref{fig:MB} provides a visual depiction of the underlying notion.

\subsection{Conjugate exponential model} \label{cem-wbm}

\begin{figure}[H]
	\centering
	{\includegraphics[width=0.7\textwidth]{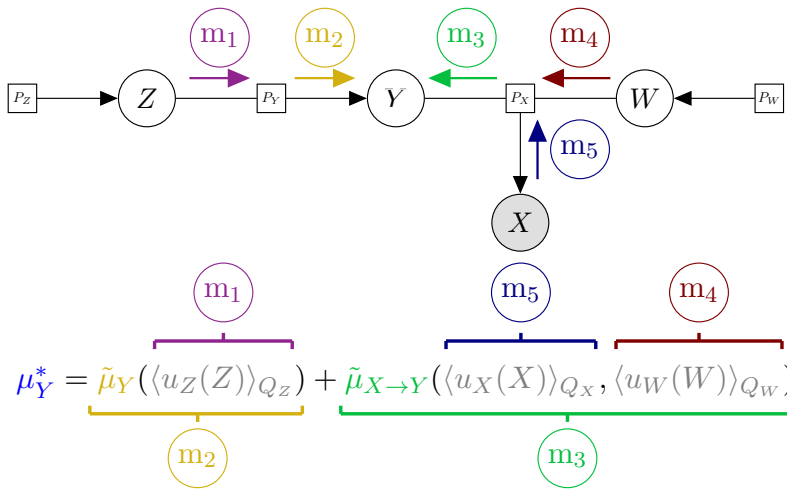}}
  \caption{This figure illustrates the computation of the optimal posterior parameters for the variable $Y$ as a message passing procedure, which requires the transmission of messages from the parent (${\color{Yellow}\text{m}_2}$) and child (${\color{Green}\text{m}_3}$) factors. Additionally, the message from the child factor (${\color{Green}\text{m}_3}$) requires the computation of messages from the co-parent (${\color{Red}\text{m}_4}$) and child (${\color{Blue}\text{m}_5}$) variables. Also, the message from the parent (${\color{Yellow}\text{m}_2}$) factor requires the computation of a message (${\color{Violet}\text{m}_1}$) from the parent variable. Set notation and associated brackets \{\} have been dropped, since there is only ever one parent or co-parent.}
   \label{fig:messages-ffg}
\end{figure}

The variational message passing algorithm can be derived for the class of conjugate exponential models \citep{VMP_TUTO}. Those models have a likelihood function and a prior in the exponential family. Furthermore, the prior and the likelihood are conjugate, meaning that the posterior will have the same form as the prior. We follows the steps in Winn and Bishop, while referring the interested reader to \citep{VMP_TUTO} for more details. The derivations in equations \ref{eq:vmp_prior_2}-\ref{eq:op_param} are clarified in the example in Figure \ref{fig:messages-ffg}.

Returning to our goal of computing the posterior over $x_k$ (cf. Equation \ref{eq:VMP_42}), we assume that $P(x_k|\text{pa}_k)$ and $P(c_j|x_k, \text{cp}_{kj})$ are in the exponential family, i.e.
\begin{align}\label{eq:vmp_prior_2}
\ln P(x_k|\text{pa}_k) = {\color{Yellow}\mu_k}(\text{pa}_k) \cdot {\color{purple}u_k}(x_k) + {\color{red}h_k}(x_k) + z_k(\text{pa}_k)
\end{align}
\begin{align}\label{eq:vmp_prior_x_2}
\ln P(c_j|x_k, \text{cp}_{kj}) = \mu_j(x_k,\text{cp}_{kj}) \cdot u_j(c_j) + h_j(c_j) + z_j(x_k,\text{cp}_{kj})
\end{align}
where $\mu_k(\text{pa}_k)$, $u_k(x_k)$, $h_k(x_k)$ and $z_k(\text{pa}_k)$ are the parameters, the sufficient statistics, the underlying measure and the log partition, respectively. For a specific example, Equation \ref{eq:D:0} shows the Dirichlet distribution written in the form of the exponential family. The first step of the Winn and Bishop method takes advantage of the conjugacy constraint to re-arrange Equation \ref{eq:vmp_prior_x_2} as a function of $u_k(x_k)$ that appears in Equation \ref{eq:vmp_prior_2}:
\begin{align}\label{eq:vmp_prior_x_3}
\ln P(c_j|x_k,\text{cp}_{kj}) = {\color{Green}\mu_{j \rightarrow k}}(c_j,\text{cp}_{kj}) \cdot {\color{purple}u_k}(x_k) + \lambda(c_j,\text{cp}_{kj}),
\end{align}
where $\mu_{j \rightarrow k}(c_j,\text{cp}_{kj})$ and $\lambda(c_j,\text{cp}_{kj})$ emerge from the re-arrangement. For a specific example of this first step, the reader is referred to the derivation from (\ref{eq:D:1}) to (\ref{eq:D:2}), Figure \ref{fig:messages-ffg} also provides an example of $\mu_{j \rightarrow k}(c_j,\text{cp}_{kj})$. The second step substitutes Equations \ref{eq:vmp_prior_x_3} and \ref{eq:vmp_prior_2} within the variational message passing equation leading to:
\begin{align*}
\ln Q_k^*(x_k) &= \langle {\color{Yellow}\mu_k}(\text{pa}_k) \cdot {\color{purple}u_k}(x_k) + {\color{red}h_k}(x_k) + z_k(\text{pa}_k) \rangle_{\sim Q_k}\\
&+ \sum_{c_j \in \text{ch}_k} \langle {\color{Green}\mu_{j \rightarrow k}}(c_j,\text{cp}_{kj}) \cdot {\color{purple}u_k}(x_k) + \lambda(c_j,\text{cp}_{kj}) \rangle_{\sim Q_k} + \text{Const}.
\end{align*}

The third step relies on taking the exponential of both sides, using the linearity of expectation and factorising by $u_k(x_k)$ to obtain:
\begin{align}\label{eq:opt_post_1}
Q_k^*(x_k) &= \exp \Bigg\{ \Big[ \langle {\color{Yellow}\mu_k}(\text{pa}_k) \rangle_{\sim Q_k} + \sum_{c_j \in \text{ch}_k} \langle {\color{Green}\mu_{j \rightarrow k}}(c_j,\text{cp}_{kj})\rangle_{\sim Q_k} \Big] \cdot {\color{purple}u_k}(x_k) + {\color{red}h_k}(x_k) + \text{Const} \Bigg\},
\end{align}
where the above constant just absorbed $z_k(\text{pa}_k)$ and $\lambda(c_j,\text{cp}_{kj})$, which does not depend on $x_k$. At this point, we already see that the prior (\ref{eq:vmp_prior_2}) and the approximate posterior (\ref{eq:opt_post_1}) have the same functional form, i.e., only their parameters differ. The fourth step re-parameterizes $\mu_k(\text{pa}_k)$ and $\mu_{j \rightarrow k}(c_j,\text{cp}_{kj})$ in terms of the expectation of the sufficient statistics of the children, parents and the co-parents:
\begin{align*}
Q_k^*(x_k) &= \exp \Bigg\{ {\color{blue}\mu_k^*} \cdot {\color{purple}u_k}(x_k) + {\color{red}h_k}(x_k) + \text{Const} \Bigg\}
\end{align*}
\begin{align}\label{eq:op_param}
{\color{blue}\mu_k^*} = {\color{Yellow}\tilde{\mu}_{k}}(\{{\color{gray}\langle u_{i}(i) \rangle_{Q_{i}}} \}_{i \in \text{pa}_k} ) + \sum_{c_j \in \text{ch}_k} {\color{Green}\tilde{\mu}_{j \rightarrow k}}({\color{gray}\langle u_j(c_j) \rangle_{Q_j}}, \{{\color{gray}\langle u_{l}(l)\rangle_{Q_l}} \}_{l \in \text{cp}_{kj}} ),
\end{align}
where ${\color{Yellow}\tilde{\mu}_{k}}$ is a re-parameterization of ${\color{Yellow}\mu_k}(pa_k)$ in terms of the expectation of the sufficient statistic of the parents of $x_k$, and similarly ${\color{Green}\tilde{\mu}_{j \rightarrow k}}$ is a re-parameterization of ${\color{Green}\mu_{j \rightarrow k}}$. The exact form of ${\color{Yellow}\tilde{\mu}_{k}}$ and ${\color{Green}\mu_{j \rightarrow k}}$ vary from distribution to distribution. An example of those re-parameterizations is visible from Equation \ref{eq:D:3.0} to \ref{eq:D:3}.

To understand the intuition behind (\ref{eq:op_param}), let us consider the following example: given the Forney factor graph illustrated in Figure \ref{fig:messages-ffg}, we wish to compute the posterior of $Y$. Then, the only parent of $Y$ is $Z$, the only child of $Y$ is $X$ and the only co-parent of $Y$ with respect to $X$ is $W$. Therefore, applying equation \ref{eq:op_param} to our example leads to the equation presented in Figure \ref{fig:messages-ffg} whose components can be interpreted as messages. Indeed, each variable (i.e. $X$, $Z$ and $W$) sends the expectation of their sufficient statistic (i.e. a message) to the square node in the direction of Y (i.e. either $P_X$ or $P_Y$). Those messages are then combined using a function (i.e. either $\tilde{\mu}_{Y}$ or $\tilde{\mu}_{X \rightarrow Y}$) whose output (i.e. another set of messages) are summed to obtain the optimal parameters $\mu_Y^*$. The computation of the optimal parameters (\ref{eq:op_param}) can then be understood as a message passing procedure.

Returning to the \citet{VMP_TUTO} method, the last step computes the (set of) expectations associated with $\{{\color{gray}\langle u_{j}(j) \rangle_{Q_{j}}} \}_{j \in \text{pa}_Y}$, ${\color{gray}\langle u_X(X) \rangle_{Q_X}}$, and $\{{\color{gray}\langle u_{j}(j)\rangle_{Q_{j}}} \}_{j \in \text{cp}_{YX}}$. Because all nodes of the model are in the exponential family, the moment generating function can be used to prove the following:
\begin{align}\label{eq:last_step}
{\color{gray}\langle u_{N}(N) \rangle_{Q_{N}}} = -\frac{\partial \tilde{z}_N(\theta_N)}{\partial \theta_N},
\end{align}
where $N$ is any node of the graphical model, $\theta_N$ are the natural parameters of the distribution over $N$, and $\tilde{z}_Z(\theta_N)$ is a re-parameterisation of the log partition w.r.t the natural parameters of the distribution over $Z$. Note that another way to compute those expectations will be presented in Section \ref{sec:messages:D}. 

\section{The link between Active Inference and Variational Message Passing}\label{sec:vmp-connection}

The previous sections have presented the theory behind active inference and variational message passing. This section focuses on the link between those two frameworks. First, we slightly modify the generative model and the variational distribution. These modifications concern a small part of the generative model and to ensure conjugacy between the random variables of the model. Then, we derive new update equations based on the Winn and Bishop method \citep{VMP_TUTO}. As we will see, those updates can be interpreted as a passing of messages that highlight the connection between variational message passing and belief updating in (planning as) active inference.

\subsection{Generative model modifications} 

\begin{figure}[H]
	\centering
	{\includegraphics[width=0.5\textwidth]{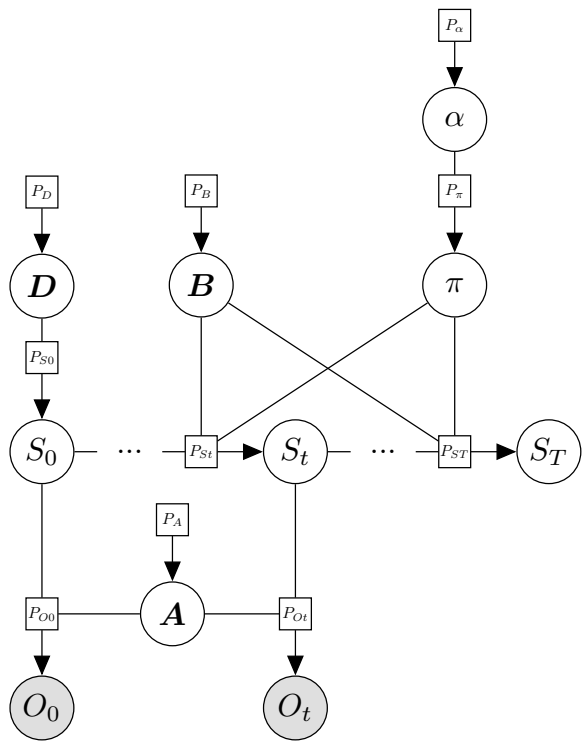}}
    \caption{The new generative model obtained after replacing the gamma distribution by a Dirichlet distribution.}
    \label{fig:modeified_GM}
\end{figure}

In order to perform variational message passing, we have made three modifications to the generative model described by Equation \ref{eq:gm}. First, the prior over the precision parameter $\gamma$ is removed. Second, the softmax function forming the prior over the policies is transformed into a categorical distribution with parameters $\alpha$. This is a mild modification because the softmax function is frequently used to represent a categorical distribution, e.g. neural classifiers using a softmax function as output layer or similarly to the updates of $Q(s_\tau)$ and $Q(\pi)$ presented in Section \ref{sec:ueq}. Finally, we assume a Dirichlet distribution over the parameters $\alpha$. Figure \ref{fig:modeified_GM} illustrates this new generative model where:
\begin{align*}
P(\pi|\alpha) &= \text{Cat}(\pi;\alpha)\\
P(\alpha) &= \text{Dir}(\alpha;\theta).
\end{align*}

The conjugacy between the Dirichlet and categorical distributions enables us to derive update equations that can be interpreted as messages. Recall that the prior over policies was used to bias the policy selection towards the policies that minimise expected free energy. This can be implemented in a straightfoward way --- while preserving conjugacy --- by setting the parameters of the Dirichlet as follows:
\begin{align*}
\theta = \overrightarrow{c} - \bm{G},
\end{align*}
where $\bm{G}$ is the expected free energy and $\overrightarrow{c}$ is a vector of constants whose elements satisfy the following properties:
\begin{enumerate}
\item $\forall i,j : \overrightarrow{c}_i = \overrightarrow{c}_j$, i.e. all elements are equal;
\item $\forall j : \overrightarrow{c}_j > \max_i \bm{G}_i$, i.e. all $\theta_j$ are strictly positive.
\end{enumerate}

To better understand the influence of $P(\alpha)$ on the selection of policies, we imagine a Dirichlet with $K$ parameters as a distribution over a $(K - 1)$-simplex. Assuming that all $\theta_i$ are greater than one, the point of this simplex with the highest probability, i.e. the mode $m_{\alpha}$, has the following coordinates:
$$m_{\alpha} = \begin{bmatrix}
\frac{\theta_{1} - 1}{\big(\sum_{k=1}^{K}\theta_{k}\big) - K} & \hdots & \frac{\theta_{K} - 1}{\big(\sum_{k=1}^{K}\theta_{k}\big) - K}
\end{bmatrix}.$$

Studying a few special cases of the above equation sheds some light on how policy selection is influenced by $P(\alpha)$. If the i-th numerator of the coordinates, i.e. $\theta_i - 1$, equal one and all others equal zero, then the mode $m_{\alpha}$ is at the corner of the simplex corresponding to the i-th axis. If all numerators are equal to one, then the mode is at the centre of the simplex. Intuitively, this means that the bigger $\theta_i$ is relative to the other $\theta_j\,\, \forall j \neq i$, the closer $m_{\alpha}$ is to the i-th corner of the simplex. Additionally, the closer $m_{\alpha}$ is to the i-th corner of the simplex, the more likely the i-th policy will be. Therefore, the bigger $\theta_i$ the more likely the i-th policy. Finally, the only part of the numerators that is not a constant is $\bm{G}_i$ and the smaller $\bm{G}_i$ the bigger the i-th numerator. Thus, in accord with the active inference literature, $P(\alpha)$ favours policies that minimise the expected free energy.

Another perspective on this parameterisation of priors over policies is to think of $\overrightarrow{c}$ as pseudo-counts that `promote' each policy according to how often it was previously pursued, before adding (-ve) expected free energy. If these pseudo-counts are suitably small, adding expected free energy will have a greater effect in the sense that expected free energy scores the number of times each policy would be pursued. Quantitatively, this means that a difference in the expected free energy between one policy and another can now be interpreted in terms of Dirichlet parameters or pseudo-counts.

It could be argued that the Dirichlet parameterisation of the prior over policies is a more natural parameterisation than the gamma distribution used to explain dopamine. Furthermore, as noted above, in most applications, gamma is set to one. More importantly, the precision parameter is only relevant for generative models where policies entail past transitions. In look-ahead policies or tree search implementations of planning, policies only concern future states. This means the precision of prior beliefs about policies relative to posterior beliefs (based upon the evidence a particular policy is being pursued) becomes irrelevant. In this case, the Dirichlet parameterisation above may be preferred.

\subsection{Variational distribution modifications} \label{sec:vd_modif}

\begin{figure}[H]
	\centering
	{\includegraphics[width=0.5\textwidth]{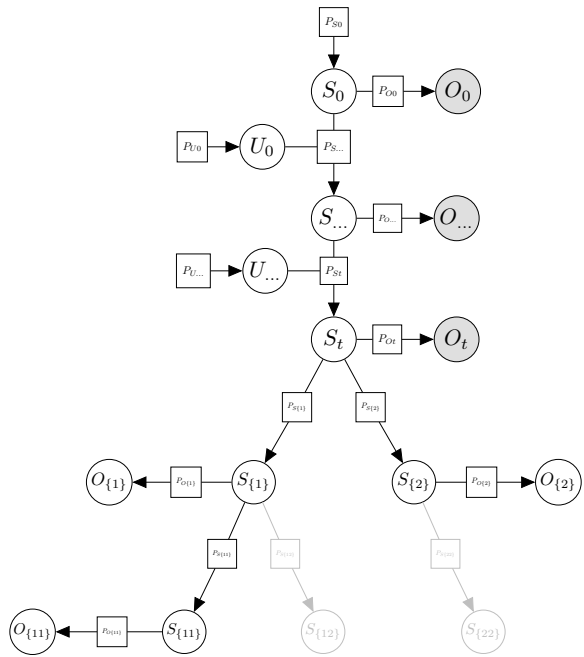}}
    \caption{
This figure illustrates an alternative new (expandable) generative model allowing planning under active inference. In this model, the future is now a tree like generative model whose branches correspond to the policies considered by the agent. Each edge connecting two states in the future correspond to an action and the nodes in light grey represent possible expansions of the current generative model.}
    \label{fig:AITS}
\end{figure}

The variational distribution presented in Section \ref{section:VD} is an example of a structured variational distribution, because factors such as $Q(S_\tau,\pi) = Q(S_\tau|\pi)Q(\pi)$ model the (posterior) dependency between $S_\tau$ and $\pi$. Performing inference with such a joint distribution falls under the category of structured variational inference \citep{SVI1,SVI2} and will not be covered in this paper. Instead, we assume a fully factorised distribution such that:
\begin{align*}
Q(S_{0:T}, \pi, \bm{A}, \bm{B}, \bm{D}, \gamma) = Q(\pi)Q(\bm{A})Q(\bm{B})Q(\bm{D})Q(\gamma) \prod_{\tau=0}^{T} Q(S_\tau),
\end{align*}
where $Q(\pi) = \text{Cat}(\pi;\tilde{\alpha})$, $Q(S_\tau) = \text{Cat}(S_\tau; \tilde{\bm{D}}_\tau)$ and all the other factors remain unchanged. This is a rather severe mean-field approximation: although it allows for straightforward application of variational message passing, removing the conditional dependencies of hidden states in the future on action means the agent cannot individuate the consequences of action. Under this functional form the expected free energy reduces to:
\begin{align*}
\bm{G}(\pi) = \sum_{\tau=1}^T \mathbb{E}_{Q(S_{\tau - 1}, \bm{B})}\Big[\text{H}[P(S_\tau | S_{\tau - 1}, \bm{B}, \pi)]\Big].
\end{align*}
Namely, the expected conditional entropy of the hidden states. Also, we refer the interested reader to Appendix H for a derivation of the above equation. Intuitively, this means that good policies select actions that lead to unambiguous hidden states. This highlights a major limitation of the mean-field approximation required by the variational message passing proposed by \citep{VMP_TUTO} in the context of active inference. In other words, when removing key structure from the variational distribution, the factor over the hidden states $Q(S_\tau|\pi)$ no longer depends on the policy $\pi$ and most of the terms in the expected free energy become constants w.r.t $\pi$. Figure \ref{fig:AITS} illustrates an alternative generative model, implementing tree search as a form of structure learning, which is not impacted by this issue because the future states in this model still depend upon the action undertaken by the agent. We refer the reader to our companion paper \citep{AI_TS_ME} for details. A related treatment that performs exact Bayesian inference by considering a slightly different generative model can be found in \citep{friston2020sophisticated}.

\begin{figure}[H]
	\centering
	{\includegraphics[width=0.7\textwidth]{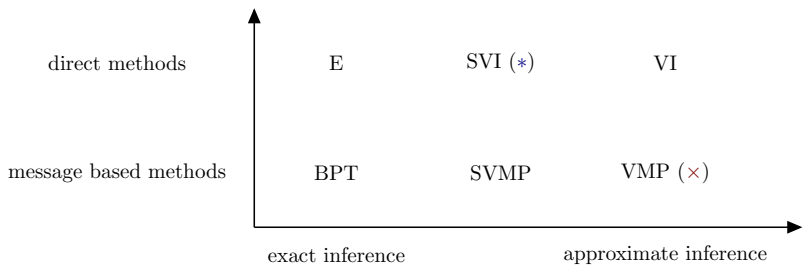}}
    \caption{
    This figure illustrates the differences between the framework presented in Section \ref{sec:ai} that belongs to the field of structured variational inference \citep{bishop2003structured} denoted by (${\color{Blue}*}$), and the work presented below that belongs to the field of variational message passing \citep{VMP_TUTO} denoted by (${\color{Red}\times}$). The other abbreviations BPT, E, VI and SVMP correspond to belief propagation on tree graphical models \citep{910572}, the elimination algorithm \citep{30008396546}, variational inference \citep{doi:10.1080/01621459.2017.1285773} and structured (or cluster) variational message passing \citep{lin2018variational}, respectively. Importantly, note that BPT is a specific kind of belief propagation which does not involve generalized BP \citep{10.5555/3008751.3008848} or loopy belief propagation \citep{murphy2013loopy}.
    }
    \label{fig:Framing}
\end{figure}

Before we turn to the derivation of the messages, we highlight the differences between active inference as presented in Section \ref{sec:ai} and the current treatment. The former is an example of structured variational inference $({\color{Blue}*})$. In contrast, the work presented in this section assumes a fully factorised variational distribution and will be strictly framed as a message passing algorithm, i.e. variational message passing $({\color{Red}*})$. Figure \ref{fig:Framing} illustrates those differences. Finally, in the remaining sections, we present the derivation of the messages for $\bm{D}$, $\bm{A}$, $\pi$ and $\alpha$, and we refer the reader to Appendices F and G for the derivations of the messages for $\bm{B}$ and $S_\tau$, respectively.

\subsection{Messages for D} \label{sec:messages:D}

This section applies the method of Winn and Bishop discussed in Section \ref{cem-wbm} to compute the messages of $\bm{D}$. Let us start with the definition of the Dirichlet and categorical distributions written in the form of the exponential family:
\begin{align}\label{eq:D:0}
\ln P(\bm{D};d) &= \underbrace{\begin{bmatrix}
d_1 - 1\\
...\\
d_{|S|} - 1
\end{bmatrix}}_{\mu_{D}(d)} \cdot
\underbrace{\begin{bmatrix}
\ln \bm{D}_1\\
...\\
\ln \bm{D}_{|S|}
\end{bmatrix}}_{u_{D}(\bm{D})}
\underbrace{- \ln B(d)}_{z_{D}(d)}
\end{align}

\begin{align}\label{eq:D:1}
\ln P(S_0;\bm{D}) &= \underbrace{\begin{bmatrix}
\ln \bm{D}_1\\
...\\
\ln \bm{D}_{|S|}
\end{bmatrix}}_{\mu_{S0}(\bm{D})} \cdot
\underbrace{\begin{bmatrix}
[S_0 = 1]\\
...\\
[S_0 = |S|]
\end{bmatrix}}_{u_{S0}(S_0)}
\end{align}
where $B(d)$ is the Beta function and $|S|$ is the number of values a hidden state can take. The first step requires us to re-write Equation \ref{eq:D:1} as a function of $u_{D}(\bm{D})$, this is straightforward because $\mu_{S0}(\bm{D})$ is just another name for $u_{D}(\bm{D})$. Using the fact that the inner product is commutative:
\begin{align}\label{eq:D:2}
\ln P(S_0;\bm{D}) &= \underbrace{\begin{bmatrix}
[S_0 = 1]\\
...\\
[S_0 = |S|]
\end{bmatrix}}_{\mu_{S_0\rightarrow \bm{D}}(S_0)}
\cdot
\underbrace{\begin{bmatrix}
\ln \bm{D}_1\\
...\\
\ln \bm{D}_{|S|}
\end{bmatrix}}_{u_{D}(\bm{D})}.
\end{align}

The second step aims to substitute Equations \ref{eq:D:0} and \ref{eq:D:2} within the variational message passing equation (\ref{eq:VMP_42}), i.e.
\begin{align*}
\ln Q^*(\bm{D}) &= \Big\langle \underbrace{\begin{bmatrix}
{\color{orange}d_1 - 1}\\
...\\
{\color{orange}d_{|S|} - 1}
\end{bmatrix}}_{\mu_{D}(d)} \cdot
\underbrace{\begin{bmatrix}
\ln \bm{D}_1\\
...\\
\ln \bm{D}_{|S|}
\end{bmatrix}}_{u_{D}(\bm{D})}
\underbrace{- \ln B(d)}_{z_{D}(d)}
\Big\rangle + \Big\langle \underbrace{\begin{bmatrix}
{\color{purple}[S_0 = 1]}\\
...\\
{\color{purple}[S_0 = |S|]}
\end{bmatrix}}_{\mu_{S_0\rightarrow \bm{D}}(S_0)}
\cdot
\underbrace{\begin{bmatrix}
\ln \bm{D}_1\\
...\\
\ln \bm{D}_{|S|}
\end{bmatrix}}_{u_{D}(\bm{D})} \Big\rangle + \text{Const},
\end{align*}
where $\langle \bigcdot \rangle$ refers to $\langle \bigcdot \rangle_{\sim Q_{\bm{D}}}$. Note that in the above equation, $d_i$ are fixed parameters, therefore there is not any posterior over $d$ and the first expectation $\langle \cdot \rangle_{\sim Q_{\bm{D}}}$ can be removed. The third step rests on taking the exponential of both sides, using the linearity of expectation and factorising by $u_{D}(\bm{D})$ to obtain:
\begin{align}\label{eq:D:3.0}
Q^*(\bm{D}) &= \exp \Bigg\{ \begin{bmatrix}
{\color{orange}d_1 - 1} + \langle {\color{purple}[S_0 = 1]} \rangle\\
...\\
{\color{orange}d_{|S|} - 1} + \langle {\color{purple}[S_0 = |S|]} \rangle
\end{bmatrix}
\cdot u_{D}(\bm{D}) + \text{Const} \Bigg\},
\end{align}

where $z_{D}(d)$ have been absorbed into the constant term because it does not depend on $\bm{D}$. The fourth step is a re-parameterisation done by observing that $\langle [S_0 = i] \rangle$ is the i-th element of the expectation of the vector $u_{S0}(S_0)$, i.e. $\langle u_{S0}(S_0) \rangle_i = \langle [S_0 = i] \rangle$:

\begin{align}\label{eq:D:3}
Q^*(\bm{D}) &= \exp \Bigg\{ \underbrace{\begin{bmatrix}
{\color{orange}d_1 - 1} + \langle {\color{purple}u_{S0}(S_0)} \rangle_1\\
...\\
{\color{orange}d_{|S|} - 1} + \langle {\color{purple}u_{S0}(S_0)} \rangle_{|S|}
\end{bmatrix}}_{\tilde{\mu}_{\bm{D}}(...) + \tilde{\mu}_{S_0 \rightarrow 	\bm{D}}(...)}
\cdot u_{D}(\bm{D}) + \text{Const} \Bigg\}.
\end{align}

The last step consists of computing the expectation of $\langle u_{S0}(S_0) \rangle_i$ for all $i$. This can be achieved by realising that the probability of an indicator function for an event is the probability of this event, i.e $\langle u_{S_0}(S_0) \rangle_i = \langle [S_0 = i] \rangle = Q(S_0 = i) = \tilde{\bm{D}}_{0i}$. Substituting this result in Equation \ref{eq:D:3}, leads to the final result:
\begin{align*}
Q^*(\bm{D}) &= \exp \Bigg\{ \begin{bmatrix}
{\color{orange}d_1 - 1} + {\color{purple}\tilde{\bm{D}}_{01}}\\
...\\
{\color{orange}d_{|S|} - 1} + {\color{purple}\tilde{\bm{D}}_{0|S|}}
\end{bmatrix}
\cdot u_{D}(\bm{D}) + \text{Const}\Bigg\}.
\end{align*}

Indeed, the above equation is in fact a Dirichlet distribution in exponential family form, and can be re-written into its usual form to obtain the final update equation:
\begin{align*}
Q^*(\bm{D}) &= \text{Dir}(\bm{D}; {\color{orange}d} + {\color{purple}\tilde{\bm{D}}_{0}}).
\end{align*}

In the following sections, we provide derivations for the messages of $\bm{A}$, $\bm{B}$, $\pi$, $\alpha$, and $S_\tau$. Those derivations are similar to the one presented above. We encourage technical readers to go through those derivations because they constitute the main contribution of this paper. However, a reader uninterested in the algebraic details of the proofs may want to jump to Section \ref{sec:msg_sum}.

\subsection{Messages for A}

In the previous section, we have shown how to compute the messages for $\bm{D}$, which are based on the conjugacy between a categorical $P(S_0|\bm{D})$ and a Dirichlet $P(\bm{D};d)$ distributions. In this section, we dive into the derivation of the messages of $\bm{A}$, which relies on the same kind of conjugacy. We start with the definition of $P(\bm{A};a)$, which is a product of Dirichlet distributions. This product can be turned into a sum by taking the logarithm of both sides and using the log property to obtain:
\begin{align}\label{eq:A:0}
\ln P(\bm{A};a) &= \ln \prod_{i} P(\bm{A}_{\bigcdot i};a_{\bigcdot i}) = \sum_{i} \ln \text{Dir}(\bm{A}_{\bigcdot i};a_{\bigcdot i})\nonumber\\
&= \sum_{i}  \underbrace{\begin{bmatrix}
a_{1i} - 1\\
...\\
a_{|O|i} - 1
\end{bmatrix} \cdot
\begin{bmatrix}
\ln \bm{A}_{1i}\\
...\\
\ln \bm{A}_{|O|i}
\end{bmatrix}
- \ln B(a_{\bigcdot i})}_{\text{Logarithm of Dirichlet}}\nonumber\\
&= \underbrace{\begin{bmatrix}
a_{11} - 1\\
...\\
a_{|O||S|} - 1
\end{bmatrix}}_{\mu_{A}(a)} \cdot
\underbrace{\begin{bmatrix}
\ln \bm{A}_{11}\\
...\\
\ln \bm{A}_{|O||S|}
\end{bmatrix}}_{u_{A}(\bm{A})}
\underbrace{- \sum_i \ln B(a_{\bigcdot i})}_{z_{A}(a)},
\end{align}
where $|O|$ is the number of possible outcomes. Note that the vectors $u_{A}(\bm{A})$ and $\mu_{A}(a)$ step through all the elements of the matrices $\bm{A}$ and $a$, respectively. Also, for each time step $\tau$ up to the present time $t$, the random matrix $\bm{A}$ has one child $O_\tau$ (see Figure \ref{fig:modeified_GM}), and its probability mass function $P(O_\tau | \bm{A}, S_\tau)$ is a product of categorical distributions that can be written as:
\begin{align}\label{eq:A:1}
\ln P(O_\tau = k | \bm{A}, S_\tau = l) &= \ln \bm{A}_{kl}\nonumber\\
&= \sum_{i,j} [O_\tau = i][S_{\tau} = j] \ln \bm{A}_{ij}\nonumber\\
&= \underbrace{\begin{bmatrix}
[S_{\tau} = 1]\ln \bm{A}_{1 1}\\
...\\
[S_{\tau} = |S|]\ln \bm{A}_{|O||S|}
\end{bmatrix}}_{\mu_{O_\tau}(\bm{A},S_{\tau})} \cdot
\underbrace{\begin{bmatrix}
[O_\tau = 1]\\
...\\
[O_\tau = |O|]
\end{bmatrix}}_{u_{O_\tau}(O_\tau)}.
\end{align}

Finally, the re-parameterisation in the fourth step will require the probability mass function of $S_\tau$ (see Figure \ref{fig:modeified_GM}), i.e. the co-parent of $\bm{A}$ with respect to $O_\tau$, to be written in the form of the exponential family as follows:
\begin{align}\label{eq:A:11}
\ln P(S_\tau = k | \bm{B}, S_{\tau - 1} = l, \pi = m) &= \ln \bm{B}[U_{\tau - 1}^m]_{kl}\nonumber\\
&= \sum_{i,j,k,u} [S_\tau = i][S_{\tau - 1} = j][\pi = k][U_{\tau - 1}^k = u] \ln \bm{B}[u]_{ij}\nonumber\\
&= \mu_{S_\tau}(\bm{B},S_{\tau - 1},\pi) \cdot u_{S_\tau}(S_\tau),
\end{align}
where:
\begin{align*}
\mu_{S_\tau}(\bm{B},S_{\tau - 1},\pi) = \begin{bmatrix}
\sum_{j,k,u} [S_{\tau - 1} = j][\pi = k][U_{\tau - 1}^k = u]\ln \bm{B}[u]_{1 j}\\
...\\
\sum_{j,k,u} [S_{\tau - 1} = j][\pi = k][U_{\tau - 1}^k = u]\ln \bm{B}[u]_{|S| j}
\end{bmatrix},
\end{align*}
and:
\begin{align*}
u_{S_\tau}(S_\tau) = \begin{bmatrix}
[S_\tau = 1]\\
...\\
[S_\tau = |S|]
\end{bmatrix}.
\end{align*}

The first step requires us to re-write Equation \ref{eq:A:1} as a function of $u_{A}(\bm{A})$, this is done by expanding the inner product and re-arranging:
\begin{align}\label{eq:A:2}
\ln P(O_\tau | \bm{A}, S_\tau) &= \underbrace{\begin{bmatrix}
[O_\tau = 1][S_\tau = 1]\\
...\\
[O_\tau = |O|][S_\tau = |S|]
\end{bmatrix}}_{\mu_{O_\tau \rightarrow \bm{A}}(O_\tau,S_\tau)} \cdot \underbrace{\begin{bmatrix}
\ln \bm{A}_{1 1}\\
...\\
\ln \bm{A}_{|O||S|}
\end{bmatrix}}_{u_{A}(\bm{A})}.
\end{align}

The second step aims to substitute Equations \ref{eq:A:0} and \ref{eq:A:2} within the variational message passing equation (\ref{eq:VMP_42}), i.e.
\begin{align*}
\ln Q^*(\bm{A}) &= \Big\langle \begin{bmatrix}
{\color{orange}a_{11} - 1}\\
...\\
{\color{orange}a_{|O||S|} - 1}
\end{bmatrix} \cdot u_{A}(\bm{A}) \Big\rangle + \sum_{\tau = 0}^{t} \Big\langle \begin{bmatrix}
{\color{purple}[O_\tau = 1][S_\tau = 1]}\\
...\\
{\color{purple}[O_\tau = |O|][S_\tau = |S|]}
\end{bmatrix} \cdot u_{A}(\bm{A}) \Big\rangle + \text{Const},
\end{align*}
where $\langle \bigcdot \rangle$ refers to $\langle \bigcdot \rangle_{\sim Q_{\bm{A}}}$. The third step builds on this equation by pulling the sum over all time steps $\tau$ inside the vector, using the linearity of expectation, factorising $u_{A}(\bm{A})$, and taking the exponential of both sides:
\begin{align*}
Q^*(\bm{A}) &= \exp \Bigg\{ \begin{bmatrix}
{\color{orange}a_{11} - 1} + \sum_{\tau = 0}^{t} \langle {\color{purple}[O_\tau = 1]} \rangle \langle {\color{purple}[S_\tau = 1]} \rangle\\
...\\
{\color{orange}a_{|O||S|} - 1} + \sum_{\tau = 0}^{t} \langle {\color{purple}[O_\tau = |O|]} \rangle \langle {\color{purple}[S_\tau = |S|]} \rangle
\end{bmatrix} \cdot u_{A}(\bm{A}) + \text{Const} \Bigg\},
\end{align*}
where we used that $a_{ji}$ are hyperparameters that are constant w.r.t the expectation $\langle \bigcdot \rangle_{\sim Q_{\bm{A}}}$. The fourth step consists of two re-parameterisations performed by observing that $\langle [O_\tau = j] \rangle$ and $\langle [S_\tau = i] \rangle$ are the expectations of the j-th and i-th elements of the vectors $u_{O_\tau}(O_\tau)$ and $u_{S_\tau}(S_\tau)$, respectively (cf. Equation \ref{eq:A:1} and \ref{eq:A:11}). Substituting those re-parameterisations in the above equation leads to:
\begin{align}\label{eq:A:3}
Q^*(\bm{A}) &= \exp \Bigg\{ \underbrace{\begin{bmatrix}
{\color{orange}a_{11} - 1} + \sum_{\tau = 0}^{t} \langle {\color{purple}u_{O_\tau}(O_\tau)} \rangle_1  \langle {\color{purple}u_{S_\tau}(S_\tau)} \rangle_1\\
...\\
{\color{orange}a_{KN} - 1} + \sum_{\tau = 0}^{t} \langle {\color{purple}u_{O_\tau}(O_\tau)} \rangle_{|O|} \langle {\color{purple}u_{S_\tau}(S_\tau)} \rangle_{|S|}
\end{bmatrix}}_{\tilde{\mu}_{\bm{A}}(...) + \sum_\tau \tilde{\mu}_{O_\tau \rightarrow \bm{A}}(...)}
\cdot u_{A}(\bm{A}) + \text{Const} \Bigg\}.
\end{align}

The last step consists of computing the expectation of $\langle {\color{purple}u_{O_\tau}(O_\tau)} \rangle_i$ and $\langle {\color{purple}u_{S_\tau}(S_\tau)} \rangle_j$ for all $i$ and $j$. Since, the probability of an indicator function for an event is the probability of this event, we are searching for the probabilities of $O_\tau = j$ and $S_\tau = i$. The probability of $O_\tau = j$ is the j-th element of the vector $\bm{o_\tau}$, which is a one hot vector containing the observation from the environment at time $\tau$. The posterior probability of $S_\tau$ is by definition $Q(S_\tau) = \tilde{\bm{D}}_{\tau}$. Substituting the probabilities of $O_\tau = j$ and $S_\tau = i$ in Equation \ref{eq:A:3}, leads to:
\begin{align}
Q^*(\bm{A}) &= \exp \Bigg\{ \begin{bmatrix}
{\color{orange}a_{11} - 1} + \sum_{\tau = 0}^{t} {\color{purple}\bm{o}_{\tau 1} \tilde{\bm{D}}_{\tau 1}}\\
...\\
{\color{orange}a_{|O||S|} - 1} + \sum_{\tau = 0}^{t} {\color{purple}\bm{o}_{\tau |O|} \tilde{\bm{D}}_{\tau |S|}}
\end{bmatrix}
\cdot u_{A}(\bm{A}) + \text{Const} \Bigg\}\label{eq:A:4}\\
&= \prod_i \exp \Bigg\{ \begin{bmatrix}
{\color{orange}a_{1i} - 1} + \sum_{\tau = 0}^{t} {\color{purple}\bm{o}_{\tau 1} \tilde{\bm{D}}_{\tau i}}\\
...\\
{\color{orange}a_{|O|i} - 1} + \sum_{\tau = 0}^{t} {\color{purple}\bm{o}_{\tau |O|} \tilde{\bm{D}}_{\tau i}}
\end{bmatrix}
\cdot \begin{bmatrix}
\ln \bm{A}_{1 i}\\
...\\
\ln \bm{A}_{|O| i}
\end{bmatrix}+ \text{Const} \Bigg\}.\label{eq:A:5}
\end{align}

Finally, one can recognise in Equation \ref{eq:A:5} the product of Dirichlet distributions written into their exponential form, i.e.
\begin{align*}
Q^*(\bm{A}) &= \prod_i \text{Dir}(\bm{A}_{\bigcdot i}, \bm{a}_{\bigcdot i}) \text{ where } \bm{a} = {\color{orange}a} + \sum_\tau {\color{purple}\bm{o}_\tau} \otimes {\color{purple}\tilde{\bm{D}}_\tau}.
\end{align*}

The origin of the outer product in the computation of the parameters can be understood by considering $P^\tau$ the outer product between $\bm{o_{\tau}}$ and $\bm{s_{\tau}}$ such that $P^\tau_{ij} = \bm{o}_{\tau i}\bm{s}_{\tau j}$. Then, Equation \ref{eq:A:4} shows that: $\bm{a}_{ij} = a_{ij} + \sum_\tau P^\tau_{ij} \Leftrightarrow \bm{a} = a + \sum_\tau \bm{o}_\tau \otimes \bm{s}_\tau$.

\subsection{Messages for $\pi$}

We now turn to the messages for $\pi$. Note, that the definition of the $P(S_\tau| \bm{B}, S_{\tau - 1}, \pi)$ and $P(\pi|\alpha)$ are given by Equations \ref{eq:A:11} and \ref{eq:pi:def}, respectively. The first step requires us to re-write Equation \ref{eq:A:11} as a function of $u_{\pi}(\pi)$. Using the inner product definition and re-arranging we obtain:
\begin{align}\label{eq:pi:2}
\ln P(S_\tau = k | \bm{B}, S_{\tau - 1} = l, \pi = m) &= \begin{bmatrix}
\sum_{i,j,u} [S_\tau = i][U_{\tau - 1}^1 = u][S_{\tau - 1} = j]\ln \bm{B}[u]_{i j}\\
...\\
\sum_{i,j,u} [S_\tau = i][U_{\tau - 1}^{|\pi|} = u][S_{\tau - 1} = j]\ln \bm{B}[u]_{i j}
\end{bmatrix} \cdot u_\pi(\pi).
\end{align}

The second step aims to substitute Equations \ref{eq:pi:def} and \ref{eq:pi:2} within the variational message passing equation, i.e.
\begin{align*}
\ln Q^*(\pi) &= \Big\langle \begin{bmatrix}
{\color{orange} \ln \alpha_1} \\
...\\
{\color{orange} \ln \alpha_{|\pi|}}
\end{bmatrix} \cdot u_\pi(\pi) \Big\rangle\\
&+ \sum_{\tau = 1}^T \Big\langle \begin{bmatrix}
\sum_{i,j,u} {\color{purple}[S_\tau = i][U_{\tau - 1}^1 = u][S_{\tau - 1} = j]\ln \bm{B}[u]_{i j}}\\
...\\
\sum_{i,j,u} {\color{purple}[S_\tau = i][U_{\tau - 1}^{|\pi|} = u][S_{\tau - 1} = j]\ln \bm{B}[u]_{i j}}
\end{bmatrix} \cdot u_\pi(\pi) \Big\rangle + \text{Const},
\end{align*}
where $\langle \bigcdot \rangle$ refers to $\langle \bigcdot \rangle_{\sim Q_{\pi}}$. The third step relies on pulling the summation over all time steps inside the vector, taking the exponential of both sides, using the linearity of expectation and factorising by $u_{\pi}(\pi)$ to obtain:
\begin{align*}
Q^*(\pi) &\propto \exp \Bigg\{ \underbrace{\begin{bmatrix}
\langle {\color{orange}\ln \alpha_1} \rangle + \sum_{\tau,i,j,u} {\color{purple}[U_{\tau - 1}^1 = u]}\langle {\color{purple}[S_\tau = i]} \rangle\langle {\color{purple}[S_{\tau - 1} = j]} \rangle\langle {\color{purple}\ln \bm{B}[u]_{i j}} \rangle\\
...\\
\langle {\color{orange}\ln \alpha_{|\pi|}} \rangle + \sum_{\tau,i,j,u} {\color{purple}[U_{\tau - 1}^{|\pi|} = u]} \langle {\color{purple}[S_\tau = i]} \rangle\langle {\color{purple}[S_{\tau - 1} = j]} \rangle\langle {\color{purple}\ln \bm{B}[u]_{i j}} \rangle
\end{bmatrix}}_{\mu^*_\pi} \cdot u_\pi(\pi) \Bigg\}.
\end{align*}

The fourth step is a re-parameterisation implemented by observing that $\langle \ln \alpha_k \rangle$, $\langle [S_\tau = i] \rangle$, $\langle [S_{\tau - 1} = j] \rangle$ and $\langle \ln \bm{B}[u]_{i j} \rangle$ are elements of the vectors $\langle u_\alpha(\alpha) \rangle$, $\langle u_{S_\tau}(S_\tau) \rangle$, $\langle u_{S_{\tau - 1}}(S_{\tau - 1}) \rangle$ and $\langle u_B(\bm{B}) \rangle$, respectively:
\begin{align}\label{eq:pi:3}
\mu^*_\pi &= \begin{bmatrix}
\langle {\color{orange}u_\alpha(\alpha)} \rangle_1 + \sum_{\tau,i,j,u} {\color{purple}[U_{\tau - 1}^1 = u]}\langle {\color{purple}u_{S_\tau}(S_\tau)} \rangle_i \langle {\color{purple}u_{S_{\tau - 1}}(S_{\tau - 1})} \rangle_j \langle {\color{purple}u_B(\bm{B})} \rangle_{u,i,j}\\
...\\
\langle {\color{orange}u_\alpha(\alpha)} \rangle_{|\pi|} + \sum_{\tau,i,j,u} {\color{purple}[U_{\tau - 1}^{|\pi|} = u]} \langle {\color{purple}u_{S_\tau}(S_\tau)} \rangle_i \langle {\color{purple}u_{S_{\tau - 1}}(S_{\tau - 1})} \rangle_j \langle {\color{purple}u_B(\bm{B})} \rangle_{u,i,j}
\end{bmatrix}.
\end{align}

The last step consists of computing the expectation of $\langle u_\alpha(\alpha) \rangle_k$, $\langle u_{S_\tau}(S_\tau) \rangle_i$, $\langle u_{S_{\tau - 1}}(S_{\tau - 1}) \rangle_j$ and $\langle u_B(\bm{B}) \rangle_{u,i,j}$ for all $i$, $j$, $k$ and $u$:

\begin{itemize}
\item $\langle u_\alpha(\alpha) \rangle_k = \langle \ln \alpha_k \rangle = \psi(\tilde{\alpha}_k) - \psi(\sum_l \tilde{\alpha}_l) \delequal \bar{\alpha}_k$
\item $\langle u_{S_\tau}(S_\tau) \rangle_i = \langle [S_\tau = i] \rangle = \tilde{\bm{D}}_{\tau i}$
\item $\langle u_{S_{\tau - 1}}(S_{\tau - 1}) \rangle_j = \langle [S_{\tau - 1} = j] \rangle = \tilde{\bm{D}}_{(\tau - 1)j}$
\item $\langle u_B(\bm{B}) \rangle_{u,i,j} = \langle \ln \bm{B}[u]_{i j} \rangle = \psi(\bm{b}[u]_{i j}) - \psi(\sum_l \bm{b}[u]_{l j}) \delequal \bar{\bm{B}}[u]_{i j}$
\end{itemize}

Furthermore, the indicator function in the k-th row of Equation \ref{eq:pi:3} filters out all elements where $u \neq U_{\tau - 1}^k$. Substituting those results in Equation \ref{eq:pi:3}, leads to the final result:
\begin{align*}
Q^*(\pi) &\propto \exp \Bigg\{\begin{bmatrix}
{\color{orange} \bar{\alpha}_1} + \sum_{\tau,i,j} {\color{purple} \tilde{\bm{D}}_{\tau i}} {\color{purple}\tilde{\bm{D}}_{(\tau - 1)j}} {\color{purple} \bar{\bm{B}}[U_{\tau - 1}^1]_{i j}}\\
...\\
{\color{orange}\bar{\alpha}_{|\pi|}} + \sum_{\tau,i,j} {\color{purple}\tilde{\bm{D}}_{\tau i}} {\color{purple}\tilde{\bm{D}}_{(\tau - 1)j}} {\color{purple} \bar{\bm{B}}[U_{\tau - 1}^{|\pi|}]_{i j}}
\end{bmatrix} \cdot u_\pi(\pi) \Bigg\}.
\end{align*}

Indeed, the above equation is a Categorical distribution in the exponential family form, and can be re-written into its usual form as follows:
\begin{align*}
Q^*(\pi) &= \text{Cat}(\pi; \alpha^*) \,\, \text{ where } \,\, \alpha^* = \sigma\Bigg({\color{orange}\bar{\alpha}} + \sum_{\tau = 1}^T {\color{purple} \mathbb{F}_\tau}\Bigg) \,\, \text{ and } \,\, \mathbb{F}_\tau = \begin{bmatrix}
\langle \tilde{\bm{D}}_{\tau} \otimes \tilde{\bm{D}}_{\tau - 1} , \bar{\bm{B}}[U_{\tau - 1}^1] \rangle_F \\
...\\
\langle \tilde{\bm{D}}_{\tau} \otimes \tilde{\bm{D}}_{\tau - 1} , \bar{\bm{B}}[U_{\tau - 1}^{|\pi|}] \rangle_F
\end{bmatrix},
\end{align*}
where it should be stressed that $\langle \bigcdot , \bigcdot \rangle_F$ is not an expectation but the Frobenius product, i.e. a generalisation of the inner product to matrices.

\subsection{Messages for $\alpha$}

In this section, we focus on the messages for $\alpha$, whose derivation is identical to the messages of $\bm{D}$. To see this, note that $P(\bm{D})$ was a Dirichlet with parameters $d$. Furthermore, the only child of $\bm{D}$ was $S_0$ whose prior and posterior were categorical distributions with parameters $\bm{D}$ and $\tilde{\bm{D}}$. Similarly, note that $P(\alpha)$ is a Dirichlet with parameters $\theta$. Furthermore, the only child of $\alpha$ is $\pi$ whose prior and posterior are categorical distributions with parameters $\alpha$ and $\tilde{\alpha}$. From this observation, we directly obtain the following result:
\begin{align*}
Q^*(\alpha) &= \text{Dir}(\alpha; {\color{orange}\theta} + {\color{purple}\tilde{\alpha}}).
\end{align*}

\subsection{Summary of messages} \label{sec:msg_sum}

Next, we focus on explaining the intuition behind the resulting equations. The first point is the coloration of the equations in orange and purple. The orange colour corresponds to messages from the parent factors, which correspond to messages of type $\text{m}_2$ in Figure \ref{fig:messages-ffg}. This means that each orange message is a function of the expectation of the sufficient statistic of the parent variables, i.e. a function of messages of type $\text{m}_1$. Similarly, the purple colour corresponds to messages from the child factors, which correspond to messages of type $\text{m}_3$ in Figure \ref{fig:messages-ffg}. Once again, this means that each purple message is a function of the sufficient statistics of the co-parent and child variables, i.e. a function of messages of type $\text{m}_4$ and $\text{m}_5$, respectively. Let's see how these play out in our newly derived equations.

\paragraph{Messages for $\alpha$:}
\begin{align*}
Q^*(\alpha) &= \text{Dir}(\alpha; {\color{orange}\theta} + {\color{purple}\tilde{\alpha}})
\end{align*}

Recall that $\mu_\alpha = \theta$ is an $\text{m}_2$ message (orange colour). However, $\alpha$ does not have any parent variables thus $\mu_\alpha$ is a constant, i.e. a function of zero $\text{m}_1$ messages. Furthermore, we know that $\alpha$ has only one child variable ($\pi$) and no co-parent variables. Therefore, $\mu_{\pi \rightarrow \alpha}(\tilde{\alpha}) = \tilde{\alpha}$ is the only $\text{m}_3$ message (purple colour) for $\alpha$, where $\tilde{\alpha} = \langle u_\pi(\pi) \rangle_{Q_\pi}$ is an $\text{m}_5$ message.

\paragraph{Messages for $\bm{D}$:}
\begin{align*}
Q^*(\bm{D}) &= \text{Dir}(\bm{D}; {\color{orange}d} + {\color{purple}\tilde{\bm{D}}_{0}})
\end{align*}

Similarly for the messages of $\alpha$, $\mu_{\bm{D}} = d$ and $\mu_{S_0 \rightarrow \bm{D}}(\tilde{\bm{D}}_{0}) = \tilde{\bm{D}}_{0}$, where $\tilde{\bm{D}}_{0}$ should be thought of as a message from a child variable ($\text{m}_5$ message).

\paragraph{Messages for $\bm{A}$:}
\begin{align*}
Q^*(\bm{A}) &= \prod_i \text{Dir}(\bm{A}_{\bigcdot i}, \bm{a}_{\bigcdot i}) \text{ where } \bm{a} = {\color{orange}a} + \sum_\tau {\color{purple}\bm{o}_\tau} \otimes {\color{purple}\tilde{\bm{D}}_\tau}
\end{align*}

\begin{figure}[H]
	\centering
	{\includegraphics[width=0.7\textwidth]{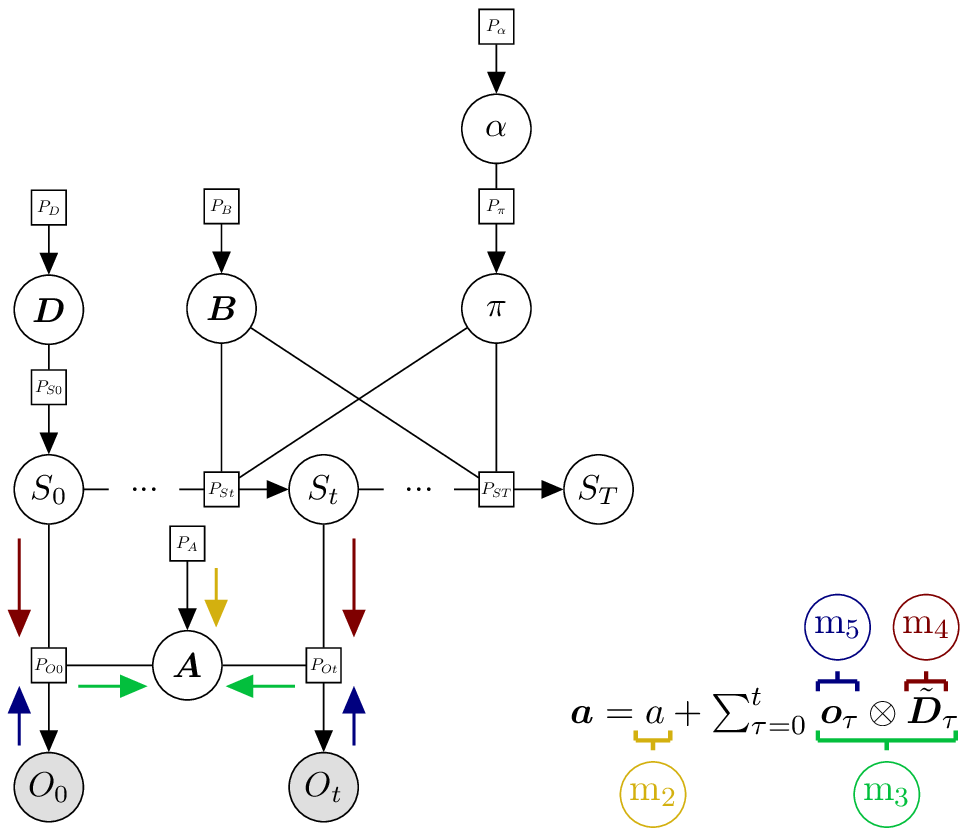}}
    \caption{
This figure illustrates the passing of messages required to update the posterior over $\bm{A}$. The messages of type ${\color{Yellow}\text{m}_2}$, ${\color{Green}\text{m}_3}$, ${\color{Red}\text{m}_4}$ and ${\color{Blue}\text{m}_5}$ come from the parent factors, child factors, co-parent variables and child variables, respectively.}
    \label{fig:A_msgs}
\end{figure}

Following the same reasoning, $\mu_{\bm{A}} = a$ is an $\text{m}_2$ message and because $\bm{A}$ does not have any parent variables then $\mu_{\bm{A}}$ is a constant. Also, $\bm{A}$ has one child variable ($O_\tau$) for each time step $\tau \in \llbracket 0, t \rrbracket$ and one co-parent variable ($S_\tau$) for each of them, which implies that there are $t + 1 \,\, \text{m}_3$ messages for $\bm{A}$, i.e. $\mu_{O_\tau \rightarrow \bm{A}}(\bm{o}_\tau, \tilde{\bm{D}}_\tau) = \bm{o}_\tau \otimes \tilde{\bm{D}}_\tau\,\, \forall \tau \in \llbracket 0, t \rrbracket$. Because the $O_\tau$ are observed, we know that the $\text{m}_5$ messages transmitted by this node will be the observation made at time $\tau$ ($\bm{o}_\tau$). Additionally, the $\text{m}_4$ message from the hidden variables $S_\tau$ are the expectation of their sufficient statistics, i.e. $\langle u_{S_\tau}(S_\tau) \rangle_{Q_{S_\tau}} = \tilde{\bm{D}}_\tau$. This confirms the idea that $\mu_{O_\tau \rightarrow \bm{A}}$ is a function of the sufficient statistics of the child and co-parent variables. Figure \ref{fig:A_msgs} concludes this paragraph with a visual representation of the messages for $\bm{A}$.

\paragraph{Messages for $\bm{B}$:}
\begin{align*}
Q^*(\bm{B}) &= \prod_{u,i} \text{Dir}(\bm{B}[u]_{\bigcdot i}, \bm{b}[u]_{\bigcdot i}) \quad \text{ where } \quad \bm{b}[u] = {\color{orange}b[u]} + \sum_{(k,\tau) \in \Omega_u} {\color{purple}\tilde{\alpha}_k} {\color{purple}\tilde{\bm{D}}_{\tau}} \otimes {\color{purple}\tilde{\bm{D}}_{\tau - 1}}
\end{align*}

Sticking with this reasoning, $\mu_{\bm{B}} = b$ is an $\text{m}_2$ message and because $\bm{B}$ does not have any parent variables then $\mu_{\bm{B}}$ is a constant equal to $b$. Also, $\bm{B}$ has one child variable ($S_\tau$) for each time step $\tau \in \llbracket 1, T \rrbracket$ and all policies $\forall \pi \in \llbracket 1, |\pi| \rrbracket$, along with two co-parent variables ($S_{\tau - 1}$ and $\pi$) for each of those child variables. This implies that there are $T \times |\pi| \,\, \text{m}_3$ messages for $\bm{B}$, i.e. $\mu_{S_{\tau} \rightarrow S_{\bm{B}}}(\tilde{\alpha}_k, \tilde{\bm{D}}_{\tau}, \tilde{\bm{D}}_{\tau - 1}) = \tilde{\alpha}_k \tilde{\bm{D}}_{\tau} \otimes \tilde{\bm{D}}_{\tau - 1}, \,\, \forall \tau \in \llbracket 1, T \rrbracket, \,\, \forall \pi \in \llbracket 1, |\pi| \rrbracket$ where $\tilde{\bm{D}}_{\tau}$ is an $\text{m}_5$ message and $\tilde{\alpha}_k$ along with $\tilde{\bm{D}}_{\tau - 1}$ are $\text{m}_4$ messages. 

\paragraph{Messages for $\pi$:}
\begin{align*}
Q^*(\pi) &= \text{Cat}(\pi; \alpha^*) \,\, \text{ where } \,\, \alpha^* = \sigma\Bigg({\color{orange}\bar{\alpha}} + \sum_{\tau = 1}^T {\color{purple} \mathbb{F}_\tau}\Bigg) \,\, \text{ and } \,\, \mathbb{F}_\tau = \begin{bmatrix}
\langle \tilde{\bm{D}}_{\tau} \otimes \tilde{\bm{D}}_{\tau - 1} , \bar{\bm{B}}[U_{\tau - 1}^1] \rangle_F \\
...\\
\langle \tilde{\bm{D}}_{\tau} \otimes \tilde{\bm{D}}_{\tau - 1} , \bar{\bm{B}}[U_{\tau - 1}^{|\pi|}] \rangle_F
\end{bmatrix}
\end{align*}

If we keep applying the same reasoning, we see that $\mu_\pi(\bar{\alpha}) = \bar{\alpha}$ is an $\text{m}_2$ message, which is a function of the sufficient statistics of the parent variable $\alpha$ ($\text{m}_1$ message). Moreover, $\pi$ has one child variable ($S_\tau$) for each time step $\tau \in \llbracket 1, T \rrbracket$, and for each of those child variables, $\pi$ has two co-parent variables ($S_{\tau - 1}$ and $\bm{B}$). Therefore, $\mu_{S_\tau \rightarrow \pi} = \mathbb{F}_\tau \,\, \forall \tau \in \llbracket 1, T \rrbracket$ correspond to $T \,\, \text{m}_3$ messages. Those messages are function of two $\text{m}_4$ messages ($\tilde{\bm{D}}_{\tau - 1}$ and $\bar{\bm{B}}$) and one $\text{m}_5$ message ($\tilde{\bm{D}}_{\tau}$).

\paragraph{Messages for $S_\tau$:}
\begin{align*}
Q^*(S_\tau) &= \text{Cat}(S_\tau; \sigma(\mu_{S_\tau}^*))
\end{align*}
$$\mu_{S_\tau}^* = [\tau = 0] {\color{orange} \bar{\bm{D}}} + [\tau \neq 0] \sum_{k} {\color{orange} \tilde{\alpha}_k \bar{\bm{B}}[U_{\tau - 1}^k]\tilde{\bm{D}}_{\tau - 1}} + [\tau \leq t] {\color{purple} \bm{o}_\tau \cdot \bar{\bm{A}}} + [\tau \neq T] \sum_{k} {\color{purple} \tilde{\alpha}_k \tilde{\bm{D}}_{\tau + 1} \cdot \bar{\bm{B}}[U_\tau^k]}$$

\begin{figure}[H]
	\centering
	{\includegraphics[width=0.8\textwidth]{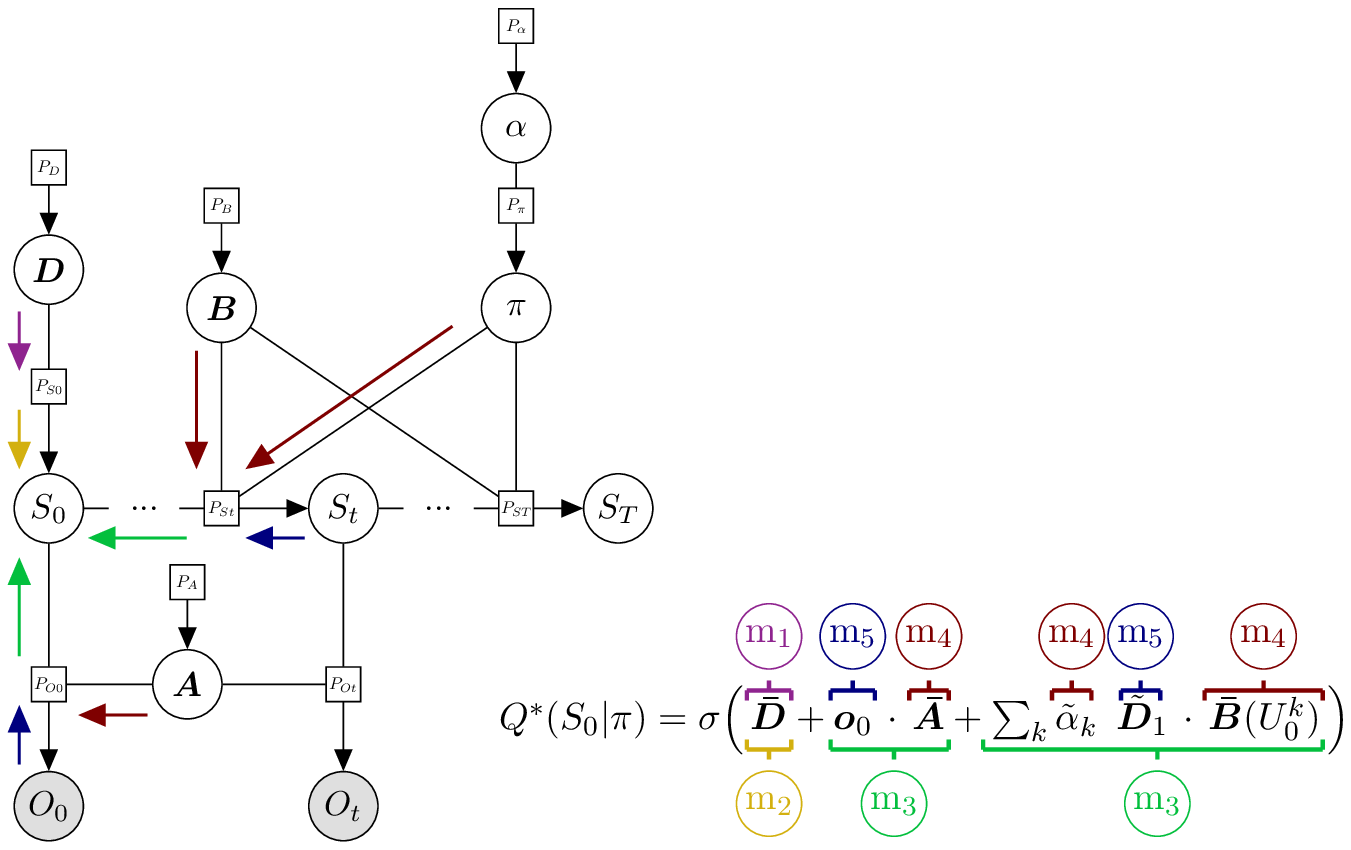}}
    \caption{
This figure illustrates the passing of messages required to update the posterior over $S_0$. The messages of type ${\color{Violet}\text{m}_1}$, ${\color{Yellow}\text{m}_2}$, ${\color{Green}\text{m}_3}$, ${\color{Red}\text{m}_4}$ and ${\color{Blue}\text{m}_5}$ come from the parent variables, parent factors, child factors, co-parent variables and child variables, respectively.}
    \label{fig:S0_msgs}
\end{figure}

To understand the above equation, we can consider two cases: $\tau = 0$ and $\tau \neq 0$. In the first case, $S_0$ only has one parent variable ($\bm{D}$), and $\mu_{S_0}(\bar{\bm{D}}) = \bar{\bm{D}}$ where $\bar{\bm{D}} = \langle u_{D}(\bm{D}) \rangle_{Q_D}$ is a message from a parent variable ($\text{m}_1$ message). In the second case, $S_\tau$ has three parent variables ($S_{\tau - 1}$, $\bm{B}$ and $\pi$), and $\mu_{S_\tau}(\tilde{\bm{D}}_{\tau - 1}, \bar{\bm{B}}, \tilde{\alpha}) = \sum_{k} \tilde{\alpha}_k \bar{\bm{B}}[U_\tau^k]\tilde{\bm{D}}_{\tau - 1}$ where $\tilde{\bm{D}}_{\tau - 1}$, $\bar{\bm{B}}$ and $\tilde{\alpha}$ are also $\text{m}_1$ messages. Let us now think about the child variable(s) of $S_\tau$. If $\tau \leq t$, then $S_\tau$ has a child variable from the likelihood mapping and $\mu_{O_\tau \rightarrow S_\tau}(\bm{o}_\tau, \bar{\bm{A}}) = \bm{o}_\tau \cdot \bar{\bm{A}}$, where $\bm{o}_\tau$ is a message from the child variable ($\text{m}_5$ message) and $\bar{\bm{A}}$ is a message from the co-parent variable ($\text{m}_4$ message). Additionally, if $\tau \neq T$, then $S_\tau$ receives a message from the future $\mu_{S_{\tau + 1} \rightarrow S_\tau}(\tilde{\alpha}_k, \tilde{\bm{D}}_{\tau + 1}, \bar{\bm{B}}) = \sum_{k} \tilde{\alpha}_k \tilde{\bm{D}}_{\tau + 1} \cdot \bar{\bm{B}}[U_\tau^k]$, where $\tilde{\alpha}_k$ and $\bar{\bm{B}}$ are $\text{m}_4$ messages and  $\tilde{\bm{D}}_{\tau + 1}$ is a $\text{m}_5$ message. Figure \ref{fig:S0_msgs} concludes this section with an illustration the message passing procedure for $S_0$.

\subsection{Messages vs update equations} \label{sec:msg_vs_up_eq}

In this section, we present a side by side comparison of the messages obtained using variational message passing and the update equations that underwrite belief updating in the active inference literature. Throughout this section, the messages will always be presented first, followed by the equivalent update equations. Let us start with the random variable $\bm{D}$:
\begin{align*}
Q^*(\bm{D}) &= \text{Dir}(\bm{D}; {\color{orange}d} + {\color{purple}\tilde{\bm{D}}_{0}})
\end{align*}
\begin{align*}
Q^*(\bm{D}) &= \text{Dir}(\bm{D}; d + \bm{s}_0)
\end{align*}
These two equations only differ in terms of labels, i.e. $\bm{s}_0$ and $\tilde{\bm{D}}_{0}$ conceptually represent the same quantity. Similarly, the updates of $\bm{A}$ are recovered up to a change of label:
\begin{align*}
Q^*(\bm{A}) &= \prod_i \text{Dir}(\bm{A}_{\bigcdot i}, \bm{a}_{\bigcdot i}) \quad \text{ where } \quad\bm{a} = {\color{orange}a} + \sum_{\tau = 0}^t {\color{purple}\bm{o}_\tau} \otimes {\color{purple}\tilde{\bm{D}}_\tau}\\
Q^*(\bm{A}) &= \prod_i \text{Dir}(\bm{A}_{\bigcdot i}, \bm{a}_{\bigcdot i}) \quad \text{ where } \quad \bm{a} = a + \sum_{\tau = 0}^t \bm{o}_\tau \otimes \bm{s}_\tau
\end{align*}
The update of $\bm{B}$ slightly differs from the messages obtained from variational message passing, which follows from the fact that we modified the variational distribution:
\begin{align*}
Q^*(\bm{B}) &= \prod_{u,i} \text{Dir}(\bm{B}[u]_{\bigcdot i}, \bm{b}[u]_{\bigcdot i}) \quad \text{ where } \quad \bm{b}[u] = {\color{orange}b[u]} + \sum_{(k,\tau) \in \Omega_u} {\color{purple}\tilde{\alpha}_k} {\color{purple}\tilde{\bm{D}}_{\tau}} \otimes {\color{purple}\tilde{\bm{D}}_{\tau - 1}}\\
Q^*(\bm{B}) &= \prod_{u,i} \text{Dir}(\bm{B}[u]_{\bigcdot i}, \bm{b}[u]_{\bigcdot i}) \,\quad \text{ where } \quad \bm{b}[u] = b[u] + \sum_{(k,\tau) \in \Omega_u} \bm{\pi}_k\bm{s}^k_{\tau} \otimes \bm{s}^k_{\tau - 1} 
\end{align*}
The only conceptual difference here is that $\bm{s}^k_{\tau}$ depended upon the policy, while $\tilde{\bm{D}}$ does not. Concerning $S_\tau$, we have re-arranged the update equation to highlight the similarity with the messages:
\begin{align*}
Q^*(S_\tau) &= \text{Cat}(S_\tau; \sigma (\mu_{S_\tau}^*))
\end{align*}
\vspace{-1.2cm}\\
\begin{align*}
\mu_{S_\tau}^* &= [\tau = 0] {\color{orange} \bar{\bm{D}}} + [\tau \neq 0] \sum_{k} {\color{orange} \tilde{\alpha}_k \bar{\bm{B}}[U_{\tau - 1}^k]\tilde{\bm{D}}_{\tau - 1}} + [\tau \leq t] {\color{purple} \bm{o}_\tau \cdot \bar{\bm{A}}} + [\tau \neq T] \sum_{k} {\color{purple} \tilde{\alpha}_k \tilde{\bm{D}}_{\tau + 1} \cdot \bar{\bm{B}}[U_\tau^k]}\\
\mu_{S_\tau}^* &= [\tau = 0]\bm{\bar{D}} + [\tau \neq 0]\,\quad \quad \quad \bm{\bar{B}}[U_{\tau - 1}^\pi] \bm{s}^\pi_{\tau - 1}\, + [\tau \leq t]  \bm{o}_{\tau} \cdot  \bm{\bar{A}} + [\tau \neq T]\quad \quad \,\,\,\,\,\,\, \bm{s}^\pi_{\tau + 1} \cdot \bm{\bar{B}}[U_{\tau}^\pi]
\end{align*}
There are two main differences here. First, as for $\bm{B}$, $\bm{s}^k_{\tau}$ is replaced by $\tilde{\bm{D}}$, which does not depend on the policies. Second, the past and future messages have an average over the policies, while the updates do not. Unsurprisingly, since we replaced $\gamma$ by $\alpha$ and changed the type of distributions, the updates are quite different:
\begin{align*}
Q^*(\alpha) &= \text{Dir}\Big(\alpha; {\color{orange}\theta} + {\color{purple}\tilde{\alpha}}\Big)\\
Q^*(\gamma) &= \Gamma \Big(\gamma; 1, \beta + \bm{G} \cdot (\bm{\pi} - \bm{\pi}_0)\Big)
\end{align*}
We conclude this section with the messages and updates of $\pi$, which are formally distinct. These differences come from the fact that we moved $\bm{G}$ from $P(\pi|\gamma)$ to $P(\alpha)$ and turned $P(\pi|\gamma)$ into a categorical distribution $P(\pi|\alpha)$:
\begin{align*}
Q^*(\pi) &= \text{Cat}(\pi; \alpha^*)
\end{align*}
\vspace{-1.2cm}\\
\begin{align*}
\alpha^* &= \sigma\Bigg(\,\quad{\color{orange}\bar{\alpha}}\,\quad + \sum_{\tau = 1}^T {\color{purple} \mathbb{F}_\tau}\Bigg) \,\, \text{ and } \,\, \mathbb{F}_\tau = \begin{bmatrix}
\langle \tilde{\bm{D}}_{\tau} \otimes \tilde{\bm{D}}_{\tau - 1} , \bar{\bm{B}}[U_{\tau - 1}^1] \rangle_F \\
...\\
\langle \tilde{\bm{D}}_{\tau} \otimes \tilde{\bm{D}}_{\tau - 1} , \bar{\bm{B}}[U_{\tau - 1}^{|\pi|}] \rangle_F
\end{bmatrix}\\
\alpha^* &= \sigma \Bigg(-\frac{1}{\bm{\beta}}\bm{G} + \sum_{\tau = 1}^T \mathbb{F}_\tau \Bigg) \,\, \text{ and } \,\, \mathbb{F}_\tau = \bm{s}^\pi_{\tau} \cdot \bm{\bar{B}}[U] \bm{s}^\pi_{\tau - 1}
\end{align*}

However, the general form of the updates remains unchanged with information coming from the parent through $\bar{\alpha}$ and $-\frac{1}{\bm{\beta}}\bm{G}$, and from each child through the summation over time steps.

\section{Conclusion}

The increasing use of active inference in neuroscience has cast many brain processes as Bayesian inference, the update equations of which can be thought of as a message passing procedure. The first goal of this paper was to present a complete overview of the active inference framework in discrete time and state space (Section \ref{sec:ai}) as well as a formal introduction to the variational message passing literature (Section \ref{sec:ai_vmp}). Then, we simplified the generative model and the variational distribution usually adopted in the active inference to derive a new set of update equations using the method of \citet{VMP_TUTO} --- and highlight the connection between active inference and variational message passing (Section \ref{sec:vmp-connection}).

We hope that the first few sections of this paper could be useful as an introduction to variational inference, Forney factor graphs, active inference or/and variational message passing. Section \ref{sec:vmp-connection} might also be of interest to researchers searching for a clear link between active inference and variational message passing or researchers seeking to derive the update equations of new generative models. Section \ref{sec:vmp-connection} explains why a fully factorised variational distribution simplifies the expected free energy in a way that precludes risk sensitive behaviour but preserves ambiguity avoidance. Finally, we note that this issue does not confound generative models implementing tree search.

One might ask why previous formulations of belief updating or message passing in active inference have not exploited the simplifications considered in the current paper. For example, using a Dirichlet distribution to parameterise Bayesian beliefs over policies --- or a fully factorised variational distribution that would simplify message passing. One answer is that much of the legacy literature in active inference is concerned with neuronal process theories and biological implementation. For example, the only reason a Gibbs form was used for the distribution over policies was to link the implicit temperature or sensitivity parameter to dopaminergic discharges. Similarly, the minimisation of variational free energy --- using a gradient descent to implement structured variational message passing --- was motivated by the need to cast belief updating in terms of differential equations that could be plausibly associated with neuronal dynamics (and accompanying electrophysiological responses to observations). However, if one frees oneself from the constraints of biological implementation, the repertoire of established schemes in machine learning and Bayesian statistics can, in principle, be leveraged to reproduce kinds of choice behaviour active inference is trying to explain and emulate. This paper has highlighted the putative usefulness of variational message passing under a rationalisation of generative models.

It is interesting to consider whether the simplified expected free-energy --- resulting from our message passing formulation of active inference --- can be linked in any sense to human behaviour, whether normative or pathological. In particular, the free-energy we have obtained reflects a very specific functional impoverishment. The full factorisation that is necessary for vanilla message passing precludes the ability to conditionalize the variational posterior on policies. This suggests a particular deficit in the ability to plan, and a blindness to future possibilities, the uncertainty associated with those possibilities and their potential to satisfy preferences. As a result, the agent's objective becomes to seek out unambiguous cues, with no concern for outcome. 

In fact, humans do exhibit patterns of behaviour that --- due to their repetitiveness --- seem to reflect a desire for high predictability. Additionally, some of these patterns do not seem obviously connected to rewarding or punishing outcomes. For example, those with autism can exhibit very stereotyped repetitive behaviour: hand flapping, hand clapping, rocking, etc \citep{GABRIELS2005}, which is often described as stimming \citep{Rajagopalan_2013_ICCV_Workshops}. These repetitive and ritualistic behaviours \citep{20001501751} suggest an objective to avoid exploration and the associated uncertainty.

This work naturally leads to future directions of research. For example, one could implement the new generative model proposed in this paper and compare its performance with the model presented in Section \ref{sec:ai}. Furthermore, additional research needs to be done to connect the original update equations of active inference to the cluster variational message passing literature. Much work has already been done on structured variational message passing; particularly relation to marginal message passing --- and its advantages over related approaches based upon Bethe free energy \citep{30019661350,ParrDec2019}. Another interesting direction of research would be to design new generative models that can tackle more complex tasks, such as playing Atari games, human-machine interaction using natural language and automatic structure learning. Partial answers to these directions of research have already been provided with the use of deep active inference \citep{DeepAIwithMCMC,Ueltzhoffer2018,9207382}, deep temporal models \citep{FRISTON2018486,10.3389/frai.2020.509354} and Bayesian model reduction \citep{BMR,curiosity_insight,sleep}. Nevertheless, we anticipate that additional work will pursue these avenues of research. Finally, one could also compare the update schemes under VMP to belief propagation \citep{BP_and_DC} or marginal message passing \citep{ParrDec2019}.


\acks{We would like to thank Karl Friston as well as the reviewers for their valuable feedback, which greatly improved the quality of the present paper.}

\appendix
\section*{Appendix A: Active Inference, KL Control and Reinforcement Learning.}

This appendix focuses on the relationship between Active Inference, KL Control and Reinforcement Learning (cf. \citet{dacosta2020relationship} and \citet{levine2018reinforcement} for more details). Let us restart with the expected free energy given by Equation \ref{eq:1}:
\begin{align*}
\bm{G}(\pi) \approx \sum_{\tau=t + 1}^T \underbrace{D_{\mathrm{KL}}[\overbrace{Q(O_\tau|\pi)}^{\text{expected outcomes}}||\overbrace{P(O_\tau)}^{\text{prior preferences}}]}_{\text{expected risk}}\,\, +\,\, \underbrace{\mathbb{E}_{Q(S_\tau|\pi)}[\text{H}[P(O_\tau | S_\tau)]]}_{\text{expected ambiguity}}.
\end{align*}

If the expected ambiguity is equal to zero, then the expected free energy reduces to the expected risk, which is the cost function minimised in the KL control literature. This highlights that active inference generalises KL control \citep{KL_CONTROL} by taking into account the ambiguity of the mapping between the hidden states and the observations. Active inference therefore selects policies leading to unambiguous states. Furthermore, the expected risk can be re-written as follows:
\begin{align*}
\text{expected risk} = D_{\mathrm{KL}}[Q(O_\tau|\pi)||P(O_\tau)] = \underbrace{\mathbb{E}_{Q(O_\tau|\pi)}[\ln Q(O_\tau|\pi)]}_{\text{negative entropy}} - \,\underbrace{\mathbb{E}_{Q(O_\tau|\pi)}[P(O_\tau)]}_{\text{expected rewards}}.
\end{align*}

If the negative entropy is zero, then the expected free energy reduces to the negative expected prior preference. Those preferences encode the notion of good outcomes, or equivalently, the notion of rewarding observations. This highlights why active inference can be thought of as a generalisation of reinforcement learning \citep{DeepRL}. Another view on the expected free energy is:
\begin{align}
\bm{G}(\tau, \pi) &= \overbrace{\mathbb{E}_{\tilde{Q}}[\ln Q(S_\tau|\pi) - \ln P(S_\tau | O_\tau, \pi)]}^{\text{(-ve) epistemic value}} - \overbrace{\mathbb{E}_{\tilde{Q}}[\ln P(O_\tau | \pi)]}^{\text{extrinsic value}},\label{eq:reward_vs_learning}
\end{align}
where $\tilde{Q} = P(O_\tau|S_\tau)Q(S_\tau)$. The extrinsic value is another term for expected prior preferences, which is equivalent to expected rewards in reinforcement learning. It is worth looking in more detail at the negative epistemic value (-EV), which differentiates the learning objectives of reinforcement learning and active inference:
\begin{align*}
-EV &= -\overbrace{-\mathbb{E}_{\tilde{Q}}[\ln Q(S_\tau|\pi) - \ln P(S_\tau | O_\tau, \pi)]}^{\text{epistemic value}}\\
\Leftrightarrow \quad \quad EV &= \underbrace{\mathbb{E}_{\tilde{Q}}[\ln P(S_\tau | O_\tau, \pi) - \ln Q(S_\tau|\pi)]}_{\text{mutual information between }S_\tau \text{ and } O_\tau}.
\end{align*}

Thus, the epistemic value is approximately equal to the mutual information between $S_\tau$ and $O_\tau$. The mutual information encodes the expected information gain over one variable by knowing the value of another. Therefore, the epistemic value tells us how knowing future observations reduces our uncertainty over future hidden states. The following should help to see that the epistemic value is approximately equal to the mutual information between $S_\tau$ and $O_\tau$:
\begin{align*}
I(S;O) &= \kl{P(S_\tau,O_\tau)}{P(S_\tau)P(O_\tau)}\\
&= \mathbb{E}_{P(S_\tau,O_\tau)}[\ln P(S_\tau|O_\tau) + \ln P(O_\tau) - \ln P(S_\tau) - \ln P(O_\tau)]\\
&= \mathbb{E}_{P(S_\tau,O_\tau)}[\ln P(S_\tau|O_\tau) - \ln P(S_\tau)].
\end{align*}

Intuitively, the more an observation tells us about future states, the more valuable this observation is. The negative epistemic value from equation \ref{eq:reward_vs_learning} directly reflects this intuition, and favours the policies with high mutual information. More importantly, equation \ref{eq:reward_vs_learning} allows the agent to compare the information gain and the reward on the same scale, i.e. using nats from information theory. This creates a sense in which an active inference agent deals optimally with the trade-off between exploration and exploitation.

\section*{Appendix B: Useful Properties.}

This appendix quickly reviews the properties used throughout this paper.

\paragraph{Product rule:}{$P(X,Y) = P(X|Y)P(Y)$,\\
where $X$ and $Y$ are random variables.}

\paragraph{Linearity of expectation:}{$\mathbb{E}_{P(Y)}[aY+b] = a\mathbb{E}_{P(Y)}[Y]+b$,\\
where $a$ and $b$ are constants, and $Y$ is a random variable.}

\paragraph{Expectation of a constant:}{$\mathbb{E}_{P(Y)}[a] = a$,\\
where $a$ is a constant, and $Y$ is a random variable}

\paragraph{Log property:}{$\ln(ab) = \ln(a) + \ln(b)$,\\
where $a$ and $b$ are real numbers}

\paragraph{Exponential product property:}{$\exp(a + b) = \exp(a)\exp(b)$,\\
where $a$ and $b$ are real numbers}

\paragraph{Exponential power property:}{$\exp(ab) = \exp(a)^b$,\\
where $a$ and $b$ are real numbers}

\section*{Appendix C: Definition and Justification of the Expected Free Energy.}

In this appendix, we focus on the definition of the expected free energy and the justification of Equation \ref{eq:1}. Another good resource on the subject is the ``expected free energy" appendix of \citet{TUTO_AI_RYAN}. For the sake of simplicity, we assume the following generative model and variational distribution:
$$P(O_{0:T},S_{0:T},\bm{B}|\pi) = P(\bm{B})P(S_0)\prod_{\tau = 1}^T P(S_\tau|S_{\tau - 1}, \bm{B}, \pi)\prod_{\tau = 0}^T P(O_\tau|S_\tau)$$
$$Q(S_{0:T},\bm{B}|\pi) = Q(\bm{B})\prod_{\tau = 0}^T Q(S_\tau|\pi).$$

Furthermore, we let $X = \{\bm{B},S_{0:T}\}$ denote the set of hidden variables of the model. Note that in this appendix, we restrict ourself to the hidden variables $X$ but new variables such as $\bm{A}$ and $\bm{D}$ can be added without changing the idea of the following derivation. Initially, the expected free energy was defined as the variational free energy conditioned on the policy, i.e.
$$\bm{G}(\pi) = \kl{Q(X|\pi)}{P(O_{0:t},X|\pi)}.$$

However, the above definition does not take into account that observations will be made in the future. To make up for this, the expected free energy can be extended as follows:
\begin{align} \label{eq:efe-app-c}
\bm{G}(\pi) = \mathbb{E}_{\tilde{Q}}\Big[\kl{Q(X|\pi)}{P(O_{0:T},X|\pi)}\Big] \text{ where } \tilde{Q} \delequal \tilde{Q}(O_{t+1:T}|\pi).
\end{align}

Since the future observations ($O_{t+1:T}$) have not been made yet, we need to predict what they could look like. This prediction relies on a predictive distribution $\tilde{Q}(O_{t+1:T}|\pi)$ that encodes our best guess about future outcomes, and is generally defined as follows:
\begin{align*}
\tilde{Q}(O_{t+1:T}|\pi) \delequal \prod_{\tau = t + 1}^T \tilde{Q}(O_{\tau}|\pi),
\end{align*}
\begin{align*}
\text{where } \,\,\,\, \tilde{Q}(O_{\tau}|\pi) \delequal \sum_{S_\tau} \tilde{Q}(O_{\tau}, S_{\tau}|\pi) \,\,\,\, \text{ and } \,\,\,\, \tilde{Q}(O_{\tau}, S_{\tau}|\pi) \delequal P(O_\tau|S_\tau)Q(S_\tau|\pi).
\end{align*}

Note that the definition of $\tilde{Q}(O_{t+1:T}|\pi)$ assumes independence between time steps and $\tilde{Q}(O_{\tau}|\pi)$ is obtained by marginalisation of $\tilde{Q}(O_{\tau}, S_{\tau}|\pi)$. By recalling the definition of the generative model as well as the definition of the variational distribution, we obtain the following from Equation \ref{eq:efe-app-c}:
\begin{align*}
\bm{G}(\pi) &= \mathbb{E}_{\tilde{Q}}\Big[\kl{Q(S_{0:T},\bm{B}|\pi)}{P(O_{0:T},S_{0:T},\bm{B}|\pi)}\Big]\\
&= \kl{Q(\bm{B})}{P(\bm{B})} + \kl{Q(S_0|\pi)}{P(S_0)} {\color{white}\sum}\\
&\quad + \,\, \sum_{\tau = 1}^t \,\,\, \mathbb{E}_{Q(S_{\tau - 1}, \bm{B}|\pi)}\Big[ \kl{Q(S_\tau|\pi)}{P(S_\tau|S_{\tau - 1}, \bm{B}, \pi)}\Big]\\
&\quad + \,\, \sum_{\tau = 0}^t \,\,\, \mathbb{E}_{Q(S_\tau|\pi)}\Big[\text{H}[P(O_\tau|S_\tau)]\Big]\\
&\quad + \sum_{\tau = t+1}^T \mathbb{E}_{Q(S_{\tau - 1}, \bm{B}|\pi)}\Big[ \kl{Q(S_\tau|\pi)}{P(S_\tau|S_{\tau - 1}, \bm{B}, \pi)}\Big] + \mathbb{E}_{Q(S_\tau|\pi)}\Big[\text{H}[P(O_\tau|S_\tau)]\Big].
\end{align*}

It must now be mentioned that the policy does not have much of an impact on the past and current hidden states ($S_{0:t}$). The terms relying on those states are then removed from the expected free energy to avoid unnecessary computational costs. Additionally, the divergence between $Q(\bm{B})$ and $P(\bm{B})$ does not depend on the policy and can be safely ignored, leading to: 
\begin{align}\label{eq:efe-app-c-2}
\bm{G}(\pi) = \sum_{\tau = t + 1}^T \bm{G}(\pi,\tau)
\end{align}
where: 
$$\bm{G}(\pi,\tau) \delequal \mathbb{E}_{Q(S_{\tau - 1},\bm{B}|\pi)}\Big[ \kl{Q(S_\tau|\pi)}{P(S_\tau|S_{\tau - 1},\bm{B}, \pi)}\Big] + \mathbb{E}_{Q(S_\tau|\pi)}\Big[\text{H}[P(O_\tau|S_\tau)]\Big].$$

We now focus on $\bm{G}(\pi,\tau)$ to bridge the gap between Equations \ref{eq:1} and \ref{eq:efe-app-c-2}. First, we merge the two terms of the above equation together:
\begin{align*}
\bm{G}(\pi,\tau) \delequal \mathbb{E}_{P(O_\tau|S_\tau)Q(S_{\tau},S_{\tau - 1},\bm{B}|\pi)}\Big[ \ln Q(S_\tau|\pi) - \ln P(O_\tau, S_\tau|S_{\tau - 1},\bm{B}, \pi) \Big].
\end{align*}

Then, we break the second term within the expectation using the product rule. Additionally, we realise that the following equation can be obtained from the product rule:
$$P(O_\tau | S_{\tau - 1}, \bm{B}, \pi) = \frac{P(O_\tau, S_{\tau - 1}, \bm{B}, \pi)}{P(S_{\tau - 1}, \bm{B}, \pi)} = \frac{P(S_{\tau - 1}, \bm{B}, \pi | O_\tau)}{P(S_{\tau - 1}, \bm{B}, \pi)} P(O_\tau) \approx P(O_\tau),$$
where we assumed that the fraction is equal to one. Doing this assumption means that the observation $O_\tau$ brings us very little information, i.e. the posterior is close to the prior. Using the above result we get:
\begin{align*}
\bm{G}(\pi,\tau) &= \mathbb{E}\Big[ \ln Q(S_\tau|\pi) - \ln P(S_\tau|O_\tau,S_{\tau - 1},\bm{B}, \pi) - \ln P(O_\tau|S_{\tau - 1},\bm{B}, \pi) \Big]\\
&\approx \mathbb{E}\Big[ \ln Q(S_\tau|\pi) - \ln P(S_\tau|O_\tau,S_{\tau - 1},\bm{B}, \pi) - \ln P(O_\tau) \Big],
\end{align*}
where the expectation is still over $P(O_\tau|S_\tau)Q(S_{\tau},S_{\tau - 1},\bm{B}|\pi)$. Then, we uses Bayes theorem on the second term, the fact that $(O_\tau \indep{}{S_{\tau - 1},\bm{B}, \pi})|S_\tau$ and the log properties to get:
\begin{align*}
\bm{G}(\pi,\tau) &= \mathbb{E}\Big[ \ln Q(S_\tau|\pi) - \ln P(S_\tau|O_\tau,S_{\tau - 1},\bm{B}, \pi) - \ln P(O_\tau) \Big]\\
&= \mathbb{E}\Big[ \ln Q(S_\tau|\pi) - \ln \frac{P(O_\tau|S_{\tau},S_{\tau - 1},\bm{B}, \pi)P(S_\tau|S_{\tau - 1},\bm{B}, \pi)}{P(O_\tau|S_{\tau - 1},\bm{B}, \pi)} - \ln P(O_\tau) \Big]\\
&\approx \mathbb{E} \Big[ \ln Q(S_\tau|\pi) - \ln \frac{P(O_\tau|S_{\tau})Q(S_\tau|\pi)}{Q(O_\tau|\pi)} - \ln P(O_\tau) \Big]\\
&= \mathbb{E} \Big[ \ln Q(O_\tau|\pi) - \ln P(O_\tau) - \ln P(O_\tau|S_{\tau}) \Big],
\end{align*}
where we assumed that $P(S_\tau|S_{\tau - 1},\bm{B}, \pi) \approx Q(S_\tau|\pi)$ and $P(O_\tau|S_{\tau - 1},\bm{B}, \pi) \approx Q(O_\tau|\pi)$. The first assumption can be supported by the variational free energy (VFE) decomposition in term of accuracy and complexity. Indeed, the VFE penalises the divergence between $Q(S_\tau|\pi)$ and $P(S_\tau|S_{\tau - 1},\bm{B}, \pi)$. The second assumption can be supported as follows:
\begin{align*}
P(O_\tau|S_{\tau - 1}, \bm{B}, \pi) &= \sum_{S_\tau} P(O_\tau, S_\tau|S_{\tau - 1}, \bm{B}, \pi)\\
&\approx \sum_{S_\tau} Q(O_\tau, S_\tau|\pi)\\
&= Q(O_\tau|\pi).
\end{align*}
Assuming that the posterior $P(O_\tau, S_\tau|S_{\tau - 1}, \bm{B}, \pi)$ can be approximated by $Q(O_\tau, S_\tau|\pi)$. The last step relies on the linearity of expectation and the expectation of a constant, leading to the final result:
\begin{align*}
\bm{G}(\pi,\tau) &= \kl{Q(O_\tau|\pi)}{P(O_\tau)} + \mathbb{E}_{Q(S_\tau|\pi)} \Big[ \text{H} [P(O_\tau|S_{\tau})] \Big].
\end{align*}

\section*{Appendix D: The simplest generative model.}

\begin{figure}[H]
	\centering
	{\includegraphics[width=0.6\textwidth]{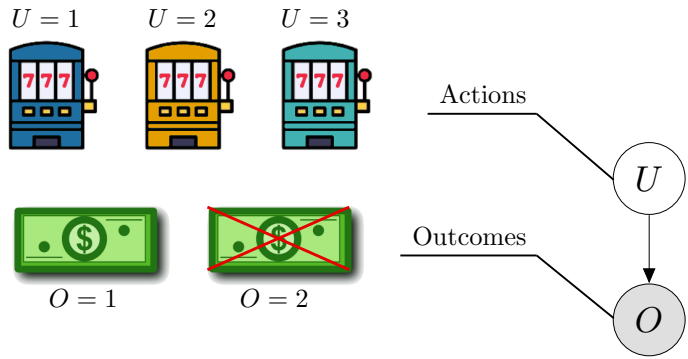}}
  \caption{This figure illustrates the 3-armed bandit problem and the generative model used by the agent. Three slot machines are available to the agent and each machine has a different probability of producing a reward. Additionally, there are two possible outcomes when pulling a lever, the agent either wins plenty of money or gets nothing. The generative model is composed of two nodes representing the possible outcomes and actions. Finally, the agent's goal is to maximize the rewards obtained, by picking the best strategy.}
   \label{fig:bandit}
\end{figure}

This appendix provides the reader with the smallest generative model that can be considered as an active inference agent and aims to solve the k-armed bandit problem. As shown in Figure \ref{fig:bandit}, this problem is composed of k slot machines or equivalently k actions that the agent can perform. Each machine has a different probability of producing a reward and the agent must chose the action to perform to maximize the rewards obtained. The agent only observes either a reward or a punishment after the execution of an action. Additional information related to the usage of active inference in the context of the multi-arms bandit (MAB) task can be found in \citep{markovic2021empirical} where active inference was compared to other major algorithms for solving MABs such as UCB sampling and Thompson sampling.

To solve the bandit problem using active inference, the first step is to create the generative model that encodes the agent's beliefs of the environment. Two random variables are used for this purpose, $O$ represents the possible outcomes and $U$ the available actions. Furthermore, $P(O|U)$ determines how the observation depends on the action performed by the agent, and $P(U)$ encodes any prior preference over the available actions. More precisely, $P(O|U)$ and $P(U)$ are categorical distributions defined as follows:
\begin{align*}
P(O = i|U = j) = A_{ij} \quad\quad\text{and}\quad\quad P(U = j) = a_j,
\end{align*}
where $A_{ij}$ defines the probability of the i-th outcome given that the j-th action is performed, and $a_j$ encodes the prior over the j-th action. Note that even if the active inference framework provides a way to learn the matrix $\bm{A}$, this section assumes that it is given to the agent. The next step is to pick an inference method to compute the posterior over the hidden state $U$. This section keeps things simple and uses Bayes theorem: 
\begin{align*}
P(U = j|O = 1) = \frac{P(O = 1|U = j)P(U = j)}{P(O = 1)} = \frac{P(O = 1|U = j)P(U = j)}{\sum_k P(O = 1|U = k)P(U = k)} = \frac{A_{1j}a_j}{\sum_k A_{1k}a_k},
\end{align*}
where the definition of the generative model has been used in the last step and we conditioned on $O = 1$ to infer the action that is more likely to be rewarding. At this point, it is possible to act in our environment either by sampling the next action to perform from the posterior $P(U|O = 1)$ or by picking the action with the highest posterior probability. Additionally, the posterior can be reused as an empirical prior for the next time step as follows:
\begin{align*}
P(U = j) \leftarrow P(U = j|O = 1) = \frac{A_{1j}a_j}{\sum_k A_{1k}a_k}.
\end{align*}

This simple example does not capture the entire theoretical power of the active inference framework. Nevertheless, it illustrates four important concepts related to the design and use of an active inference agent, namely, the design of a generative model, the inference of the latent variable(s), the action selection process, and the use of the posterior as an empirical prior.

\section*{Appendix E: Possible future research.}

In this appendix, we propose future research directions aiming to understand the relationship between $P(\pi|\gamma)$ and $P(\pi|\alpha)$. The first direction relies on the following link between Dirichlet and gamma distributions. If we let $X_1$,$...$,$X_{k}$ be mutually independent random variables, each having a gamma distribution with parameters $\theta_i$ for $i = 1, ... , k$ and if we define $Y_i = \frac{X_i}{X_{1} + ... + X_{k}}$ for $i = 1, ..., k$, then $(Y_1, ..., Y_k) \sim \text{Dir}(\theta_1, ..., \theta_k)$. This naturally leads to the hypothesis that the new generative model might be a generalisation of the old generative model when all $\theta_i$ are equal.

Another interesting fact that could be studied in more detail comes from studying the variance of the Dirichlet distribution. Recall that the variance of the random variable $Y_i$ is given by:
$$\text{Var}[Y_i] = \frac{\tilde{\theta}_i(1 - \tilde{\theta}_i)}{\theta_0 + 1},$$
where $\tilde{\theta}_i = \frac{\theta_i}{\theta_0}$ and $\theta_0 = \sum_{j = 1}^k \theta_j$. If we stick to our definition of $\theta$, i.e. $\theta_j = c - \bm{G}_j$ with $c = \overrightarrow{c}_j \,\, \forall j$, then we can study how the variance of $Y_j$ behaves as $c$ goes to infinity. Let us begin with:
$$\lim_{c \rightarrow +\infty} \tilde{\theta}_i = \lim_{c \rightarrow +\infty} \frac{\theta_i}{ \sum_{j = 1}^k \theta_j} = \lim_{c \rightarrow +\infty} \frac{c - \bm{G}_i}{ \sum_{j = 1}^k c - \bm{G}_j} = \lim_{c \rightarrow +\infty} \frac{c - \bm{G}_i}{ kc - \sum_{j = 1}^k \bm{G}_j} = \lim_{c \rightarrow +\infty} \frac{c}{ kc} = \frac{1}{k},$$
where we note that $\bm{G}_i$ and $\sum_{j = 1}^k \bm{G}_j$ become negligible as $c \rightarrow +\infty$. Returning to the limit of the variance: 
$$\lim_{c \rightarrow +\infty} \text{Var}[Y_i] = \lim_{c \rightarrow +\infty} \frac{\tilde{\theta}(1 - \tilde{\theta})}{\theta_0 + 1} = \lim_{c \rightarrow +\infty} \frac{\tilde{\theta}(1 - \tilde{\theta})}{\Big( \sum_{j = 1}^k c - \bm{G}_j \Big) + 1} = 0,$$
where we used the fact that $\tilde{\theta}_i$ tends towards $\frac{1}{k}$ (i.e. a constant w.r.t $c$) and therefore the variance is only influenced by the $c$ in the denominator, which tends towards $+ \infty$. Additionally, from the definition of the mode of the Dirichlet, we see that as $c \rightarrow + \infty$ then the mode of the distribution tends towards the centre of the simplex because the $\bm{G}_i$ becomes negligible, i.e. 
$$\lim_{c \rightarrow + \infty} m_{\alpha} = \begin{bmatrix}
\frac{1}{k} & \hdots & \frac{1}{k}
\end{bmatrix}.$$
Combining the behaviour of the variance and the mode as $c \rightarrow + \infty$, we see that as $c$ increases the prior becomes more and more compact around the centre of the simplex. In other words, the policy selection becomes more and more stochastic as $c$ increases. This is not without recalling the role of $\gamma$ as highlighted previously in the caption of Figure \ref{fig:full_GM}.

\section*{Appendix F: Messages for B.}

In this appendix, we provide the derivation of the messages for $\bm{B}$, which relies on the conjugacy between a categorical and a Dirichlet distribution. Let us start with the definition of $P(\bm{B};b)$, which is a product of Dirichlet distributions that can be written in the following form:
\begin{align}\label{eq:B:0}
\ln P(\bm{B};b) &= \ln \prod_{i,u} P(\bm{B}[u]_{\bigcdot i};b[u]_{\bigcdot i}) = \sum_{i,u} \ln \text{Dir}(\bm{B}[u]_{\bigcdot i};b[u]_{\bigcdot i})\nonumber\\
&= \sum_{i,u} \underbrace{\begin{bmatrix}
b[u]_{1i} - 1\\
...\\
b[u]_{|S|i} - 1
\end{bmatrix} \cdot
\begin{bmatrix}
\ln \bm{B}[u]_{1i}\\
...\\
\ln \bm{B}[u]_{|S|i}
\end{bmatrix}
- \ln B(b[u]_{\bigcdot i})}_{\text{Logartithm of Dirichlet}}\nonumber\\
&= \underbrace{\begin{bmatrix}
b[1]_{11} - 1\\
...\\
b[|U|]_{|S||S|} - 1
\end{bmatrix}}_{\mu_{B}(b)} \cdot
\underbrace{\begin{bmatrix}
\ln \bm{B}[1]_{11}\\
...\\
\ln \bm{B}[|U|]_{|S||S|}
\end{bmatrix}}_{u_{B}(\bm{B})}
- \underbrace{\sum_{i,u} \ln B(b[u]_{\bigcdot i})}_{z_{B}(b)},
\end{align}
where $|U|$ is the number of possible actions. Let $\llbracket a, b \rrbracket$ denotes all the natural numbers between $a$ and $b$ (inclusive). The random matrix $\bm{B}[u]$ has one child $S_\tau$ for each time step $\tau \in \llbracket 1, T \rrbracket$ where action $u$ has been predicted by the $m$-th policy, and its probability mass function is given by Equation \ref{eq:A:11}. Similarly, the probability mass function of $S_{\tau - 1}$ is obtained from Equation \ref{eq:A:11} by decreasing all indexes $\tau$ by one. The first step requires us to re-write Equation \ref{eq:A:11} as a function of $u_{B}(\bm{B})$. This can be done by using the definition of the dot product and re-arranging to obtain:
\begin{align}\label{eq:B:2}
\ln P(S_\tau = k | \bm{B}, S_{\tau - 1} = l, \pi = m) = \underbrace{\begin{bmatrix}
\sum_k [\pi = k][U_{\tau - 1}^k = 1][S_{\tau - 1} = 1][S_\tau = 1]\\
...\\
\sum_k [\pi = k][U_{\tau - 1}^k = |U|][S_{\tau - 1} = |S|][S_\tau = |S|]
\end{bmatrix}}_{\mu_{S_\tau \rightarrow \bm{B}}(S_{\tau},S_{\tau - 1},\pi)} \cdot u_{B}(\bm{B}).
\end{align}

The second step aims to substitute Equations \ref{eq:B:0} and \ref{eq:B:2} within the variational message passing equation (\ref{eq:VMP_42}), i.e.
\begin{align*}
\ln Q^*(\bm{B}) &= \Big\langle \begin{bmatrix}
{\color{orange}b[1]_{11} - 1}\\
...\\
{\color{orange}b[|U|]_{|S||S|} - 1}
\end{bmatrix} \cdot u_{B}(\bm{B}) \Big\rangle\\
&+ \sum_{\tau = 1}^{T} \Big\langle \begin{bmatrix}
{\color{purple}\sum_k [\pi = k][U_{\tau - 1}^k = 1][S_{\tau - 1} = 1][S_\tau = 1]}\\
...\\
{\color{purple}\sum_k [\pi = k][U_{\tau - 1}^k = |U|][S_{\tau - 1} = |S|][S_\tau = |S|]}
\end{bmatrix} \cdot u_{B}(\bm{B}) \Big\rangle + \text{Const},
\end{align*}
where $\langle \bigcdot \rangle$ refers to $\langle \bigcdot \rangle_{\sim Q_{\bm{B}}}$. Note that in the above Equation, $b[u]_{ij}$ are hyper parameters that can therefore be considered as constants with respect to the expectation $\langle \bigcdot \rangle_{\sim Q_{\bm{B}}}$. The third step builds on this insight, by pulling the summation over time steps inside the vector, factorising by $u_{\bm{B}}(\bm{B})$, using the linearity of expectation and by taking the exponential of both sides to obtain:
\begin{align*}
Q^*(\bm{B}) &\propto \exp \{ \mu_{B}^* \cdot u_{B}(\bm{B}) \}
\end{align*}
\begin{align*}
\mu_{B}^* &= \begin{bmatrix}
{\color{orange}b[1]_{11} - 1} + \sum_{k,\tau} \langle {\color{purple} [\pi = k][U_{\tau - 1}^k = 1][S_{\tau - 1} = 1][S_\tau = 1]} \rangle\\
...\\
{\color{orange}b[|U|]_{|S||S|} - 1} + \sum_{k,\tau} \langle {\color{purple} [\pi = k][U_{\tau - 1}^k = |U|][S_{\tau - 1} = |S|][S_\tau = |S|]} \rangle
\end{bmatrix}.
\end{align*}

By looking at Equations \ref{eq:A:11}, one can see that $\langle [S_\tau = i] \rangle$ and $\langle [S_{\tau - 1} = j] \rangle$ are the i-th and j-th elements of the vector $\langle u_{S_\tau}(S_\tau)\rangle$ and $\langle u_{S_\tau - 1}(S_\tau - 1)\rangle$, respectively. Furthermore, because $P(\pi)$ is a categorical distribution it can be expressed as:
\begin{align}\label{eq:pi:def}
P(\pi|\alpha) = \underbrace{\begin{bmatrix}
\ln \alpha_1 \\
...\\
\ln \alpha_{|\pi|}
\end{bmatrix}}_{\mu_\pi(\alpha)} \cdot \underbrace{\begin{bmatrix}
[\pi = 1]\\
...\\
[\pi = |\pi|]
\end{bmatrix}}_{u_\pi(\pi)},
\end{align}
where $|\pi|$ is the number of policies. The above equation highlights that $\langle [\pi = k] \rangle$ is the k-th element of $\langle u_\pi(\pi) \rangle$. Using those three insights, we proceed with the following re-parameterization (i.e. the fourth step):
\begin{align}\label{eq:B:3}
\mu_{B}^* = \begin{bmatrix}
{\color{orange}b[1]_{11} - 1} + \sum_{k,\tau} {\color{purple} [U_{\tau - 1}^k = 1]} \langle {\color{purple} u_\pi(\pi)} \rangle_k \langle {\color{purple}u_{S_\tau}(S_\tau)} \rangle_1\langle {\color{purple}u_{S_\tau - 1}(S_\tau - 1)} \rangle_1\\
...\\
{\color{orange}b[|U|]_{|S||S|} - 1} + \sum_{k,\tau} {\color{purple} [U_{\tau - 1}^k = |U|]} \langle {\color{purple} u_\pi(\pi)}\rangle_k \langle {\color{purple}u_{S_\tau}(S_\tau)}\rangle_{|S|}\langle {\color{purple}u_{S_\tau - 1}(S_\tau - 1)}\rangle_{|S|}
\end{bmatrix},
\end{align}
where we focused on the optimal parameters because the rest remains unchanged. The last step consists of computing the expectation of $\langle u_{S_\tau - 1}(S_\tau - 1) \rangle_i$, $\langle u_{S_\tau}(S_\tau) \rangle_j$, and $\langle u_\pi(\pi) \rangle_k$ for all $i$, $j$ and $k$:

\begin{itemize}
\item $\langle u_{S_\tau - 1}(S_\tau - 1) \rangle_i = \langle [S_\tau - 1 = i] \rangle = \tilde{\bm{D}}_{(\tau - 1)i}$
\item $\langle u_{S_\tau}(S_\tau) \rangle_j = \langle [S_\tau = j] \rangle = \tilde{\bm{D}}_{\tau j}$
\item $\langle u_\pi(\pi) \rangle_k = \langle [\pi = k] \rangle = \tilde{\alpha}_{k}$
\end{itemize}

One last thing we need to look at is the interaction between the summation and the indicator function in the i-th line of Equation \ref{eq:B:3}. Indeed, the sum iterates over all time steps $\tau$ and all policies $k$, but the indicator function filters out all elements where the k-th policy does not predict the i-th action at time $\tau - 1$. Building on this insight, we can now substitute the above results in Equation \ref{eq:B:3}:
\begin{align*}
Q^*(\bm{B}) &\propto \exp \Bigg\{ \begin{bmatrix}
{\color{orange}b[1]_{11} - 1} + \sum_{(k,\tau) \in \Omega_1} {\color{purple}\tilde{\alpha}_{k}\tilde{\bm{D}}_{\tau 1}\tilde{\bm{D}}_{(\tau - 1)1}}\\
...\\
{\color{orange}b[|U|]_{|S||S|} - 1} + \sum_{(k,\tau) \in \Omega_{|U|}} {\color{purple}\tilde{\alpha}_{k}\tilde{\bm{D}}_{\tau |S|}\tilde{\bm{D}}_{(\tau - 1)|S|}}
\end{bmatrix}
\cdot u_{B}(\bm{B}) \Bigg\}.
\end{align*}

Finally, one can recognise in the above equation the logarithm of a product of Dirichlet distributions written into their exponential form, i.e.
\begin{align*}
Q^*(\bm{B}) &= \prod_{u,i} \text{Dir}(\bm{B}[u]_{\bigcdot i}, \bm{b}[u]_{\bigcdot i}) \quad \text{ where } \quad \bm{b}[u] = {\color{orange}b[u]} + \sum_{(k,\tau) \in \Omega_u} {\color{purple}\tilde{\alpha}_k} {\color{purple}\tilde{\bm{D}}_{\tau}} \otimes {\color{purple}\tilde{\bm{D}}_{\tau - 1}}.
\end{align*}

\section*{Appendix G: Messages for $S_\tau$.}

This appendix shows how to derive the messages for $S_\tau$ for all time steps. We will use Equations \ref{eq:D:1} and \ref{eq:A:11} that describe $P(S_0|\bm{D})$ and $P(S_\tau|S_{\tau - 1},\bm{B},\pi)$ as a function of $u_{S_\tau}(S_\tau)$. The first step requires us to re-arrange Equation \ref{eq:A:1} and $P(S_{\tau+1}|S_\tau,\bm{B},\pi)$ as a functions of $u_{S_\tau}(S_\tau)$, where $P(S_{\tau+1}|S_\tau,\bm{B},\pi)$ is obtained by adding one to all instances of $\tau$ in Equation \ref{eq:A:11}. Those two re-arrangements lead to the following results:
\begin{align}\label{eq:S0:000}
\ln P(O_\tau = k | \bm{A}, S_\tau = l) &= \begin{bmatrix}
\sum_i [O_\tau = i]\ln \bm{A}_{i 1}\\
...\\
\sum_i [O_\tau = i]\ln \bm{A}_{i |S|}
\end{bmatrix} \cdot
\underbrace{\begin{bmatrix}
[S_{\tau} = 1]\\
...\\
[S_{\tau} = |S|]
\end{bmatrix}}_{u_{S_\tau}(S_\tau)}
\end{align}

\begin{align}\label{eq:S0:111}
\ln P(S_{\tau + 1} = k | \bm{B}, S_\tau = l, \pi = m) &= \begin{bmatrix}
\sum_{j,k,u} [\pi = k][U_\tau^k = u][S_{\tau + 1} = j]\ln \bm{B}[u]_{1 j}\\
...\\
\sum_{j,k,u} [\pi = k][U_\tau^k = u][S_{\tau + 1} = j]\ln \bm{B}[u]_{|S| j}
\end{bmatrix} \cdot u_{S_\tau}(S_\tau).
\end{align}

For the second step, we need to substitute Equations \ref{eq:D:1}, \ref{eq:A:11}, \ref{eq:S0:000} and \ref{eq:S0:111} into the variational message passing equation. If $\tau = 0$, the parent message will come from the prior (i.e. Equation \ref{eq:D:1}) otherwise from the past (i.e. Equation \ref{eq:A:11}). Also, for all time steps such that $\tau \leq t$ there is a message from the likelihood mapping (i.e. Equation \ref{eq:S0:000}) and for all time steps except $\tau = T$ there is a message from the future (i.e. Equation \ref{eq:S0:111}). Putting everything together we obtain:
\begin{align*}
\ln Q^*(S_\tau) &= \Big\langle [\tau = 0] \begin{bmatrix}
{\color{orange}\ln \bm{D}_1}\\
...\\
{\color{orange}\ln \bm{D}_{|S|}}
\end{bmatrix} \cdot u_{S_\tau}(S_\tau) \Big\rangle \\
&+ \Big\langle [\tau \neq 0] \begin{bmatrix}
\sum_{j,k,u} {\color{orange}[\pi = k][U_{\tau - 1}^k = u][S_{\tau - 1} = j]\ln \bm{B}[u]_{1 j}}\\
...\\
\sum_{j,k,u} {\color{orange}[\pi = k][U_{\tau - 1}^k = u][S_{\tau - 1} = j]\ln \bm{B}[u]_{|S| j}}
\end{bmatrix} \cdot u_{S_\tau}(S_\tau) \Big\rangle \\
&+ \Big\langle [\tau \leq t] \begin{bmatrix}
\sum_i {\color{purple}[O_\tau = i]\ln \bm{A}_{i 1}}\\
...\\
\sum_i {\color{purple}[O_\tau = i]\ln \bm{A}_{i |S|}}
\end{bmatrix} \cdot u_{S_\tau}(S_\tau) \Big\rangle \\
&+ \Big\langle [\tau \neq T] \begin{bmatrix}
\sum_{j,k,u} {\color{purple} [\pi = k][U_\tau^k = u][S_{\tau + 1} = j]\ln \bm{B}[u]_{1 j}}\\
...\\
\sum_{j,k,u} {\color{purple} [\pi = k][U_\tau^k = u][S_{\tau + 1} = j]\ln \bm{B}[u]_{|S| j}}
\end{bmatrix} \cdot u_{S_\tau}(S_\tau) \Big\rangle\\
&+ \text{Const}.
\end{align*}

The third step requires us to factorise by $u_{S_\tau}(S_\tau)$, use the linearity of expectation and take the exponential of both sides:
\begin{align} \label{eq:S0:222}
Q^*(S_\tau) &\propto \exp \Bigg\{ \Big[ [\tau = 0] \mu_1^* + [\tau \neq 0] \mu_2^* + [\tau \leq t] \mu_3^* + [\tau \neq T] \mu_4^* \Big] \cdot u_{S_\tau}(S_\tau) \Bigg\},
\end{align}
where:
\begin{align*}
\mu_1^* = \begin{bmatrix}
\langle {\color{orange} \ln \bm{D}_1} \rangle\\
...\\
\langle {\color{orange} \ln \bm{D}_{|S|}} \rangle
\end{bmatrix}
\end{align*}
\begin{align*}
\mu_2^* = \begin{bmatrix}
\sum_{j,k,u} {\color{orange} [U_{\tau - 1}^k = u]} \langle {\color{orange} [\pi = k]} \rangle \langle {\color{orange} [S_{\tau - 1} = j]} \rangle \langle {\color{orange} \ln \bm{B}[u]_{1 j}} \rangle \\
...\\
\sum_{j,k,u} {\color{orange} [U_{\tau - 1}^k = u]} \langle {\color{orange} [\pi = k]}\rangle \langle {\color{orange} [S_{\tau - 1} = j]} \rangle \langle {\color{orange} \ln \bm{B}[u]_{|S| j}} \rangle 
\end{bmatrix}
\end{align*}
\begin{align*}
\mu_3^* = \begin{bmatrix}
\sum_i {\color{purple} [O_\tau = i]}\langle {\color{purple} \ln \bm{A}_{i 1}} \rangle \\
...\\
\sum_i {\color{purple} [O_\tau = i]}\langle {\color{purple} \ln \bm{A}_{i |S|}} \rangle 
\end{bmatrix}
\end{align*}
\begin{align*}
\mu_4^* = \begin{bmatrix}
\sum_{j,k,u} {\color{purple} [U_\tau^k = u]} \langle {\color{purple} [\pi = k]} \rangle \langle {\color{purple} [S_{\tau + 1} = j]}\rangle \langle {\color{purple} \ln \bm{B}[u]_{1 j}} \rangle\\
...\\
\sum_{j,k,u} {\color{purple} [U_\tau^k = u] } \langle {\color{purple} [\pi = k]} \rangle \langle {\color{purple} [S_{\tau + 1} = j]}\rangle \langle {\color{purple} \ln \bm{B}[u]_{|S| j}} \rangle
\end{bmatrix}.
\end{align*}

The fourth step is the re-parameterization relying on the fact that $\langle \ln \bm{D}_i \rangle$, $\langle [\pi = j] \rangle$, $\langle [S_{\tau - 1} = k] \rangle$, $\langle \ln \bm{B}[l]_{m n} \rangle$, $\langle \ln \bm{A}_{o p} \rangle$ and $\langle [S_{\tau + 1} = q] \rangle$ are elements of $\langle u_{D}(\bm{D})\rangle$, $\langle u_{\pi}(\pi)\rangle$, $\langle u_{S_{\tau - 1}}(S_{\tau - 1}) \rangle$, $\langle u_{B}(\bm{B}) \rangle$, $\langle u_{A}(\bm{A}) \rangle$ and $\langle u_{S_{\tau + 1}}(S_{\tau + 1})\rangle$, respectively. Focusing on the $\mu_i^*$ because the rest remains unchanged, the result of the the re-parameterisation is:

\begin{align*}
\mu_1^* = \begin{bmatrix}
\langle {\color{orange} u_{D}(\bm{D})} \rangle_1\\
...\\
\langle {\color{orange} u_{D}(\bm{D})} \rangle_{|S|}
\end{bmatrix}
\end{align*}
\begin{align*}
\mu_2^* = \begin{bmatrix}
\sum_{j,k,u} {\color{orange} [U_{\tau - 1}^k = u]} \langle {\color{orange} u_{\pi}(\pi)} \rangle_k \langle {\color{orange} u_{S_{\tau - 1}}(S_{\tau - 1})} \rangle_j \langle {\color{orange} u_{B}(\bm{B})} \rangle_{u 1 j}\\
...\\
\sum_{j,k,u} {\color{orange} [U_{\tau - 1}^k = u]} \langle {\color{orange} u_{\pi}(\pi)} \rangle_k \langle {\color{orange} u_{S_{\tau - 1}}(S_{\tau - 1})} \rangle_j \langle {\color{orange} u_{B}(\bm{B})} \rangle_{u |S| j} \rangle 
\end{bmatrix} 
\end{align*}
\begin{align*}
\mu_3^* = \begin{bmatrix}
\sum_i {\color{purple} [O_\tau = i]}\langle {\color{purple} u_{A}(\bm{A})} \rangle_{i 1} \\
...\\
\sum_i {\color{purple} [O_\tau = i]}\langle {\color{purple} u_{A}(\bm{A})} \rangle_{i |S|}
\end{bmatrix}
\end{align*}
\begin{align*}
\mu_4^* = \begin{bmatrix}
\sum_{j,k,u} {\color{purple} [U_\tau^k = u]} \langle {\color{purple} u_{\pi}(\pi)} \rangle_k \langle {\color{purple} u_{S_{\tau + 1}}(S_{\tau + 1})} \rangle_j \langle {\color{purple} u_{B}(\bm{B})} \rangle_{u 1 j}\\
...\\
\sum_{j,k,u} {\color{purple} [U_\tau^k = u] } \langle {\color{purple} u_{\pi}(\pi)} \rangle_k \langle {\color{purple} u_{S_{\tau + 1}}(S_{\tau + 1})} \rangle_j \langle {\color{purple} u_{B}(\bm{B})} \rangle_{u |S| j}
\end{bmatrix}.
\end{align*}

Finally, the last step consists of computing the expectations of all sufficient statistics as follows:

\begin{itemize}
\item $\langle u_{D}(\bm{D})\rangle_{i} = \langle \ln \bm{D}_i \rangle = \psi(\bm{d}_i) - \psi(\sum_r \bm{d}_r) \delequal \bar{\bm{D}}_i$
\item $\langle u_{\pi}(\pi)\rangle_{j} = \langle [\pi = j] \rangle = \tilde{\alpha}_j$
\item $\langle u_{S_{\tau - 1}}(S_{\tau - 1}) \rangle_{k} = \langle [S_{\tau - 1} = k] \rangle = \tilde{\bm{D}}_{(\tau - 1)k}$
\item $\langle u_{B}(\bm{B}) \rangle_{lmn} = \langle \ln \bm{B}[l]_{m n} \rangle = \psi(\bm{b}[l]_{m n}) - \psi(\sum_r \bm{b}[l]_{r n}) \delequal \bar{\bm{B}}[l]_{m n}$
\item $\langle u_{A}(\bm{A}) \rangle_{op} = \langle \ln \bm{A}_{o p} \rangle = \psi(\bm{a}_{op}) - \psi(\sum_r \bm{a}_{r p}) \delequal \bar{\bm{A}}_{op}$
\item $\langle u_{S_{\tau + 1}}(S_{\tau + 1})\rangle_{q} = \langle [S_{\tau + 1} = q] \rangle = 
\tilde{\bm{D}}_{(\tau + 1)q}$
\end{itemize}

Substituting those expectations into the equations for the $\mu_i^*$ leads to the following results: $\mu_1^* = \bar{\bm{D}}$, $\mu_2^* = \sum_{k} \tilde{\alpha}_k \bar{\bm{B}}[U_\tau^k]\tilde{\bm{D}}_{\tau - 1}$, $\mu_3^* = \bm{o}_\tau \cdot \bar{\bm{A}}$ and $\mu_4^* = \sum_{k} \tilde{\alpha}_k \tilde{\bm{D}}_{\tau + 1} \cdot \bar{\bm{B}}[U_\tau^k]$. Where $\bm{o}_\tau$ is a one hot vector containing the observation made by the agent and we used the fact that the indicator function $[U_\tau^k = u]$ filters out elements from the sum where $u \neq U_\tau^k$. The final result is obtained by substituting the values of the $\mu_i^*$'s in Equation \ref{eq:S0:222} to obtain the following categorical distribution:

\begin{align*}
Q^*(S_\tau) &\propto \exp \Big\{ \mu_{S_\tau}^* \cdot u_{S_\tau}(S_\tau) \Big\}
\end{align*}
$$\mu_{S_\tau}^* = [\tau = 0] {\color{orange} \bar{\bm{D}}} + [\tau \neq 0] \sum_{k} {\color{orange} \tilde{\alpha}_k \bar{\bm{B}}[U_{\tau - 1}^k]\tilde{\bm{D}}_{\tau - 1}} + [\tau \leq t] {\color{purple} \bm{o}_\tau \cdot \bar{\bm{A}}} + [\tau \neq T] \sum_{k} {\color{purple} \tilde{\alpha}_k \bar{\bm{B}}[U_\tau^k]\tilde{\bm{D}}_{\tau + 1}}.$$

\section*{Appendix H: Derivation of the new expected free energy.}

In this appendix, we derive the expected free energy of our new model. First, we restate the factorisation of the generative model and the variational distribution:
\begin{align} \label{eq:gm_app_h}
P(O_{0:t}, S_{0:T}, \pi, \bm{A}, \bm{B}, \bm{D}, \alpha)\,\, =\,\,\,\, &P(\pi|\alpha) P(\alpha) P(\bm{A}) P(\bm{B}) P(S_0|\bm{D}) P(\bm{D})\nonumber\\
&\prod^{t}_{\tau = 0} P(O_\tau|S_\tau,\bm{A}) \prod^{T}_{\tau = 1} P(S_\tau|S_{\tau-1},\bm{B},\pi)
\end{align}
\begin{align} \label{eq:vd_app_h}
Q(S_{0:T}, \pi, \bm{A}, \bm{B}, \bm{D}, \alpha) = Q(\pi)Q(\bm{A})Q(\bm{B})Q(\bm{D})Q(\alpha) \prod_{\tau=0}^{T} Q(S_\tau).
\end{align}
Remembering from Appendix C that the expected free energy is defined as:
\begin{align} \label{eq:efe-app-h}
\bm{G}(\pi) = \mathbb{E}_{\tilde{Q}}\Big[\kl{Q(X|\pi)}{P(O_{0:T},X|\pi)}\Big],
\end{align}
where the latent variables are $X = \{S_{0:T}, \bm{A}, \bm{B}, \bm{D}, \alpha\}$, $\tilde{Q} = \tilde{Q}(O_{t+1:T}) \delequal \prod_{\tau = t + 1}^T \tilde{Q}(O_{\tau})$ and $\tilde{Q}(O_{\tau}) \delequal \sum_{S_\tau} \tilde{Q}(O_{\tau}, S_{\tau})$.
Now we substitute Equation \ref{eq:gm_app_h} and \ref{eq:vd_app_h} into Equation \ref{eq:efe-app-h} and simplify by removing the terms that are constant w.r.t the policy $\pi$:
\begin{align*}
\bm{G}(\pi) &= \mathbb{E}_{\tilde{Q}}\Big[\kl{Q(S_{0:T}, \bm{A}, \bm{B}, \bm{D}, \alpha | \pi)}{P(O_{0:T}, S_{0:T}, \bm{A}, \bm{B}, \bm{D}, \alpha | \pi)}\Big]{\color{white}\sum}\\
&= \kl{Q(\bm{A})}{P(\bm{A})} + \kl{Q(\bm{B})}{P(\bm{B})} + \kl{Q(\bm{D})}{P(\bm{D})}{\color{white}\sum}\\
&+ \kl{Q(\alpha)}{P(\alpha)} +  \mathbb{E}_{Q(\bm{D})}\Big[\kl{Q(S_0)}{P(S_0|\bm{D})}\Big] {\color{white}\sum}\\
&+ \sum_{\tau=1}^{T} \mathbb{E}_{Q(S_{\tau-1},\bm{B})}\Big[ \kl{Q(S_\tau)}{P(S_\tau|S_{\tau-1},\bm{B},\pi)}\Big]\\
&- \sum_{\tau=0}^{T} \mathbb{E}_{Q(S_{\tau},\bm{A})\tilde{Q}(O_{t+1:T})}\Big[ \ln P(O_\tau|S_{\tau},\bm{A}) \Big]\\
&= \sum_{\tau=1}^{T} \mathbb{E}_{Q(S_{\tau-1},\bm{B})}\Big[ \kl{Q(S_\tau)}{P(S_\tau|S_{\tau-1},\bm{B},\pi)} \Big] + C {\color{white}\sum_{\tau=0}^{T}}\\
&= \sum_{\tau=1}^{T} \mathbb{E}_{Q(S_{\tau},S_{\tau-1},\bm{B})}\Big[ \ln Q(S_\tau) - \ln P(S_\tau|S_{\tau-1},\bm{B},\pi)\Big] + C {\color{white}\sum_{\tau=0}^{T}}\\
&= \sum_{\tau=1}^{T} \mathbb{E}_{Q(S_{\tau-1},\bm{B})}\Big[ \underbrace{- \mathbb{E}_{Q(S_{\tau})}\big[ \ln P(S_\tau|S_{\tau-1},\bm{B},\pi)\big]}_{H[\bigcdot]} \Big] + C {\color{white}\sum_{\tau=0}^{T}}\\
&= \sum_{\tau=1}^{T} \mathbb{E}_{Q(S_{\tau-1},\bm{B})}\Big[ H \big[ P(S_\tau|S_{\tau-1},\bm{B},\pi)\big] \Big] + C {\color{white}\sum_{\tau=0}^{T}},
\end{align*}
where $H[\bigcdot]$ refer to $- \mathbb{E}_{Q(S_{\tau})}\big[ \ln P(S_\tau|S_{\tau-1},\bm{B},\pi)\big]$ in the last equation.

\end{document}